\documentclass[twoside]{article}
\usepackage[accepted]{aistats2019}
\usepackage[round]{natbib}

\usepackage[utf8]{inputenc} 
\usepackage[T1]{fontenc}    
\usepackage{hyperref}       
\usepackage{url}            
\usepackage{booktabs}       
\usepackage{amsfonts}       
\usepackage{nicefrac}       
\usepackage{microtype}      

\usepackage[round]{natbib}
\usepackage{setspace}
\usepackage{geometry}
\usepackage{placeins}
\usepackage{graphicx}
\usepackage{xcolor}
\usepackage{titlesec}
\usepackage[page]{appendix}
\usepackage{enumerate}
\usepackage[shortlabels]{enumitem}

\usepackage{lscape}
\usepackage{booktabs}
\usepackage{rotating}
\usepackage{multirow}
\usepackage{longtable}
\usepackage{caption}
\usepackage{subcaption}
\usepackage{float}
\usepackage{tabularx}
\usepackage{ragged2e}
\newcolumntype{Y}{>{\RaggedRight\arraybackslash}X}
\usepackage{pdflscape}
\usepackage{afterpage}

\usepackage{amsmath}
\usepackage{amssymb}
\usepackage{amsthm}
\usepackage{mathtools}
\usepackage{dsfont}

\usepackage[ruled,vlined]{algorithm2e}

\SetCommentSty{algcommstyle}

\usepackage{tikz}
\usetikzlibrary{bayesnet}

\usepackage{textcomp}
\usepackage{sourcecodepro}
\usepackage{listings}
\definecolor{commentgrey}{gray}{0.45}
\definecolor{backgray}{gray}{0.96}
\DeclareMathOperator*{\argmax}{arg\,max}
\DeclareMathOperator*{\argmin}{arg\,min}
\DeclareMathOperator*{\tr}{tr}
\DeclarePairedDelimiter{\abs}{\lvert}{\rvert}
\DeclarePairedDelimiter{\norm}{\lVert}{\rVert}
\DeclarePairedDelimiter{\inner}{\langle}{\rangle}

\newcommand*{\st}{\text{ s.t. }}

\newcommand*{\R}{\mathbb R}
\newcommand*{\I}{\mathds{1}}

\newcommand*{\cond}{\;\ifnum\currentgrouptype=16 \middle\fi|\;}

\newcommand*{\dif}{\,d}

\newcommand*{\m}[1]{\textbf{#1}}

\newcommand*{\ttilde}{{\raise.17ex\hbox{$\scriptstyle\sim$}}}

\makeatletter
\newsavebox{\mybox}\newsavebox{\mysim}
\newcommand*{\distas}[1]{%
  \savebox{\mybox}{\hbox{\kern3pt$\scriptstyle#1$\kern3pt}}%
  \savebox{\mysim}{\hbox{$\sim$}}%
  \mathbin{\overset{#1}{\kern\z@\resizebox{\wd\mybox}{\ht\mysim}{$\sim$}}}%
}
\makeatother
\newcommand*{\dist}{\sim}

\makeatletter
\def\moverlay{\mathpalette\mov@rlay}
\def\mov@rlay#1#2{\leavevmode\vtop{%
   \baselineskip\z@skip \lineskiplimit-\maxdimen
   \ialign{\hfil$\m@th#1##$\hfil\cr#2\crcr}}}
\newcommand*{\charfusion}[3][\mathord]{
  #1{\ifx#1\mathop\vphantom{#2}\fi\mathpalette\mov@rlay{#2\cr#3}}
  \ifx#1\mathop\expandafter\displaylimits\fi}
\makeatother

\newcommand*{\mt}[1]{\text{\normalfont #1}}

\newtheorem{theorem*}{Theorem}

\newtheorem{corollary*}[theorem*]{Corollary}

\newtheorem{proposition*}[theorem*]{Proposition}

\newtheorem{lemma*}[theorem*]{Lemma}

\theoremstyle{definition}

\newtheorem{definition*}{Definition}
\newtheorem*{definition**}{Definition}

\newtheorem*{remark}{Remark}

\newtheoremstyle{algodesc}{}{}{}{}{\bfseries}{.}{ }{}%
\theoremstyle{algodesc}

\newcommand*{\Gammahat}{\hat{\Gamma}_{n}}

\begin{document}

\runningauthor{Fan, Jang, Sun, Zhou}

\twocolumn[

\aistatstitle{Precision Matrix Estimation with Noisy and Missing Data}

\aistatsauthor{
    Roger Fan\textsuperscript{1} \And
    Byoungwook Jang\textsuperscript{1} \And
    Yuekai Sun\textsuperscript{1} \And
    Shuheng Zhou\textsuperscript{2} }
\vspace{5pt}
\aistatsaddress{
    \textsuperscript{1}University of Michigan \\
    \textsuperscript{2}University of California, Riverside} ]

\begin{abstract}
Estimating conditional dependence graphs and precision matrices are some of the most common problems in modern statistics and machine learning. When data are fully observed, penalized maximum likelihood-type estimators have become standard tools for estimating graphical models under sparsity conditions. Extensions of these methods to more complex settings where data are contaminated with additive or multiplicative noise have been developed in recent years. In these settings, however, the relative performance of different methods is not well understood and algorithmic gaps still exist. In particular, in high-dimensional settings these methods require using non-positive semidefinite matrices as inputs, presenting novel optimization challenges. We develop an alternating direction method of multipliers (ADMM) algorithm for these problems, providing a feasible algorithm to estimate precision matrices with indefinite input and potentially nonconvex penalties. We compare this method with existing alternative solutions and empirically characterize the tradeoffs between them. Finally, we use this method to explore the networks among US senators estimated from voting records data.
\end{abstract}

\section{Introduction}

Undirected graphs are often used to describe high-dimensional distributions. Under sparsity conditions, these graphs can be estimated using penalized methods such as
\begin{equation} \label{eq:glasso_opt_problem}
	\hat{\Theta} \in
	 \argmin_{\Theta \succeq 0} \left\{ \tr(\Gammahat \Theta) - \log \det (\Theta) + g_\lambda(\Theta) \right\},
\end{equation}
where $\Gammahat$ is the sample covariance or correlation matrix and $g_\lambda$ is a separable (entry-wise) sparsity-inducing penalty function. Although this approach has proven successful in a variety of application areas such as neuroscience and genomics, its soundness hinges on the positive semidefiniteness (PSD) of $\Gammahat$. If $\Gammahat$ is indefinite, the objective may be unbounded from below.

In order to ensure this penalized $M$-estimator is well-behaving, \cite{lohwainwright15} impose a side constraint of the form $\rho(\Theta) < R$, where $\rho$ is a convex function. Here we focus on the estimator using the operator norm as a side constraint
\begin{equation} \label{eq:glasso_opt_problem_side}
\hat{\Theta}
  \in \argmin_{\Theta \succeq 0, \norm{\Theta}_2 \leq R} \left\{ \tr(\Gammahat \Theta) - \log \det (\Theta) + g_\lambda(\Theta) \right\}.
\end{equation}
\cite{lohwainwright17} adopt this method and show in theory the superior statistical properties of this constrained estimator.
Their results suggest that the addition of a side constraint is not only sufficient but also almost necessary to effectively untangle the aforementioned complications.

Unfortunately, this additional constraint precludes using existing methods to solve the penalized objective with non-PSD input. To close this gap, we develop an alternating direction method of multipliers (ADMM) algorithm to implement \eqref{eq:glasso_opt_problem_side} efficiently. We conduct empirical studies comparing this new method to several other precision matrix estimators. Our simulation study reveals several trends that are not present in the fully observed case. Finally, we illustrate the performance of our methods in analyzing the US senate voting data, uncovering both known and novel phenomena from the modern political landscape.

The remainder of this paper is organized as follows. In Section~2, we provide an overview of existing related work and describe in detail the optimization issues that arise from indefinite inputs and nonconvex penalties.
In Section~3, we present the proposed ADMM algorithm and present some convergence results.
Section~4 provides numerical examples and comparisons.
Section~5 presents an exploratory analysis of US Senate voting records data using this method and details several interesting conclusions that can be drawn from the estimated graphs.
Finally, we summarize the empirical results and
their practical implications regarding choice of method
in Section~6.

\section{Problem formulation and existing work}
\label{sec:problemformulation}

There is a wide body of work proposing methods to perform precision matrix estimation in the fully observed case, including \citet{meinshausenbuhlmann06}, \citet{yuanlin07modselggm}, \citet{rothmanetal08}, \citet{friedmanhastietib08graphicallasso}, \citet{banerjeeetal08sparsemlegaussian}, and \citet{zhouetal10}, most of which are essentially a $\ell_1$-penalized likelihood approach \eqref{eq:glasso_opt_problem} which we will refer to as the graphical Lasso.

Recent work has focused on using nonconvex regularizers such as SCAD and MCP for model selection in the regression setting \citep{fan2001variable,zhang10mcp,brehenyhuang11coorddescnonconvex,zhangzhang12concavereg}. \citet{lohwainwright15,lohwainwright17} extend this analysis to general $M$-estimators, including variants of the graphical Lasso objective, and show their statistical convergence and support recovery properties. Estimators with these penalties have been shown to attain model selection under weaker theoretical conditions, but require more sophisticated optimization algorithms to solve, such as the local linear approximation (LLA) method of \citet{fanetal14oracleoptfoldedconcave}.

In a fully observed and noiseless setting, $\hat{\Gamma}_n$ is the sample covariance and guaranteed to be at least positive semidefinite. Then, if $g_\lambda$ is the $\ell_1$-penalty, the objective of \eqref{eq:glasso_opt_problem} is convex and bounded from below. In this setting, one can show that for $\lambda > 0$ a unique optimum $\hat{\Theta}$ exists with bounded eigenvalues and that the iterates for any descent algorithm will also have bounded eigenvalues \citep[for example, see Lemma 2 in][]{hsieh_etal14quic}.

When working with missing, corrupted, and dependent data, the likelihood is nonconvex, and the expectation-maximization (EM) algorithm has traditionally been used to perform statistical inference. However, in these noisy settings, the convergence of the EM algorithm is difficult to guarantee and is often slow in practice. For instance, \citet{stadler2012missing} implement a likelihood-based method for inverse covariance estimation with missing values, but their EM algorithm requires solving a full graphical Lasso optimization problem in each M-step.

An alternative approach is to develop $M$-estimators that account for missing and corrupted data. For graphical models, \citet{lohwainwright15} establish that the graphical Lasso, including a version using nonconvex penalties, can be modified to accommodate noisy or missing data by adjusting the sample covariance estimate.

These modified estimators depend on the observation that statistical theory for the graphical Lasso generally requires that $\norm{\Gammahat - \Sigma}_\infty$ converges to zero at a sufficiently fast rate \citep[e.g.][]{rothmanetal08,zhouetal10,lohwainwright17}. When considering missing or corrupted data, it is often possible to construct covariance estimates $\Gammahat$ that satisfy this convergence criteria but are not necessarily positive semidefinite. In fact, in high-dimensional settings $\Gammahat$ may even be guaranteed to be indefinite. Attempting to input these indefinite covariance estimates into the graphical Lasso, however, presents novel optimization issues.

\textbf{Unbounded objective.} When attempting to move beyond the $\ell_1$ penalized case with positive semidefinite input, the problem in \eqref{eq:glasso_opt_problem} becomes unbounded from below, so an optimum may not necessarily exist. This issue comes from two potential sources: 1) negative eigenvalues in $\Gammahat$, or 2) zero eigenvalues combined with the boundedness of the nonconvex penalty $g_\lambda$.
For example, consider the restriction of the objective in \eqref{eq:glasso_opt_problem} to a ray defined by an eigenvalue-vector pair $\sigma_1, v_1$ of $\Gammahat$:
\begin{equation} \label{eq:glasso_unbounded}
\begin{aligned}
& f(I + t v_1 v_1^T)  \\
  & \quad =  \tr(\Gammahat) + t \tr(\Gammahat v_1 v_1^T) - \log ( 1+t ) + g_\lambda(t v_1 v_1^T) \\
  & \quad =  \tr(\Gammahat) + t \sigma_1 - \log ( 1+t ) + g_\lambda(t v_1 v_1^T).
\end{aligned}
\end{equation}
If $\sigma_1 < 0$, we see that $f$ is unbounded from below due to the $t\sigma_1$ and $- \log(1 + t)$ terms. In fact, if $\sigma_1 = 0$ and $g_\lambda$ is bounded from above, as is the case when using standard nonconvex penalties, the objective is also unbounded from below.

So unboundedness can occur anytime there is a negative eigenvalue in the input matrix, or whenever there are zero eigenvalues combined with a nonconvex penalty function $g_\lambda$. Unboundedness creates optimization issues, as an optimum no longer necessarily exists.

\textbf{Handling unboundedness.} In order to guarantee that an optimum exists for \eqref{eq:glasso_opt_problem}, an additional constraint of the form $\rho(\Theta) \leq R$ can be imposed, where $\rho$ is some convex function.
In this paper, we consider the estimator \eqref{eq:glasso_opt_problem_side}, which uses a side constraint of the form $\norm{\Theta}_2 \leq R$. \cite{lohwainwright17} show the rates of convergence of this estimator \eqref{eq:glasso_opt_problem_side} and show that it can attain model selection consistency and spectral norm convergence without the incoherence assumption when used with a nonconvex penalty (see Appendix~E therein), but do not discuss implementation or optimization aspects of the problem.

To our knowledge, there is currently no feasible optimization algorithm for the estimator defined in \eqref{eq:glasso_opt_problem_side}, particularly when the input is indefinite.
\citet{lohwainwright15} present a composite gradient descent method for optimizing a subset of side-constrained versions of \eqref{eq:glasso_opt_problem}. However, their algorithm requires a side constraint  of the form
$\rho(\Theta)
  = \frac{1}{\lambda} ( g_\lambda(\Theta) + \frac{\mu}{2} \norm{\Theta}_F^2 )$,
which does not include the spectral norm constraint and therefore cannot attain the better theoretical results it achieves (Section~\ref{subsec:side_comp} compares the performance of different side constraints).
It may be possible to develop heuristic algorithms that alternate performing a proximal gradient update ignoring the side constraint and projecting to the constraint set, but as far as we know there has not been any analysis of algorithms of this type (we discuss this in more detail in Section~\ref{subsec:optimization_performance}).

An alternative approach to solving this unbounded issue is to project the input matrix $\Gammahat$ to the positive semidefinite cone before inputting into \eqref{eq:glasso_opt_problem}. We discuss this further in Section~\ref{sec:alternatives}, but this only solves the unbounded issue when using the $\ell_1$ penalty; nonconvex penalties still require a side constraint to have a bounded objective and therefore our algorithm is still useful even for the projected methods.

\section{ADMM Algorithm}
\label{sec:ADMM}

Our algorithm is similar to the algorithm in \citet{guozhang17l1condnumprec}, which applies ADMM to the closely related problem of condition number-constrained sparse precision matrix estimation using the same splitting scheme as below. We discuss their method in more detail in Section~\ref{sec:guozhang_comp}. The following algorithm is specialized to the case where the spectral norm is used as the side constraint. In Section~\ref{sec:admm_genside} we derive a similar ADMM algorithm that can be used for any side constraint with a computable projection operator.

Rewrite the objective from \eqref{eq:glasso_opt_problem_side} as
\begin{equation} \label{eq:glasso_obj_foradmm}
f(\Theta) = \tr(\Gammahat \Theta) - \log \det (\Theta) + g_\lambda(\Theta) + \I_{\mathcal{X}_R} (\Theta)
\end{equation}
where $\mathcal{X}_R = \{\Theta : \Theta \succeq 0, \norm{\Theta}_2 \leq R\}$ and
$\I_\mathcal{X} (\Theta) = 0 \mt{ if } \Theta \in \mathcal{X} \mt{ and } \infty \mt{ otherwise}$.

Let $\rho>0$ be a penalty parameter and let $\mt{Prox}_{g_{\lambda} / \rho}$ be the prox operator of $g_\lambda / \rho$. We derive these updates for SCAD and MCP in Section~\ref{subsec:ADMM_derivation}. Let $T_\rho(A)$ be the following prox operator for $- \log \det \Theta + \I_{\mathcal{X}_R}(\Theta)$, which we derive in Section~\ref{sec:logdet_prox_ind},
\begin{equation*}
\begin{aligned}
  T_\rho(A) & = T_\rho(U M U^T) = U \tilde{D} U^T \\
  & \quad \mt{ where } \tilde{D}_{ii} = \min \left\{ \frac{M_{ii} + (M_{ii}^2 + \frac{4}{\rho})^{1/2}}{2}, R \right\},
\end{aligned}
\end{equation*}
where $U M U^T$ is the eigendecomposition of $A$. Then the ADMM algorithm for solving \eqref{eq:glasso_obj_foradmm}, which we derive in Section~\ref{subsec:ADMM_derivation}, is described in Algorithm~\ref{alg:admm_indefinite}. Computationally this algorithm is dominated by the eigendecomposition used to evaluate $T_\rho$, and therefore has a complexity of $O(m^3)$, which matches the scaling of other graphical Lasso solvers \citep[e.g.][]{meinshausenbuhlmann06,friedmanhastietib08graphicallasso,hsieh_etal14quic}.
\begin{algorithm}[t]
\caption{ADMM for graphical Lasso with a side constraint} \label{alg:admm_indefinite}
  \SetAlgoLined
  \KwIn{$\Gammahat$, $\rho$, $g_\lambda$, $R$}
  \KwOut{$\hat{\Theta}$}
  Initialize $V^0 = \Theta^0 \succ 0$, $\Lambda^0 = \m{0}$ \;
  \While{not converged}{
    $V^{k+1} = \mt{Prox}_{g_{\lambda} / \rho} \left( \frac{\rho \Theta^k + \Lambda^k}{\rho} \right)$\\
    $\Theta^{k+1} = T_\rho \left( \frac{\rho V^{k+1} - \Gammahat - \Lambda^k}{\rho} \right)$\\
    $\Lambda^{k+1} = \Lambda^k + \rho(\Theta^{k+1} - V^{k+1})$
  }
\end{algorithm}

\subsection{Convergence}
\label{subsec:ADMM_convergence}

The following proposition applies standard results on the convergence of ADMM for convex problems to show convergence when the $\ell_1$ penalty is used. Details are in Section~\ref{subsec:admm_conv_convex_pf}.

\begin{proposition*} \label{cor:admm_conv_convex}
If the penalty is convex and satisfies the conditions in Section~\ref{sec:nonconvex_penalties}, Algorithm~\ref{alg:admm_indefinite} converges to a global minimum of \eqref{eq:glasso_obj_foradmm}.
\end{proposition*}
\begin{remark}
Regarding the nonconvex penalty, recent work has established ADMM convergence results in some nonconvex settings \citep[see][]{hong16convadmm,wangeta15_admm_nonconvex}, but to our knowledge there is no convergence result that encompasses this nonsmooth and nonconvex application. We can show convergence if a fairly strong assumption is made on the iterates, but we are currently working on extending existing results to this case.
\end{remark}

Proposition~\ref{thm:admm_stationary} shows that any limiting point of Algorithm~\ref{alg:admm_indefinite} is a stationary point of the original objective \eqref{eq:glasso_obj_foradmm}. This is proved in Section~\ref{subsec:admm_stationary_pf}. When using the $\ell_1$ penalty or a nonconvex penalty with $R \leq \sqrt{2/\mu}$, where $\mu$ is the weak convexity constant of $g_\lambda$, the objective $f$ is convex and therefore any stationary point is unique and also the global optimum. See Section~\ref{subsec:R_mu_cond} for a more detailed discussion.

\begin{proposition*} \label{thm:admm_stationary}
Assume that the penalty $g_\lambda$ satisfies the conditions in Section~\ref{sec:nonconvex_penalties}. Then for any limit point $(\Theta^*, V^*, \Lambda^*)$ of the ADMM algorithm defined in Algorithm~\ref{alg:admm_indefinite}, $\Theta^*$ is also a stationary point of the objective $f$ as defined in \eqref{eq:glasso_obj_foradmm}.
\end{proposition*}
The assumptions on $g_\lambda$ in Section~\ref{sec:nonconvex_penalties} are the same as those assumed in \citet{lohwainwright15,lohwainwright17}, and are satisfied by the Lasso, SCAD, and MCP functions.

Note that if a limiting point is found to exist when using a nonconvex penalty the result in Proposition~\ref{thm:admm_stationary} will still hold. Empirically we find that the algorithm performs well and converges consistently when used with nonconvex penalties, but there is no existing theoretical guarantee that a limiting point of ADMM will exist in that setting.

\section{Simulations}
We evaluate the proposed estimators using the relative Frobenius norm and the sum of the false positive rate and false negative rate ($\mt{FPR}+\mt{FNR}$). We present results over a range of $\lambda$ values, noting that all the compared methods would use similar techniques to perform model tuning. Section~\ref{subsec:tuning} presents an example of how to use BIC or cross-validation to tune these methods. We present results using covariance matrices from auto-regressive and Erd\H{o}s-R\'{e}nyi random graph models. See Section~\ref{supp:simulation_supp} for descriptions of these models as well as additional simulation results.

\subsection{Alternative methods}
\label{sec:alternatives}

When faced with indefinite input, there are two alternative graphical Lasso-style estimators that can be used besides \eqref{eq:glasso_opt_problem_side}, which involve either $\ell_\infty$ projection to the positive semidefinite cone or nodewise regression in the style of \citet{meinshausenbuhlmann06}.

\textbf{Projection.}
Given an indefinite input matrix $\Gammahat$, \citet{park16quantile_errorinvar} and \citet{greenewaldetal17timedepspatialgraph} propose performing the projection
$\Gammahat^+ = \argmin_{\Gamma \succeq 0} \norm{\Gamma - \Gammahat}_\infty$.
They then input $\Gammahat^+$ into the optimization problem \eqref{eq:glasso_opt_problem}. This is similar to the projection done in \citet{datta2017cocolasso}. In terms of the upper bound on statistical convergence rates, this method pays a constant factor cost, though in practice projection may result in a loss of information and therefore a decrease in efficiency.

After projecting the input, existing algorithms can be used to optimize \eqref{eq:glasso_opt_problem} with the $\ell_1$ penalty. However, as mentioned in Section~\ref{sec:problemformulation}, using a nonconvex penalty still leads to an unbounded objective and therefore still requires using our ADMM algorithm to solve \eqref{eq:glasso_opt_problem_side}.

\textbf{Nodewise regression.}
\citet{lohwainwright12} and \citet{rudelsonzhou17errinvardependent} both study the statistical and computational convergence properties of using errors-in-variables regression to handle indefinite input matrices in high-dimensional settings. Following the nodewise regression ideas of \citet{meinshausenbuhlmann06} and \citet{yuan10highdiminvcovest}, we can perform $m$ Lasso-type regressions to obtain estimates $\hat{\beta}_j$ and form estimates $\hat{a}_j$, where
\begin{equation} \label{eq:nodewise_lassoreg}
\begin{aligned}
\hat{\beta}_j & \in \argmin_{\norm{\beta}_1 \leq R} \left\{ \frac{1}{2} \beta^T \hat{\Gamma}_{n, -j, -j} \beta - \inner{\hat{\Gamma}_{n, -j, j}, \beta} + \lambda \norm{\beta}_1 \right\} \\
\hat{a}_j & = - (\hat{\Gamma}_{n, j, j} - \inner{\hat{\Gamma}_{n, -j, j}, \hat{\beta}_j})^{-1}
\end{aligned}
\end{equation}
and combine to get $\tilde{\Theta}$ with $\tilde{\Theta}_{-j, j} = \hat{a}_j \hat{\beta}_j$ and $\tilde{\Theta}_{j, j} = - \hat{a_j}$. Finally, we symmetrize the result to obtain $\hat{\Theta} = \argmin_{\Theta \in S^m} \norm{\Theta - \tilde{\Theta}}_1$, where $S^m$ is the set of symmetric matrices.

These types of nodewise estimators have gained popularity as they require less restrictive incoherence conditions to attain model selection consistency and often perform better in practice in the fully observed case. They have not, however, been as well studied when used with indefinite input.

\subsection{Data models}
\label{sec:data_models}

We test these methods on two models that result in indefinite covariance estimators, the non-separable Kronecker sum model from \citet{rudelsonzhou17errinvardependent} and the missing data graphical model described in \citet{lohwainwright15}. In the main paper we focus on the missing data model, but Section~\ref{supp:simulation_supp} contains a detailed description of the Kronecker sum model as well as simulation results using it.

\begin{figure*}[tbhp] \centering
\begin{subfigure}[t]{0.3\linewidth} \centering
\includegraphics[width=\linewidth]{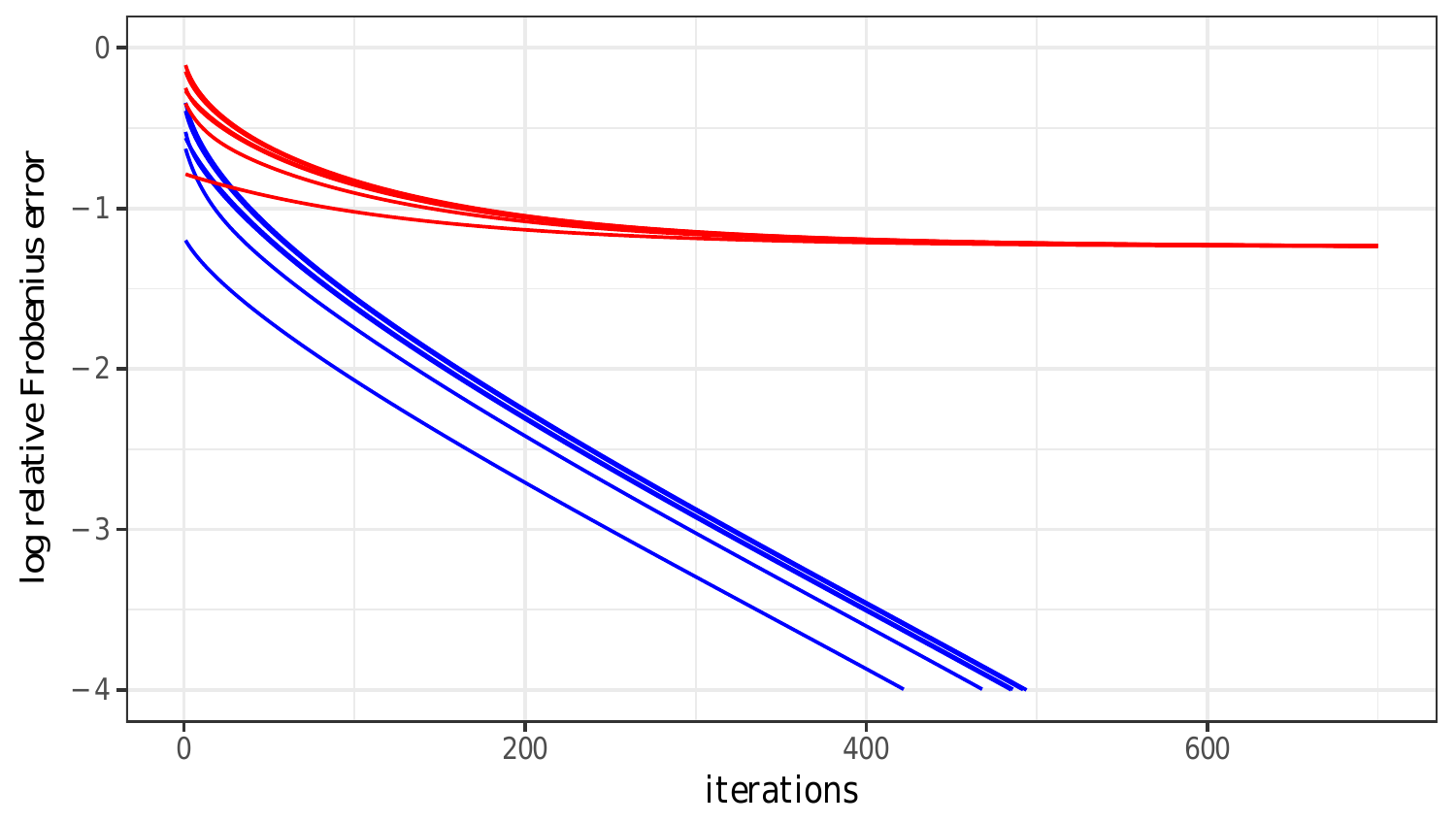}
\caption{$\ell_1$, $\zeta = 0.95$}
\end{subfigure}
\begin{subfigure}[t]{0.3\linewidth} \centering
\includegraphics[width=\linewidth]{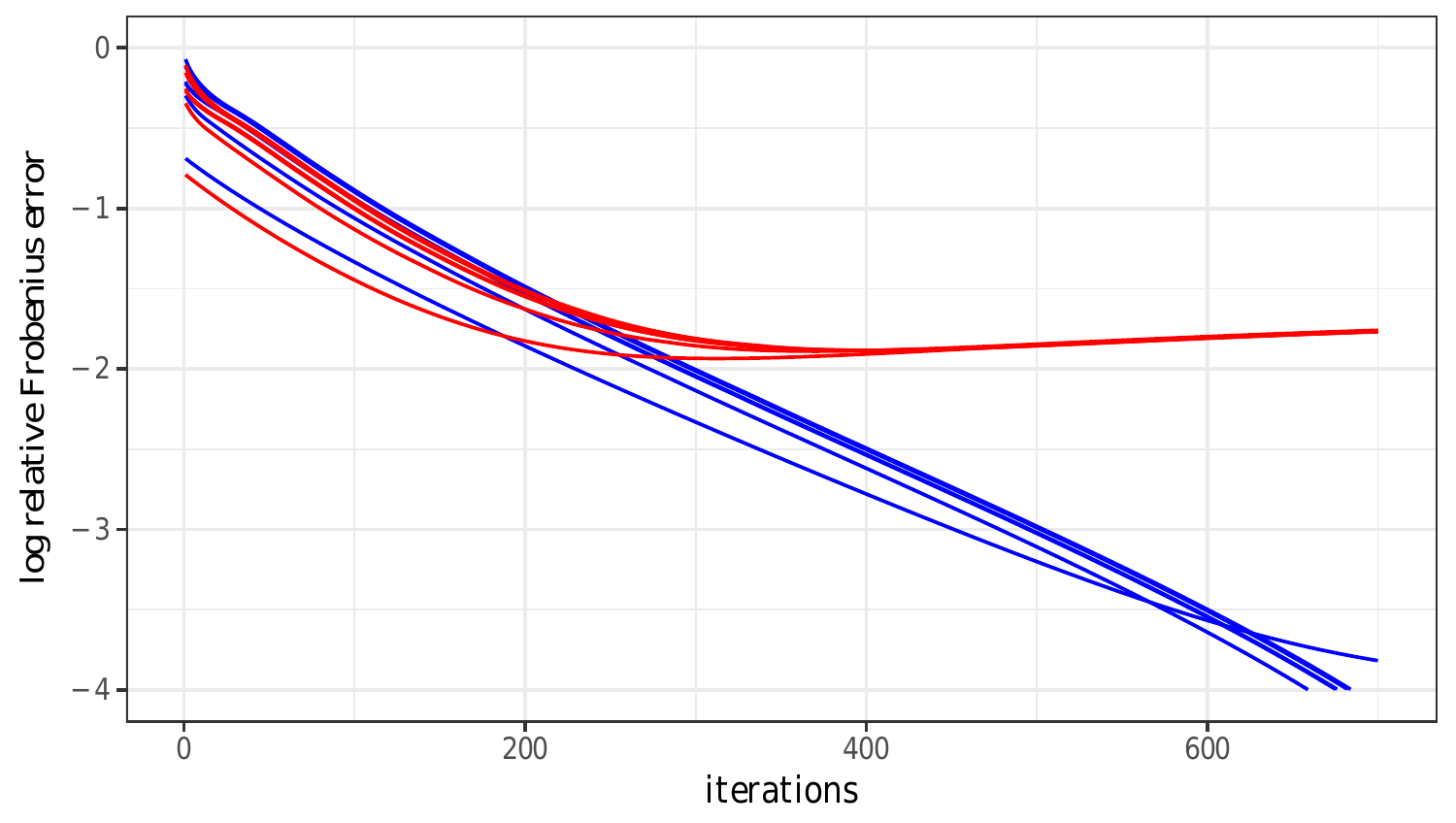}
\caption{MCP, $\zeta = 0.95$}
\end{subfigure} \\
\begin{subfigure}[t]{0.3\linewidth} \centering
\includegraphics[width=\linewidth]{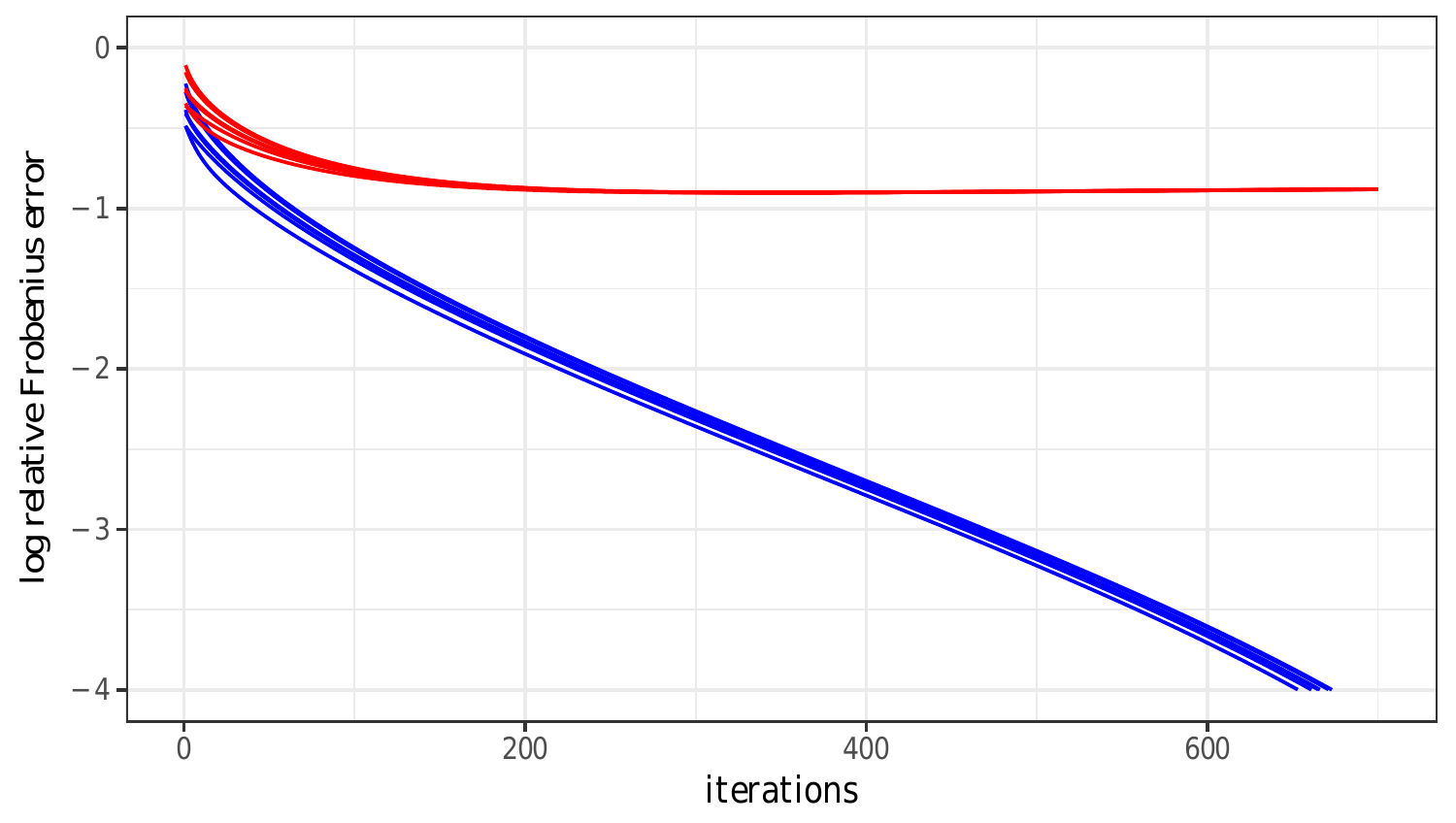}
\caption{$\ell_1$, $\zeta = 0.7$}
\end{subfigure}
\begin{subfigure}[t]{0.3\linewidth} \centering
\includegraphics[width=\linewidth]{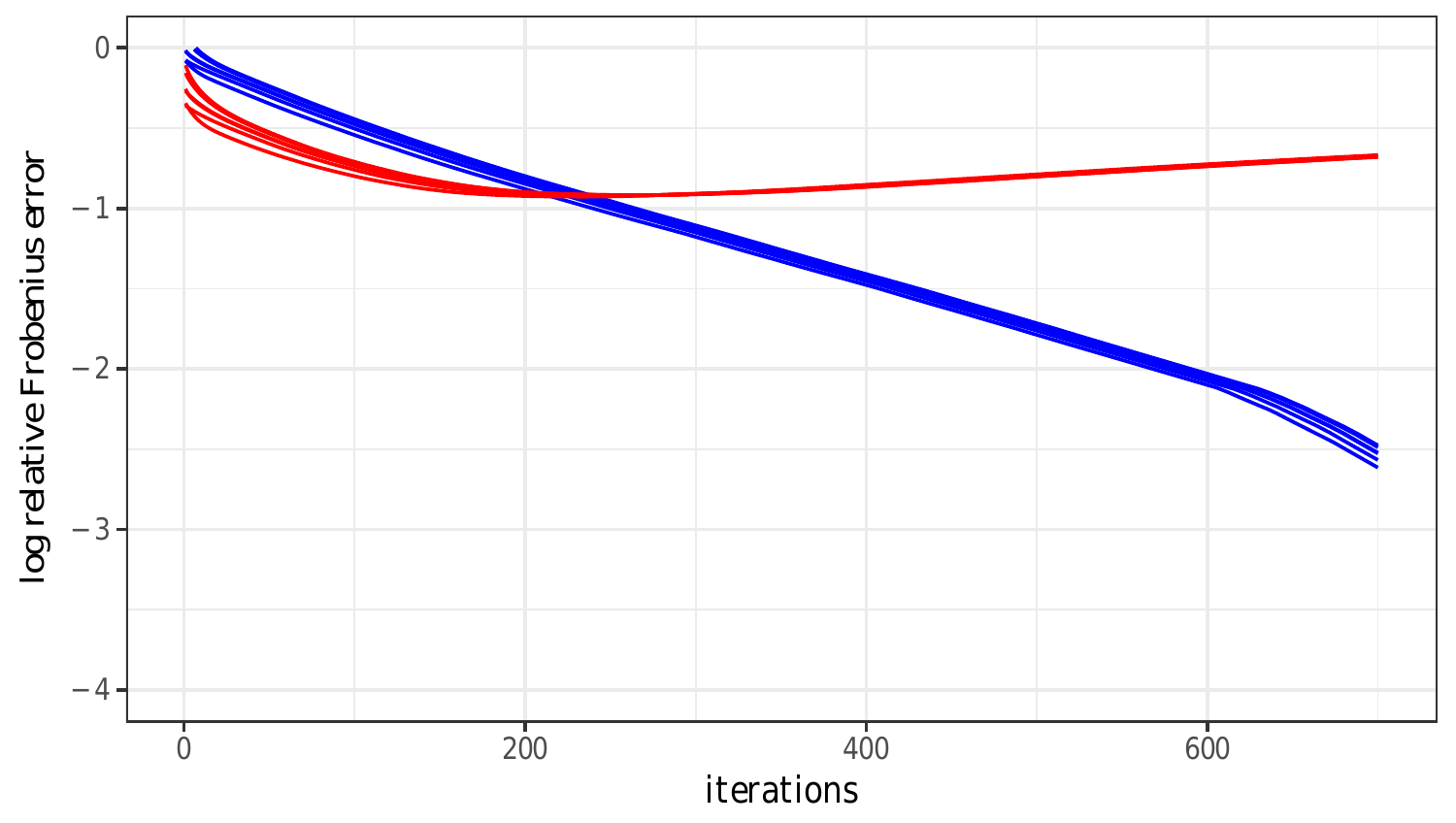}
\caption{MCP, $\zeta = 0.7$}
\end{subfigure}
\caption{Convergence of the ADMM algorithm for several initializations. Blue lines show the relative optimization error ($\norm{\Theta^k - \hat{\Theta}}_F/\norm{\Theta^*}_F$, where $\hat{\Theta}$ is the result of running our algorithm to convergence) while red lines show the statistical error ($\norm{\Theta^k - \Theta^*}_F/\norm{\Theta^*}_F$).
All panels use an $\mt{AR1}(0.7)$ covariance with $m=300$ and $n=125$ and set $\rho=12$. The left panels use an $\ell_1$ penalty, while the right panels use MCP with $a=2.5$. $R$ is set to be three times the oracle spectral norm.}
\label{fig:admm_conv}
\end{figure*}

\textbf{Missing data (MD).} As discussed above, \citet{lohwainwright13discretegraphical,lohwainwright15} propose an estimator for a graphical model with missing-completely-at-random observations.

Let $W \in \R^{n \times m}$ be a mean-zero subgaussian random matrix. Let
$U \in \{0, 1\}^{n \times m}$ where $U_{ij} \dist \mt{Bernoulli}(\zeta_j)$ are independent of $W$. This corresponds to entries of the $j$th column of the data matrix being observed with probability $\zeta_j$. Then we have an unobserved matrix $Z$ and observed matrix $X$ generated by
$Z = W A^{1/2}$ and $X = U \circ X$,
where $\circ$ denotes the Hadamard, or element-wise, product. Here the covariance estimate for $A$ is
\begin{equation}
\Gammahat = \frac{1}{n} X^T X \oslash M
  \mt{ where } M_{k\ell} =
  \begin{cases}
  \zeta_k &\mt{if } k = \ell \\
  \zeta_k \zeta_\ell &\mt{if } k \neq \ell
  \end{cases}
\end{equation}
where $\oslash$ denotes element-wise division. As we divide off-diagonal entries by smaller values, $\Gammahat$ will not necessarily be positive semidefinite.

\subsection{Simulation results}
\label{subsec:sim_optperf}

\textbf{Optimization performance.} Figure~\ref{fig:admm_conv} shows the optimization performance of Algorithm~\ref{alg:admm_indefinite} using nonprojected input matrices from the missing data model with both $\ell_1$ and nonconvex penalties (MCP). The top two panels present an ``easy'' scenario with a higher sampling rate, while the bottom two have a more challenging scenario with significant missing data. Blue lines report the optimization error while red lines are the statistical error.

All the plots in Figure~\ref{fig:admm_conv} have their optimization error quickly converge to below the statistical error. These plots also suggest that our algorithm can attain linear convergence rates. We find that the algorithm consistently converges well over a range of tested scenarios.

Comparing the statistical error of the top two plots, we see that MCP achieves significantly lower error for the easier scenario. But in the bottom two plots, where there is more missing data, it struggles relative to the $\ell_1$ penalty. This is a common trend through our simulations, as the performance of estimators using MCP degrades as missingness increases while the $\ell_1$-penalized versions are more robust.

\begin{figure*}[t] \centering
\includegraphics[width=0.69\linewidth]{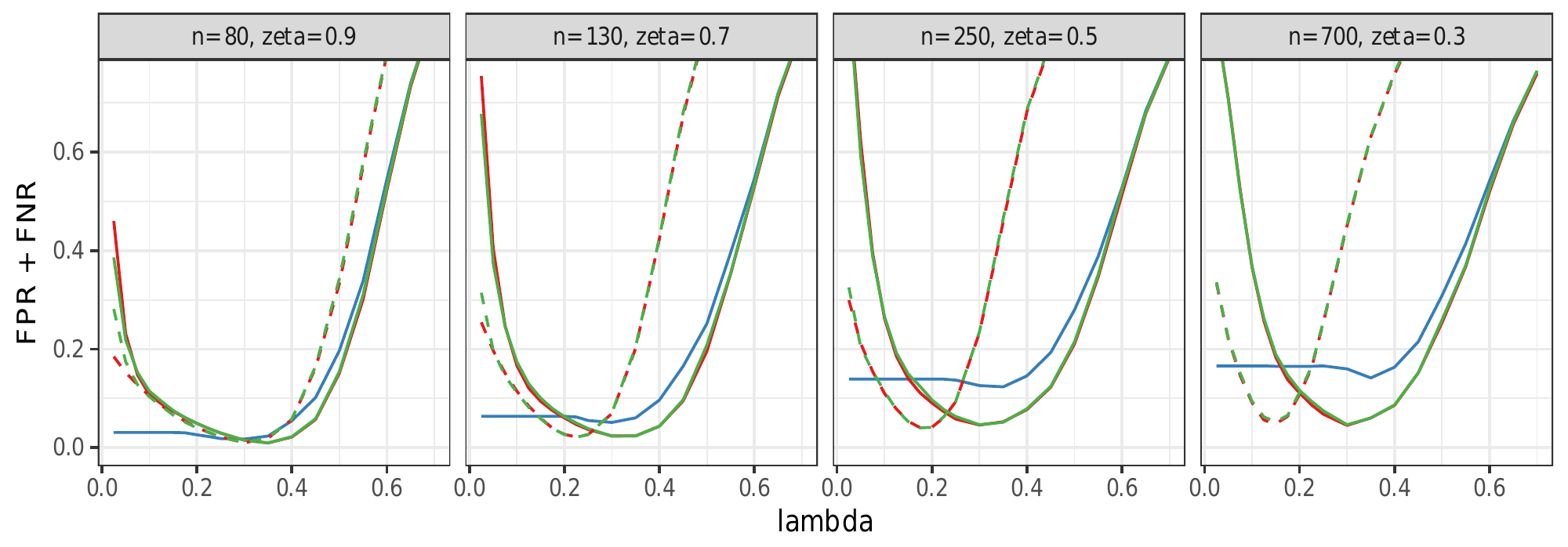} \\
\includegraphics[width=0.69\linewidth]{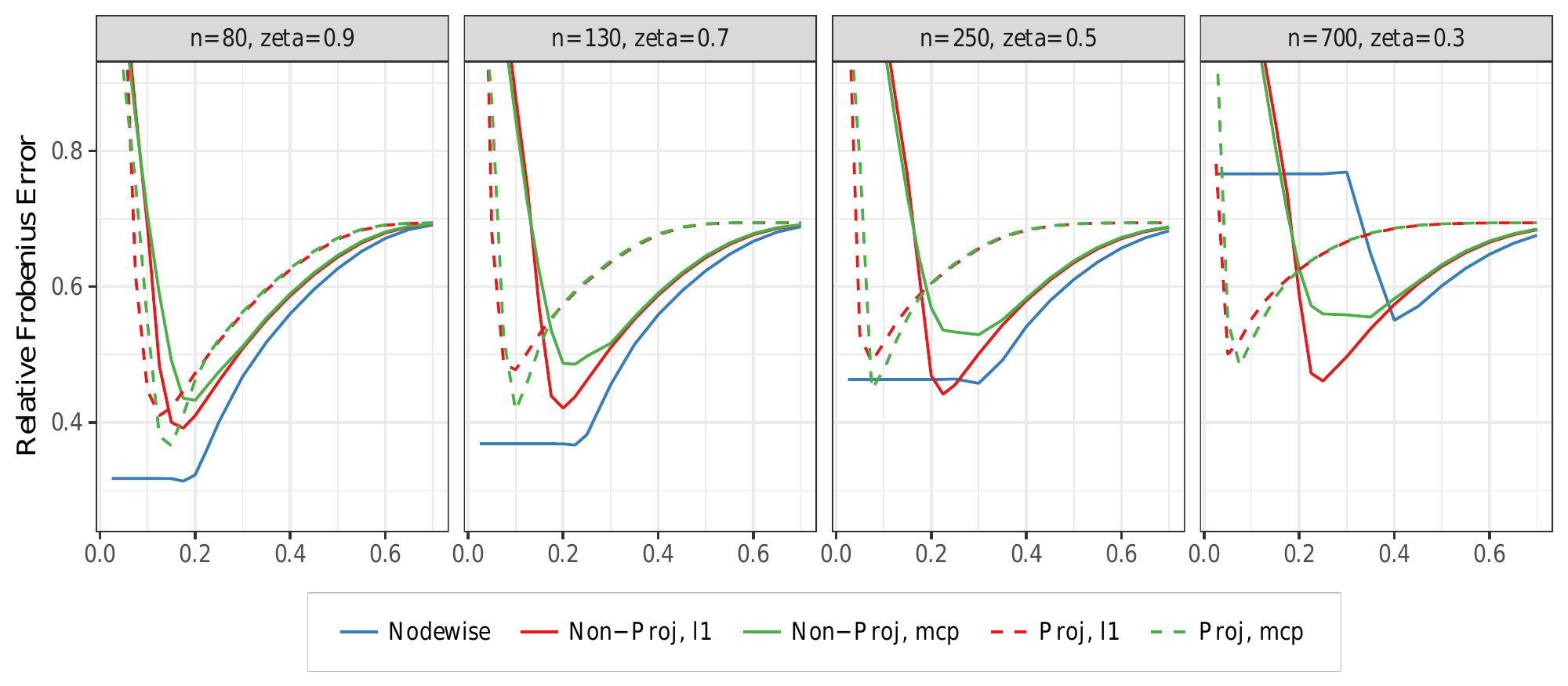}

\caption{The performance of the various estimators for the missing data model in terms of relative Frobenius error ($\norm{\hat{\Theta} - \Theta^*}_F/\norm{\Theta^*}_F$) and model selection as measured by FPR + FNR. We use an $\mt{AR}(0.6)$ covariance and set $m=1200$. Settings are chosen so that the effective sample size ($n \zeta^2$) is roughly equivalent. The MCP penalty uses $a = 2.5$. We set $R$ to be $1.5$ times the oracle value for each method and set $\rho = 24$. Our convergence criteria is $\norm{\Theta^{k+1} - \Theta^k}_F / \norm{\Theta^k}_F < 5\mathrm{e}{-5}$.}
\label{fig:missdat_ar_projvsnon}
\end{figure*}

\textbf{Method comparisons.} Figure~\ref{fig:missdat_ar_projvsnon} demonstrates the statistical performance along the full regularization path. Across the panels from left to right, the sampling rate decreases and therefore the magnitude of the most negative eigenvalue increases (see Table~\ref{tab:input_spectrums}).

In terms of Frobenius error, both projected methods and the nonprojected estimator with the $\ell_1$ penalty get slightly worse across panels, but the nodewise regression and the nonprojected MCP estimator react much more negatively to more indefinite input. The nodewise regression in particular goes from being among the best to among the worst estimators as the sampling rate decreases.

Comparing the projected and nonprojected curves in Figure~\ref{fig:missdat_ar_projvsnon}, we see that the optimal value of $\lambda$, as well as the range of optimal values, shrinks for the projected method as the sampling rate decreases. This pattern is consistently repeated across models and scenarios, likely because the $\ell_\infty$ projection is shrinking the off-diagonal entries of the input matrix. We find that the nonprojected graphical Lasso performs slightly better than the projected version when used with the $\ell_1$ penalty, likely due to the information lost in this shrinkage.

Figure~\ref{fig:missdat_ar_projvsnon} also shows how these methods perform in terms of model selection. We can see that the nonconvex penalties perform essentially identically to their $\ell_1$ penalized counterparts. In particular, the degradation of the nonprojected MCP estimator in terms of norm error does not seem to affect its model selection performance. The nodewise regression, however, still demonstrates this pattern, as its model selection performance degrades across the panels. For scenarios with more missing data, the nonprojected estimators seem to be easier to tune, maintaining a wider range of $\lambda$ values where they perform near-optimally. In Section~\ref{supp:simulation_supp} of the supplement we perform similar experiments in a variety of different noise and model settings.

\begin{figure}[tbh] \centering

\begin{subfigure}[t]{0.78\linewidth} \centering
\includegraphics[width=1\linewidth]{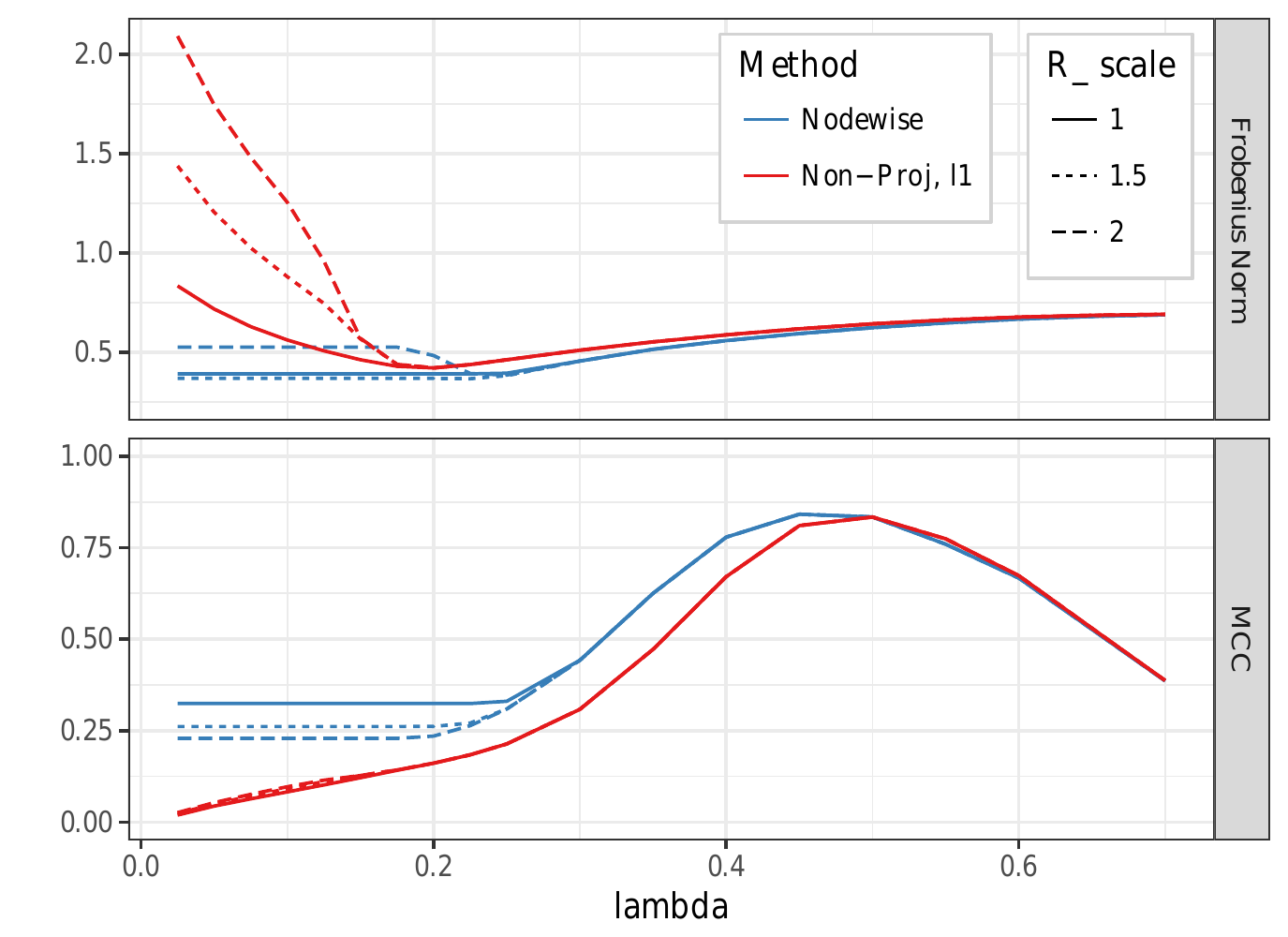}
\end{subfigure}
\\
\begin{subfigure}[t]{0.78\linewidth} \centering
\includegraphics[width=1\linewidth]{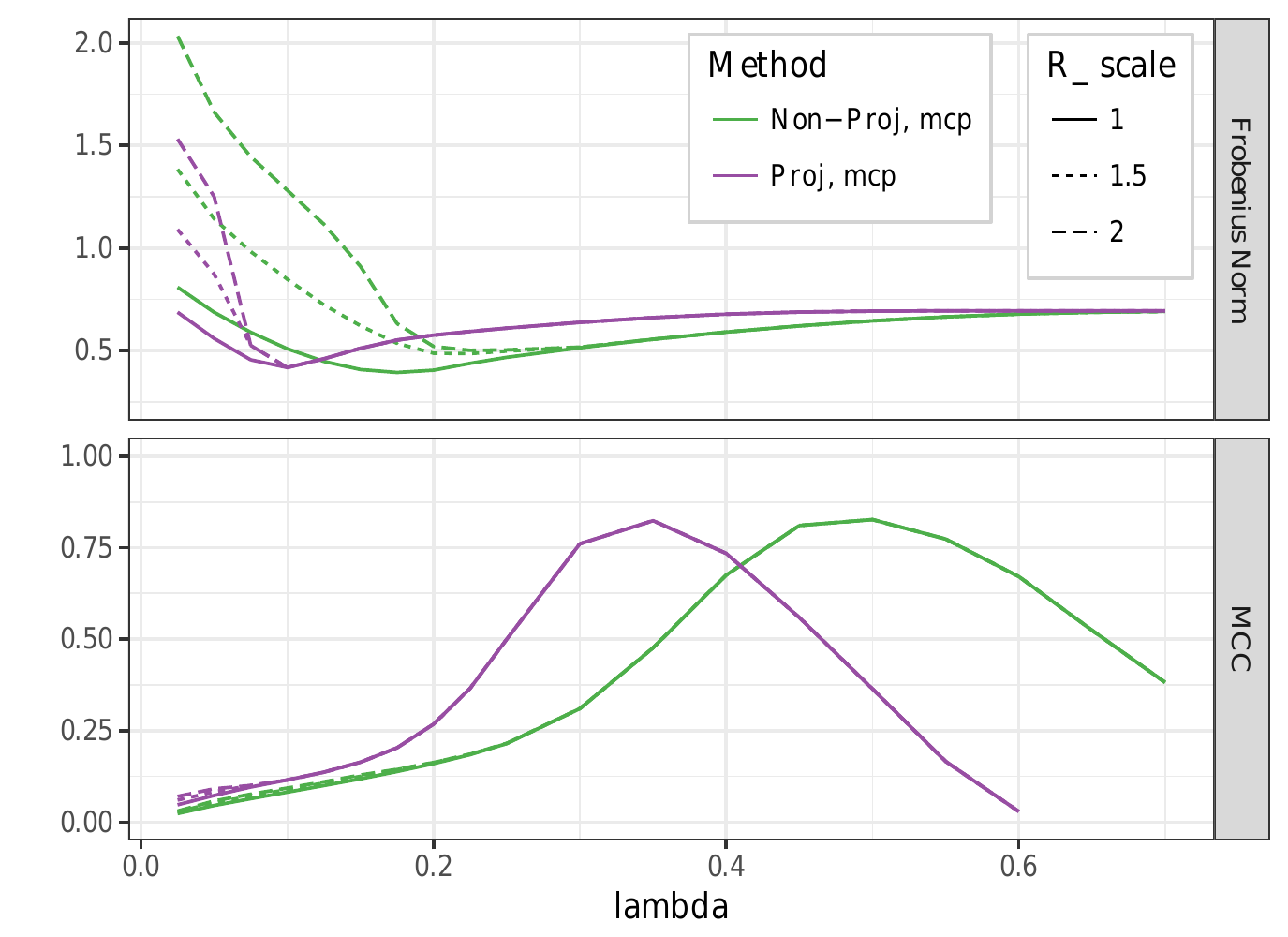}
\end{subfigure}

\caption{The performance of missing data estimators over different choices of $R$. The non-nodewise estimators set $R = \mt{R\_scale} \times \norm{A}_2$, while each node's regression in the nodewise estimator sets $R$ to be R\_scale times that node's oracle $\ell_1$ value. We use an $\mt{AR}(0.6)$ covariance, set $m=1200$, $n=130$, and choose a sampling rate of $\zeta = 0.7$. The MCP penalty is chosen with $a = 2.5$.}
\label{fig:kronsum_ar_varyR}
\end{figure}

\textbf{Sensitivity to \boldmath$R$.} Figure~\ref{fig:kronsum_ar_varyR} demonstrates the sensitivity of the nonprojected estimators to the choice of $R$, the size of the side constraint. We can see that all these methods are sensitive to the choice of $R$ for small values of $\lambda$ in terms of norm error. None of the methods are sensitive in terms of model selection.

The nonprojected graphical Lasso with MCP is the most sensitive to $R$ and is also sensitive for larger choices of $\lambda$, which is important since it never reaches its oracle minimum norm errors when $R$ is chosen to be larger than the oracle. The nonprojected graphical Lasso with $\ell_1$ and the projected graphical Lasso with MCP both still achieve the same best-case performance when $R$ is misspecified, though tuning $\lambda$ becomes more difficult.

The nodewise regression results are also plotted here. Here $R$ is the $\ell_1$ side constraint level in \eqref{eq:nodewise_lassoreg}. For smaller values of $\lambda$ the nodewise estimator levels off, corresponding to when the side constraint becomes active over the penalty. Different values of $R$ change when this occurs and, if $R$ is chosen large enough, do not significantly affect ideal performance. Note that these use a stronger oracle that knows each column-wise $\ell_1$ norm, but do show that this method can be improved with careful tuning.

\section{Senate voting analysis}
\label{sec:data_analysis}
Based on the missing data model from Section~\ref{sec:data_models}, we estimate the conditional dependence graph among senators using the ADMM algorithm from Section~\ref{sec:ADMM}. The dataset includes voting records from the United States Senate during the 112th Congress (2011-2013).
We drop senators who serve partial terms and unanimous votes, resulting in a dataset of voting records for 99 senators over 426 votes. Appendix~\ref{sec:data_analysis_supp} contains further details regarding data processing and the methods used as well as additional analysis.

Missing values in this data correspond to votes that are missed by senators and consist of roughly 2.6\% of total votes.
Note that only 109 of the votes are fully observed, so some type of correction or imputation should be used instead of omitting rows.

A major story at this time was the rise of the tea party movement in the Republican party. Across the US government tea party challengers rose to prominence. Though it was not an official party, politicians associated with the tea party movement tended to be more conservative and less likely to compromise than establishment Republicans, leading to a particularly politically polarized period of government.

\begin{figure}[tbp] \centering
\begin{subfigure}[t]{0.92\linewidth} \centering
\includegraphics[width=1\linewidth]{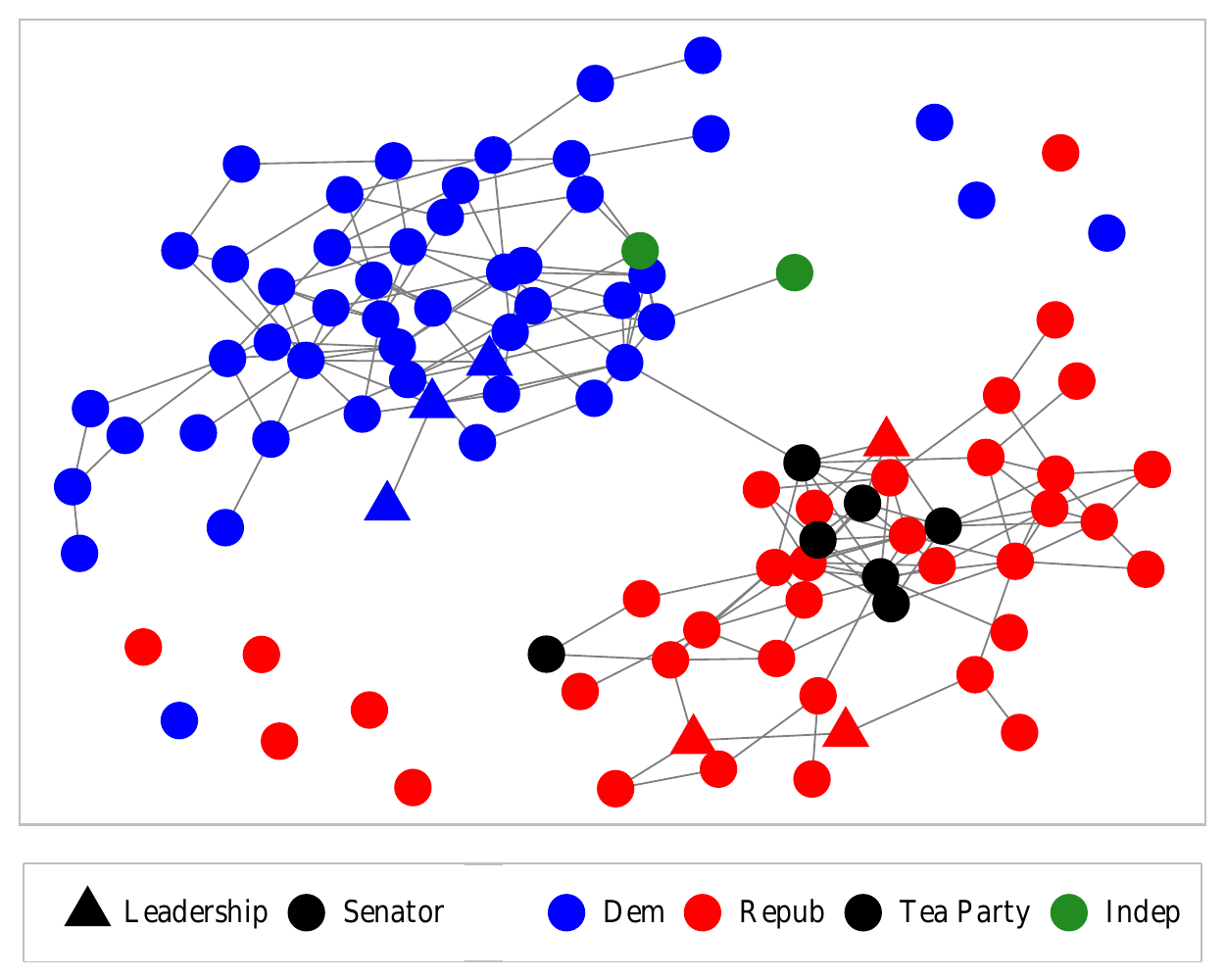}
\caption{$\lambda=0.21$}
\label{fig:senate_graphs_1}
\end{subfigure}
\begin{subfigure}[t]{0.92\linewidth} \centering
\includegraphics[width=1\linewidth]{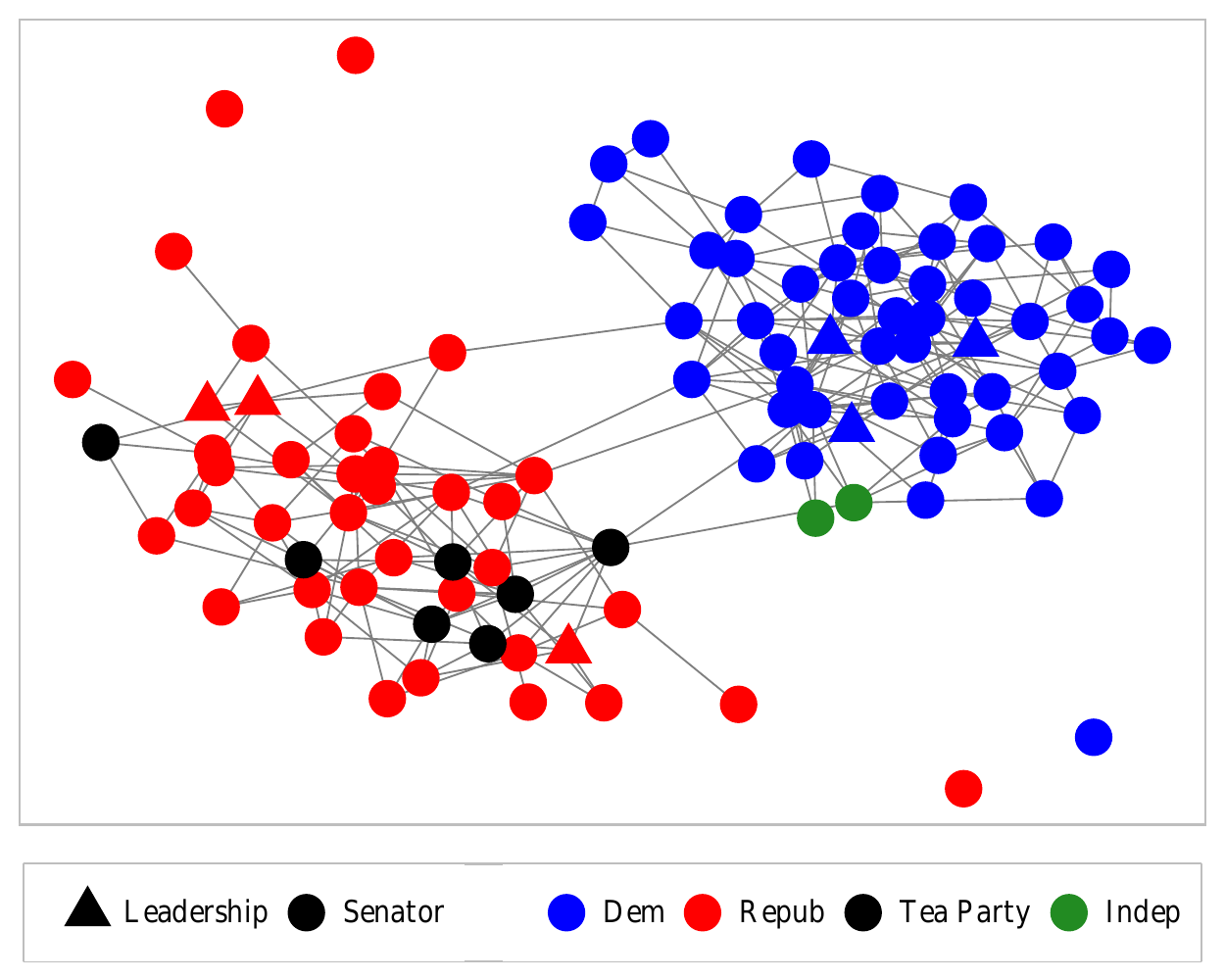}
\caption{$\lambda=0.15$}
\label{fig:senate_graphs_2}
\end{subfigure}
\begin{subfigure}[t]{0.92\linewidth} \centering
\includegraphics[width=1\linewidth]{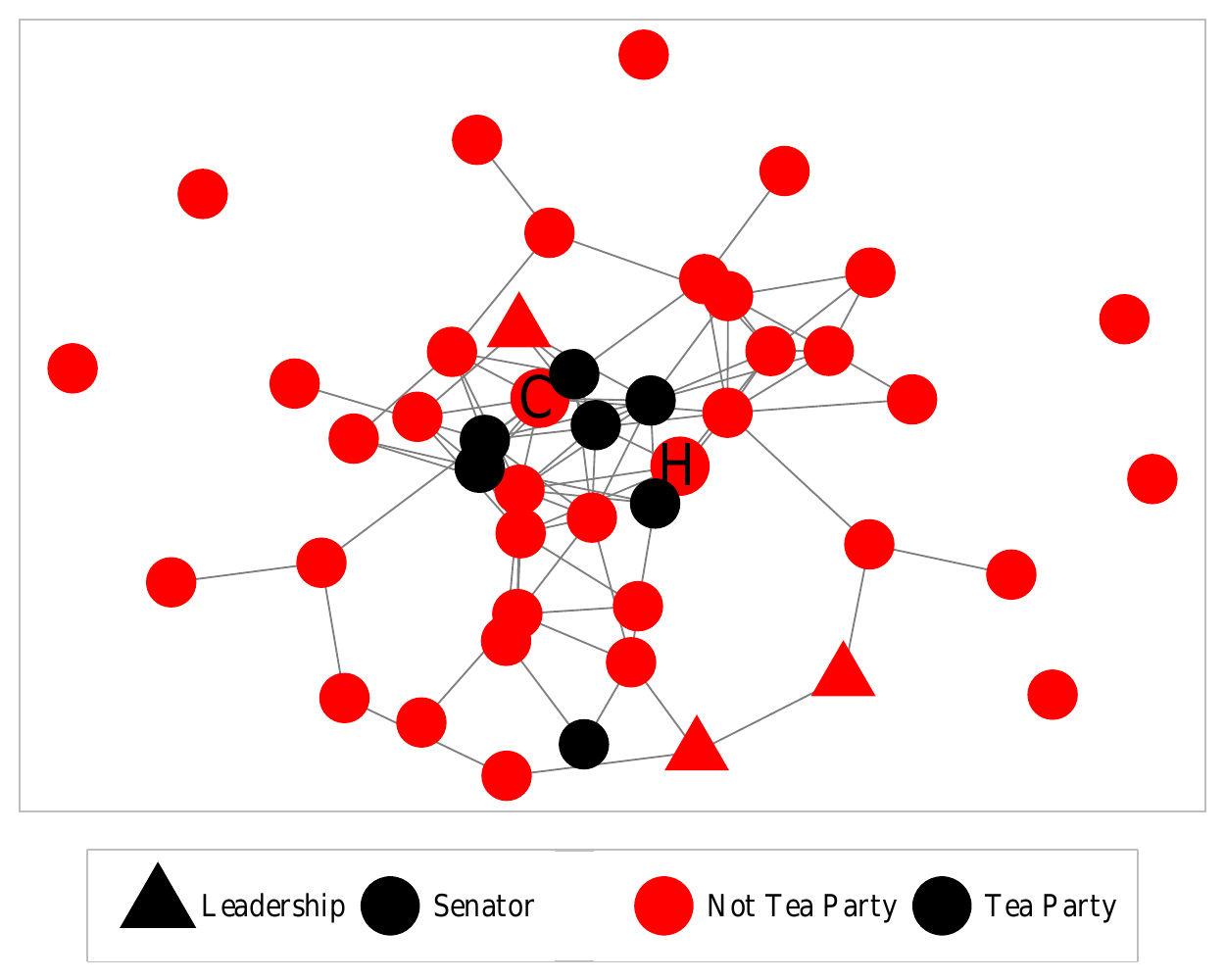}
\caption{$\lambda=0.21$, Republican subgraph}
\label{fig:senate_graphs_r}
\end{subfigure}

\caption{Graphs among senators estimated on Senate voting records from the 112th US Congress using an $\ell_1$ penalty with penalty $\lambda$ as indicated. We set $R=10$ and the ADMM algorithm was run with $\rho=10$. After estimation, the precision matrix is thresholded at 0.04 for the top and bottom panel and 0.055 for the middle one.}
\label{fig:senate_graphs}
\end{figure}

Figure~\ref{fig:senate_graphs} plots the estimated graph among senators. As expected the distinction between Republicans and Democrats is stark. Both independent senators caucus with the Democrats, so as expected they are part of the Democratic component of the graph.

We identify senators who were present at the inaugural meeting of the unofficial Senate Tea Party Caucus as well as those elected in 2010 with significant tea party support.\footnote{The marked tea party senators are Marco Rubio, Mike Lee, Jerry Moran, Jim DeMint, Rand Paul, Ron Johnson, and Pat Toomey.} These senators are colored in black, and we can see that within the Republican party they are clustered together.

In Figure~\ref{fig:senate_graphs_1} we can see that the sole connection between parties runs through the tea party (Rand Paul) and Jeff Merkley, a Democratic senator. This may be surprising, as Rand Paul is one of the most conservative senators and Merkley one of the most liberal. Paul is, however,  regarded as a relatively libertarian conservative. So though he is extremely conservative in some dimensions, he may share liberal views with Merkeley on others.

Figure~\ref{fig:senate_graphs_2} plots the same graph estimated at a lower penalization level. The Republicans who have cross-party connections include some of both the most conservative (Paul) and the most moderate (Thad Cochran, Lisa Murkowski).\footnote{Here we are measuring ideology by NOMINATE, a standard method in political science for assessing a representative's position on the political spectrum \citep{poole2005spatial}. See Appendix~\ref{sec:data_analysis_supp} for more details.} On the Democratic side the cross-connected senators also include both the most liberal (Sanders, Merkley, Tom Udall) and relatively moderate (Claire McCaskill). As expected, moderates are among those most connected opposing party, but this shows that the most extreme members of a party can also be linked to the opposing party. Appendix~\ref{sec:data_analysis_supp} discusses these cross-party links in more detail.

Figure~\ref{fig:senate_graphs_r} shows the Republican subgraph from Figure~\ref{fig:senate_graphs_1}. Here we can identify other senators who are closely associated with the tea party. In particular, two nodes near the tea party cluster are marked `H' and `C,' corresponding to Senators Orrin Hatch and Tom Coburn. Both have been linked to the tea party in the media, either as candidates supported by it or as being supportive of the movement.

It is also of interest that one marked senator is not clustered with the others, Jerry Moran. This suggests that he is not as closely connected to the tea party movement as the others we have identified.

\section{Summary and discussion}

In this paper, we study the estimation of sparse precision matrices from noisy and missing data. To close an existing algorithmic gap, we propose an ADMM algorithm that allows for fast optimization of the side-constrained graphical Lasso, which is needed to implement the graphical Lasso with either indefinite input and/or nonconvex penalties. We investigate its convergence properties and compare its performance with other methods that handle the indefinite sample covariance matrices that arise with dirty data.

We find that methods with nonconvex penalties are quite sensitive to the indefiniteness of the input covariance estimate, and are particularly sensitive to the magnitude of its negative eigenvalues.
They may have better existing theoretical guarantees, but in practice we find that with nontrivial missingness or noise they perform worst than or, at best, recover the performance of their $\ell_1$-normalized counterparts. The nonconvex methods can outperform the $\ell_1$-penalized ones when there is a small amount of missingness or noise, but in these cases we often find the nodewise estimator to perform best.

In difficult settings with significant noise or missingness, the most robust and efficient method seems to be using the graphical Lasso with nonprojected input and an $\ell_1$ penalty. As the application becomes easier -- with more observations or less missing data -- the nodewise estimator becomes more competitive, just as it is understood to be with fully observed data.

The projected graphical Lasso estimator with an $\ell_1$ penalty seems to be slightly worse than its nonprojected counterpart. Projection does, however, allow for the use of nonconvex penalties in more difficult settings without the large degradation in performance we have observed. This may be desired in some scenarios when the nonzero off-diagonal precision matrix entries are expected to be large.

Finally, we also use this new algorithm to estimate conditional dependence graphs among US senators using voting records data. We identify several interesting patterns in these graphs, especially regarding the rise of the tea party movement and cross-party connections between senators.

\FloatBarrier

\vfill
\pagebreak

\subsection*{Acknowledgements}

The research is supported in part by the NSF under grants DMS-1316731 and DMS-1830247.

\FloatBarrier

\nocite{belloni2017linearconic,rosenbaumtsybakov10matrixuncert,rosenbaumtsybakov13impmatrixuncert,datta2017cocolasso,parketal17_kroneckersum}
\bibliographystyle{apa}
\bibliography{gems,admm}

\vfill
\pagebreak
\appendix

\section{Auxiliary Results}

\subsection{Nonconvex penalties}
\label{sec:nonconvex_penalties}

The nonconvex penalties we will focus on are the SCAD and MCP functions, introduced in \citet{fan2001variable} and \citet{zhang10mcp}, respectively. Following \citet{lohwainwright15}, we make the following assumptions regarding the (univariate) penalty function $g_\lambda\colon \R \to \R$.
\begin{enumerate}[(i), itemsep=2pt, topsep=5pt]
\item
  $g_\lambda(0) = 0$ and $g_\lambda(t) = g_\lambda(-t)$.
\item
  $g_\lambda(w)$ is nondecreasing for $w >= 0$.
\item
  $g_\lambda(w)/w$ is nonincreasing for $w>0$.
\item
  $g_\lambda'(w)$ exists for all $w \neq 0$ and $\lim_{w \to 0^+} g_\lambda'(w) = \lambda$.
\item
  $g_\lambda$ is weakly convex, i.e. there exists $\mu>0$ such that $g_\lambda(w) + (\mu/2) w^2$ is convex.
\end{enumerate}
Note that \citet{lohwainwright17} show stronger model selection results under the following additional assumption.
\begin{enumerate}[(i), itemsep=2pt, topsep=5pt]
\setcounter{enumi}{5}
\item \label{assump:penalty6}
  There exists a constant $\gamma < \infty$ such that $g_\lambda'(w) = 0$ for all $w > \gamma\lambda$.
\end{enumerate}
This excludes the $\ell_1$ penalty, but is satisfied by the nonconvex penalties we consider.

The SCAD penalty takes the form
\begin{equation} \label{eq:penalty_scad}
g_\lambda(w)
  = \begin{cases}
  \lambda \abs{w} & \mt{if } \abs{w} \leq \lambda \\
  -\frac{w^2 - 2 a \lambda \abs{w} + \lambda^2}{2 (a - 1)} & \mt{if } \lambda < \abs{w} \leq a \lambda \\
  \frac{(a+1) \lambda^2}{2} & \mt{if } a \lambda < \abs{w}
  \end{cases}
\end{equation}
for some parameter $a>2$. Note that this penalty is weakly convex with constant $\mu=1/(a-1)$.

The MCP penalty has the form
\begin{equation} \label{eq:penalty_mcp}
g_\lambda(w) = \mt{sign}(w) \lambda \int_0^{\abs{w}} \left( 1 - \frac{z}{\lambda a} \right)_+ \dif z
\end{equation}
for some parameter $a>0$. This penalty is weakly convex with $\mu=1/a$.

\subsection{Derivation of Algorithm~\ref{alg:admm_indefinite}}
\label{subsec:ADMM_derivation}

Recall that we can rewrite the objective as
\begin{equation*}
f(\Theta) = \tr(\Gammahat \Theta) - \log \det (\Theta) + g_\lambda(\Theta) + \I_{\mathcal{X}_R} (\Theta)
\end{equation*}
where $\mathcal{X}_R = \{\Theta : \Theta \succeq 0, \norm{\Theta}_2 \leq R\}$ and
$\I_\mathcal{X} (\Theta) = 0 \mt{ if } \Theta \in \mathcal{X} \mt{ and } \infty \mt{ otherwise}$.

We then introduce an auxiliary optimization variable $V \in \R^{m \times m}$ and reformulate the problem as
\begin{equation*}
\begin{aligned}
  \hat{\Theta} & =\argmax_{\Theta, V \in \R^{m \times m}}
    \Big\{ \tr(\Gammahat \Theta) - \log \det (\Theta) \\
    & \qquad \qquad \qquad \qquad + \I_\mathcal{X_R}(\Theta)  + g_\lambda(V) \Big \} \; \mt{  s.t. } \Theta = V
\end{aligned}
\end{equation*}
For a penalty parameter $\rho>0$ and Lagrange multiplier $\Lambda \in \R^{m \times m}$, we consider the augmented Lagrangian
\begin{equation} \label{eq:glasso_obj_auglag}
\begin{aligned}
  \mathcal{L}_\rho(\Theta, V, \Lambda)
    & = \tr(\Gammahat \Theta) - \log \det (\Theta) + \I_{\mathcal{X}_R}(\Theta) \\
    & + g_\lambda(V) + \frac{\rho}{2} \norm{\Theta - V}_F^2 + \inner{\Lambda, \Theta - V}
\end{aligned}
\end{equation}

The ADMM algorithm is then, given current iterates $\Theta^k$, $V^k$, and $\Lambda^k$,
\begin{gather}
\begin{aligned}
V^{k+1}
  = \argmin_{V \in \R^{m \times m}} \Big\{ &g_\lambda(V) + \frac{\rho}{2} \norm{\Theta^k - V}_F^2 \\
  &\quad+ \inner{\Lambda^k, \Theta^k - V} \Big\}
  \label{eq:admm_v}
\end{aligned} \\
\begin{aligned}
\Theta^{k+1}
  = \argmin_{\Theta \in \R^{m \times m}} \Big\{
    & - \log \det \Theta + \tr(\Gammahat \Theta) \\
    &\quad+ \I_{\mathcal{X}_R}(\Theta) + \frac{\rho}{2} \norm{\Theta - V^{k+1}}_F^2 \\
    &\quad+ \inner{\Lambda^k, \Theta - V^{k+1}}
  \Big\}
\end{aligned}
  \label{eq:admm_theta} \\
\Lambda^{k+1} = \Lambda^k + \rho(\Theta^{k+1} - V^{k+1})  \label{eq:admm_lambda}
\end{gather}

Considering the $V$-subproblem, we can show that the minimization problem in \eqref{eq:admm_v} is equivalent to
\begin{equation*}
V^{k+1} = \argmin_{V \in \R^{m \times m}} \left\{ \frac{1}{\rho} g_\lambda(V) + \frac{1}{2} \norm*{V - \frac{\rho \Theta^k + \Lambda^k}{\rho}}_F^2 \right\}.
\end{equation*}
Which is a prox operator of $g_\lambda / \rho$. Let $W = \frac{\rho \Theta^k + \Lambda^k}{\rho}$ and $\nu = 1/\rho$. If $g_\lambda$ is the $\ell_1$ penalty then these updates simply soft-threshold the elements of $W$ at level $\lambda/\rho$. For SCAD, these updates have the element-wise form

\small
\begin{equation}
\mt{Prox}_{g_{\lambda} / \rho}(w)
  = \begin{cases}
  0 & \mt{if } \abs{w} \leq \nu \lambda \\
  w - \mt{sign}(w) \nu \lambda & \mt{if } \nu \lambda \leq \abs{w} \leq (\nu+1) \lambda \\
  \frac{w - \mt{sign}(w) \frac{a \nu \lambda}{a - 1}}{1 - \frac{\nu}{a-1}} & \mt{if } (\nu+1)\lambda \leq \abs{w} \leq a \lambda \\
  w & \mt{if } a \lambda \leq \abs{w}
  \end{cases}
\end{equation}
\normalsize
While for MCP the updates are
\begin{equation}
\mt{Prox}_{g_{\lambda} / \rho}(w)
  = \begin{cases}
  0 & \mt{if } \abs{w} \leq \nu \lambda \\
  \frac{w - \mt{sign}(w) \nu \lambda}{1 - \nu/a} & \mt{if } \nu \lambda \leq \abs{w} \leq a \lambda \\
  w & \mt{if } a \lambda \leq \abs{w}
  \end{cases}
\end{equation}
See \citet{lohwainwright15} for the derivations of these updates.

For the $\Theta$-subproblem, we can similarly show that \eqref{eq:admm_theta} is equivalent to
\begin{equation}
\begin{aligned}
  \Theta^{k+1}  = \argmin_{\Theta \in \R^{m \times m}} \Big\{ & - \log \det \Theta + \I_{\mathcal{X}_R}(\Theta) \\
  & + \frac{\rho}{2} \norm*{\Theta - \frac{\rho V^{k+1} - \Gammahat - \Lambda^k}{\rho}}_F^2 \Big\}
\end{aligned}
\end{equation}
For any matrix $A$ with corresponding eigendecomposition $A = R M R^T$ let us define the operator
\begin{equation} \label{eq:logdet_prox_ind}
\begin{aligned}
& T_\rho(A)  = T_\rho(U M U^T)\\
  & \quad = \argmin_{\Theta} \left\{ - \log \det \Theta + \I_{\mathcal{X}_R}(\Theta) + \frac{\rho}{2} \norm{\Theta - A}_F^2 \right\} \\
  & \quad = U \tilde{D} U^T \\
  & \qquad \mt{ where } \tilde{D}_{ii} = \min \left\{ \frac{M_{ii} + (M_{ii}^2 + \frac{4}{\rho})^{1/2}}{2}, R \right\}
\end{aligned}
\end{equation}
whose solution is derived in Section~\ref{sec:logdet_prox_ind}. Then the solution to \eqref{eq:admm_theta} is
$T_\rho ( (\rho V^{k+1} - \Gammahat - \Lambda^k)/\rho )$.

Using these results, the algorithm in \eqref{eq:admm_v}-\eqref{eq:admm_lambda} becomes
\begin{equation} \label{eq:admm_solved}
\begin{aligned}
V^{k+1}
  &= \mt{Prox}_{g_{\lambda} / \rho} \left( \frac{\rho \Theta^k + \Lambda^k}{\rho} \right) \\
\Theta^{k+1}
  &= T_\rho \left( \frac{\rho V^{k+1} - \Gammahat - \Lambda^k}{\rho} \right) \\
\Lambda^{k+1} &= \Lambda^k + \rho(\Theta^{k+1} - V^{k+1})
\end{aligned}
\end{equation}

\subsection{Solution of \texorpdfstring{$T_\rho$}{T}}
\label{sec:logdet_prox_ind}

Recall that in \eqref{eq:logdet_prox_ind} we define
\begin{equation*}
\begin{aligned}
  & T_\rho(A) \\
    & \quad = \argmin_{\Theta} \left\{ - \log \det \Theta + \I_{\mathcal{X}_R}(\Theta) + \frac{\rho}{2} \norm{\Theta - A}_F^2 \right\}\\
\end{aligned}
\end{equation*}
Let $\Theta = W D W^T$ and $A = U M U^T$ be the eigen-decompositions of the optimization variable and $A$. Then, similar to the derivation in \citet{guozhang17l1condnumprec}, we can rewrite this problem as
\begin{align*}
T_\rho(A)
  &= \argmin_{\Theta \in \R^{m \times m}}
    -\log \det \Theta + \frac{\rho}{2} \tr(\Theta \Theta) \\
  & \qquad \qquad \qquad - \rho \tr(\Theta A) + \I_{\mathcal{X}_R}(\Theta) \\
  &= \argmin_{\Theta = W D W^T}
    -\log \det D + \frac{\rho}{2} \tr(DD) \\
  & \qquad \qquad \qquad - \rho \tr(WDW^T UMU^T) + \I_{\mathcal{X}_R}(D) \\
  &= \argmin_{\Theta = W D W^T, W=U}
    -\log \det D + \frac{\rho}{2} \tr(DD) \\
  & \qquad \qquad \qquad - \rho \tr(DM) + \I_{\mathcal{X}_R}(D)
\end{align*}
The final line is since, if we denote $O(m)$ to be the set of $m \times m$ orthonormal matrices,
\begin{equation*}
\begin{aligned}
  & \tr(WDW^T UMU^T)  = \tr((U^T W) D (U^T W)^T M) \\
  & \qquad \leq \sup_{Q \in O(m)} \tr(Q D Q^T M) = \tr(D M)
\end{aligned}
\end{equation*}
Which holds with equality when $W = U$. Note that the last equality here is from Theorem~14.3.2 of \citet{farrell1985multivariate}.

We therefore get that $T_\rho(A) = U \tilde{D} U^T$ where
\begin{align*}
\tilde{D}
  &= \argmin_{D \mt{ diagonal}} - \log \det D \\
  & \qquad \qquad \qquad + \frac{\rho}{2} \tr(D^2) - \rho \tr(DM) + \I_{\mathcal{X}_R}(D) \\
  &= \argmin_{D \mt{ diagonal}} \sum_{i=1}^m \Big( - \log D_{ii} + \frac{\rho}{2} D_{ii}^2 - \rho D_{ii} M_{ii} \\
  & \qquad \qquad \qquad \qquad \qquad + \I(0 \leq D_{ii} \leq R) \Big)
\end{align*}
We can see that this is separable by element. Let
\begin{equation*}
q(d; M_{ii}) = - \log d + \frac{\rho}{2} d^2 - \rho d M_{ii}
\end{equation*}
So $\tilde{D}_{ii} = \argmin_{d} q(d; M_{ii}) + \I(0 \leq d \leq R)$. Ignoring the constraints in the indicator function for now, we can set the derivative of $q$ equal to zero to get that
\begin{equation*}
0 = - \frac{1}{d} + \rho d - \rho M_{ii}
\implies 0 = d^2 - M_{ii} d - \frac{1}{\rho}
\end{equation*}
Which we can solve with the quadratic formula to show that $q(d; M_{ii})$ has a unique minimizer over $d > 0$ at
\begin{equation*}
\argmin_{d} q(d; M_{ii}) = \frac{M_{ii} + (M_{ii}^2 + \frac{4}{\rho})^{1/2}}{2}
\end{equation*}
Adding $\I(0 \leq d \leq R)$ back and noting that $q(d; M_{ii})$ is strictly convex over $d > 0$, we get that we simply need to truncate this value at $R$. Therefore we get that
\begin{equation*}
\begin{aligned}
  & T_\rho(UMU^T) = U \tilde{D} U^T  \\
  & \qquad \mt{ where } \tilde{D}_{ii} =
    \min \left\{ \frac{M_{ii} + (M_{ii}^2 + \frac{4}{\rho})^{1/2}}{2}, R \right\}
\end{aligned}
\end{equation*}

\subsection{Proof of Proposition~\ref{cor:admm_conv_convex}}
\label{subsec:admm_conv_convex_pf}

\begin{proof}
The optimization problem \eqref{eq:glasso_obj_foradmm} is equivalently
\begin{equation} \label{eq:glasso_obj_alt}
\begin{aligned}
  \min_{\Theta, V} \ \phi(\Theta, V)
    & = \min_{\Theta, V} \left\{ f_1(\Theta) + f_2(V) \right\} \\
    & \mt{ s.t. } \ A \mt{vec}(V) + B \mt{vec}(\Theta) = 0
\end{aligned}
\end{equation}
where
$f_1(\Theta) = \tr(\Gammahat \Theta) - \log \det (\Theta) + \I_{\mathcal{X}_R}(V)$,
$f_2(V) = g_\lambda(V)$, $A = -I_{m^2}$, and $B = I_{m^2}$.

\citet{boydetal10admm} show that if $f_1$ and $f_2$ are proper convex functions and if \eqref{eq:glasso_obj_alt} is solveable then ADMM converges in terms of the objective value $\phi(\Theta^t, V^t) \to \phi^*$  and dual variable $\Lambda^t \to \Lambda^*$. \citet[Proposition~4.2]{bertsekas1989paralleldistcomp} and \citet{mota2011proof} show that if in addition $A$ and $B$ have full column rank then we get convergence of the primal iterates $\Theta^t \to \Theta^*$ and $V^t \to V^*$, where $(\Theta^*, V^*)$ is the solution to \eqref{eq:glasso_obj_alt}.
\end{proof}

\subsection{Proof of Proposition~\ref{thm:admm_stationary}}
\label{subsec:admm_stationary_pf}

Before we prove Proposition~\ref{thm:admm_stationary}, we first define directional derivatives and stationary points.
\begin{definition**}
The \emph{directional derivative} of a lower semi-continuous function $h$ at $\Theta$ in the direction $\Delta$ is
\begin{equation*}
h'(\Theta; \Delta) = \lim_{t \searrow 0} \frac{h(\Theta + t \Delta) - h(\Theta)}{t}.
\end{equation*}
Note that we allow $h'(\Theta; \Delta) = +\infty$.
We say that $\Theta$ is a \emph{stationary point} of $h$ if it satisfies the first-order necessary conditions to be a local extrema, i.e.
\begin{equation*}
h'(\Theta; \Delta) \geq 0 \mt{ for all directions } \Delta \in \R^{m \times m}
\end{equation*}
\end{definition**}
Note that this coincides with the definition of stationary point used in \citet{lohwainwright17}, though they use slightly different notation. Also note that $h'(\Theta; \Delta) = \inner{\nabla h(\Theta), \Delta}$ when $h$ is continuously differentiable.

\begin{proof}  
From the first-order necessary conditions of the subproblems \eqref{eq:admm_v}-\eqref{eq:admm_theta}, we get that, for all $\Delta \in \R^{m \times m}$,
\begin{equation} \label{eq:admm_foc}
\begin{aligned}
&0 \leq g_\lambda'(V^{k+1}; \Delta) - \inner{\rho (\Theta^k - V^{k+1}) + \Lambda^k, \Delta} \\
&0 \leq \inner{\Gammahat - (\Theta^{k+1})^{-1} + \rho (\Theta^{k+1} - V^{k+1}) \\
& \qquad \qquad \qquad \qquad + \Lambda^k, \Delta} + \I'_{\mathcal{X}_R}(\Theta^{k+1}; \Delta)
\end{aligned}
\end{equation}
And recall that
\begin{equation} \label{eq:admm_lambda_recall}
\Lambda^{k+1} = \Lambda^k + \rho (\Theta^{k+1} - V^{k+1}).
\end{equation}
We can rewrite \eqref{eq:admm_foc}-\eqref{eq:admm_lambda_recall} as
\begin{align}
&g_\lambda'(V^{k+1}; \Delta) \geq \inner{\rho (\Theta^k - \Theta^{k+1}) + \Lambda^{k+1}, \Delta} \label{eq:admm_foc2_1} \\
&0 \leq \inner{\Gammahat - (\Theta^{k+1})^{-1} + \Lambda^{k+1}, \Delta} + \I'_{\mathcal{X}_R}(\Theta^{k+1}; \Delta)
 \label{eq:admm_foc2_2} \\
& \frac{1}{\rho} (\Lambda^{k+1} - \Lambda^k) = \Theta^{k+1} - V^{k+1}. \label{eq:admm_foc2_3}
\end{align}

Now consider a fixed point $(\Theta^*, V^*, \Lambda^*)$ and consider \eqref{eq:admm_foc2_1}-\eqref{eq:admm_foc2_3} evaluated at this limit point. From \eqref{eq:admm_foc2_3} we get that $\Theta^* = V^*$. This combined with \eqref{eq:admm_foc2_1} gives us that, for all $\Delta \in \R^{m \times m}$,
\begin{equation*}
g_\lambda'(\Theta^*; \Delta) \geq \inner{\Lambda^*, \Delta}
\end{equation*}
Finally, \eqref{eq:admm_foc2_2} gives us that
\begin{equation*}
0 \leq \inner{\Gammahat - (\Theta^*)^{-1} + \Lambda^*, \Delta} + \I'_{\mathcal{X}_R}(\Theta^*; \Delta)
\end{equation*}
Using the above and recalling the objective $f$ as defined in \eqref{eq:glasso_obj_foradmm}, we get that, for all $\Delta \in \R^{m \times m}$,
\begin{align*}
0 &\leq \inner{\Gammahat - (\Theta^*)^{-1}, \Delta} + \inner{\Lambda^*, \Delta} + \I'_{\mathcal{X}_R}(\Theta^*; \Delta) \\
  &\leq \inner{\Gammahat - (\Theta^*)^{-1}, \Delta} + g_\lambda'(\Theta^*; \Delta) + \I'_{\mathcal{X}_R}(\Theta^*; \Delta) \\
  &= f'(\Theta^*; \Delta)
\end{align*}
So $\Theta^*$ is a stationary point of $f$ by definition.
\end{proof}

\subsection{Comparison to Guo and Zhang (2017)}
\label{sec:guozhang_comp}

\citet{guozhang17l1condnumprec} study the problem of condition number-constrained precision matrix estimation, where they consider the estimator
\begin{equation}
\hat{\Theta} = \argmin_{\Theta \succ 0, \mt{cond}(\Theta) \leq \kappa} - \log \det \Theta + \tr(\Gammahat \Theta) + \lambda \norm{\Theta}_{1, \mt{off}}
\end{equation}
Note that this is quite similar to the estimators we consider in \eqref{eq:glasso_opt_problem_side}, as they simply replace the maximum eigenvalue constraint with a constraint on the ratio of the maximum to minimum eigenvalues.

However, they do not study the application of their estimator to cases with indefinite input or its performance in noisy and missing data situations. In particular, constraining the condition number does not necessarily guarantee that the graphical Lasso objective \eqref{eq:glasso_opt_problem} will be lower bounded, especially when using nonconvex penalties.

As a simple example, consider the case with an input matrix and iterates
\begin{equation*}
\Gammahat = \begin{pmatrix}
  1 & 0 \\ 0 & -0.2
  \end{pmatrix}
\qquad \Theta^t = t\begin{pmatrix}
  0.1 & 0 \\ 0 & 1
  \end{pmatrix}
\end{equation*}
In this case the objective is
\begin{equation*}
f(\Theta^t)
  = \tr(\Gammahat \Theta^t) - \log \det \Theta^t
  = -0.1 \times t - \log (0.1 \times t)
\end{equation*}
which is unbounded below as $t$ grows even though the condition numbers of the iterates are constant.

More generally, whenever $\Gammahat \in \R^{m \times m}$ has eigenvalues $\sigma_1, \dots, \sigma_m$, where $\sigma_1 \geq \dots \geq \sigma_{m_1} \geq 0$ and $0 > \sigma_{m_1+1} \geq \dots \geq \sigma_m$. Denote $S_1 = \sum_{i=1}^{m_1} \sigma_i$ and $S_2 = \sum_{i=m_1+1}^m - \sigma_i$. Let $VDV^T = \Gammahat$ be the eigendecomposition of the covariance estimate. Then for some condition number bound $\kappa$, we can consider iterates of the form $\Theta^t = t V M V^T$, where $M$ is a diagonal matrix with entries
\begin{equation*}
M_{ii} = \begin{cases}
  1 &\mt{ if } i \leq m_1 \\
  \kappa &\mt{ if } i > m_1
  \end{cases}
\end{equation*}
Which we note has a condition number of $\kappa$. Then we can see that the objective becomes
\begin{align*}
f(\Theta^t)
  &= t \tr(VDV^T VMV^T) \\
  & \qquad \qquad - (m-m_1) \log(\kappa) + g_\lambda(t VMV^T) \\
  &= t (S_1 - \kappa S_2) \\
  & \qquad \qquad- (m-m_1) \log(\kappa) + g_\lambda(t VMV^T)
\end{align*}
So if $\kappa > S_1 / S_2$ then this objective is still unbounded below.

Using a spectral norm bound $\norm{\Theta}_2 \leq R$ as the side constraint with a indefinite input guarantees a lower bound on the graphical Lasso objective regardless of the choice of $R$ and is therefore a more natural side constraint to use.

\section{ADMM for general side constraints}
\label{sec:admm_genside}

In this section we develop an ADMM algorithm for general side constraints, i.e. the following variant of \eqref{eq:glasso_opt_problem_side}.\footnote{Note that we switch the notation of the side constraint function from $\rho$ to $h$ to avoid confusion with the ADMM penalty parameter $\rho$.}
\begin{equation*}
\hat{\Theta}
  \in \argmin_{\Theta \succeq 0, h(\Theta) \leq R} \tr(\Gammahat \Theta) - \log \det (\Theta) + g_\lambda(\Theta).
\end{equation*}

This algorithm has the same convergence guarantees as Algorithm~\ref{alg:admm_indefinite}, but in practice we find that Algorithm~\ref{alg:admm_indefinite} converges faster and more consistently when the spectral norm side constraint is used.

\subsection{Derivation}
\label{subsec:ADMM_genside_derivation}

We first rewrite the objective as
\begin{equation} \label{eq:glasso_obj_foradmm_genside}
f(\Theta) = \tr(\Gammahat \Theta) - \log \det (\Theta) + g_\lambda(\Theta) + \I_{\mathcal{X}_{h, R}} (\Theta)
\end{equation}
where $\mathcal{X}_{h, R} = \{\Theta : \Theta \succeq 0, h(\Theta) \leq R\}$ and
\begin{equation*}
\I_\mathcal{X} (\Theta)
  = \begin{cases}
  0 & \mt{if } \Theta \in \mathcal{X} \\
  \infty & \mt{otherwise.}
  \end{cases}
\end{equation*}
We can then introduce auxiliary optimization variables $V_1, V_2 \in \R^{m \times m}$ and reformulate the optimization problem as
\begin{align*}
\begin{aligned}
  \hat{\Theta} = &\argmax_{\Theta, V_1, V_2}
    \Big\{ \tr(\Gammahat \Theta) - \log \det (\Theta) \\
    & \qquad \qquad \qquad + g_\lambda(V_1) + \I_{\mathcal{X}_{h, R}}(V_2) \Big\} \\
    & \qquad \qquad \st \Theta = V_1 = V_2
\end{aligned}
\end{align*}

For a penalty parameter $\rho>0$ and Lagrange multiplier matrices $\Lambda_1, \Lambda_2 \in \R^{m \times m}$, we consider the augmented Lagrangian of this problem
\begin{equation} \label{eq:glasso_obj_auglag_genside}
\begin{aligned}
& \mathcal{L}_\rho(\Theta, V_1, V_2, \Lambda_1, \Lambda_2) \\
  &\quad = \tr(\Gammahat \Theta) - \log \det (\Theta) + g_\lambda(V_1) + \I_{\mathcal{X}_{h, R}}(V_2)  \\
  &\qquad+ \frac{\rho}{2} \norm{\Theta - V_1}_F^2 +  \frac{\rho}{2} \norm{\Theta - V_2}_F^2 + \inner{\Lambda_1, \Theta - V_1} \\
  & \qquad + \inner{\Lambda_2, \Theta - V_2}
\end{aligned}
\end{equation}
The ADMM algorithm is then, given current iterates $\Theta^k$, $V_1^k$, $V_2^k$, $\Lambda_1^k$, and $\Lambda_2^k$,
\begin{gather}
\begin{aligned}
  V_1^{k+1}
    & = \argmin_{V_1 \in \R^{m \times m}} \Big\{ g_\lambda(V_1)  + \frac{\rho}{2} \norm{\Theta^k - V_1}_F^2 \\
    & \qquad \qquad \qquad \qquad \qquad + \inner{\Lambda_1^k, \Theta^k - V_1}\Big\}
\end{aligned}
  \label{eq:admm_genside_v1} \\
\begin{aligned}
  V_2^{k+1}
    & = \argmin_{V_2 \in \R^{m \times m}} \Big\{ \I_{\mathcal{X}_{h, R}}(V_2)  + \frac{\rho}{2} \norm{\Theta^k - V_2}_F^2 \\
      & \qquad \qquad \qquad \qquad \qquad + \inner{\Lambda_2^k, \Theta^k - V_2}\Big\}
\end{aligned}
  \label{eq:admm_genside_v2} \\
\begin{aligned}
\Theta^{k+1}
  = \argmin_{\Theta \in \R^{m \times m}}
    &\Big\{ - \log \det \Theta + \tr(\Gammahat \Theta)  \\
    &\hspace{-8pt} + \frac{\rho}{2} \norm{\Theta - V_1^{k+1}}_F^2 + \frac{\rho}{2} \norm{\Theta - V_2^{k+1}}_F^2\\
    &\hspace{-8pt} + \inner{\Lambda_1^k, \Theta - V_1^{k+1}} + \inner{\Lambda_2^k, \Theta - V_2^{k+1}} \Big\}
\end{aligned}
  \label{eq:admm_genside_theta} \\
\Lambda_1^{k+1} = \Lambda_1^k + \rho(\Theta^{k+1} - V_1^{k+1})  \label{eq:admm_genside_lambda1}\\
\Lambda_2^{k+1} = \Lambda_2^k + \rho(\Theta^{k+1} - V_2^{k+1}) \label{eq:admm_genside_lambda2}
\end{gather}

Considering the $V_1$-subproblem, we can show that the minimization problem in \eqref{eq:admm_genside_v1} is equivalent to
\begin{equation*}
V_1^{k+1} = \argmin_{V_1 \in \R^{m \times m}} \left\{ \frac{1}{\rho} g_\lambda(V_1) + \frac{1}{2} \norm*{V_1 - \frac{\rho \Theta^k + \Lambda_1^k}{\rho}}_F^2 \right\}.
\end{equation*}
Which is a prox operator of $g_\lambda / \rho$. These have the same form as described in Section~\ref{subsec:ADMM_derivation}.

For the $V_2$-subproblem, we similarly see that \eqref{eq:admm_genside_v2} is equivalent to
\begin{equation*}
V_2^{k+1} = \argmin_{V_2 \in \R^{m \times m}} \left\{ \I_{\mathcal{X}_{h, R}}(V_2) + \frac{1}{2} \norm*{V_2 - \frac{\rho \Theta^k + \Lambda_2^k}{\rho}}_F^2 \right\}.
\end{equation*}
which is equivalent to the projection operator
\begin{equation}
\mt{Proj}_{\mathcal{X}_{h, R}} \left( \frac{\rho \Theta^k + \Lambda_2^k}{\rho} \right)
  = \min_{V_2 \in \mathcal{X}_{h, R}} \norm*{V_2 - \frac{\rho \Theta^k + \Lambda_2^k}{\rho}}_F^2
\end{equation}
Note that if directly projecting onto $\mathcal{X}_{h, R}$ does not have an closed-form solution, we can perform this step using Dykstra's alternating projection algorithm.

Finally, for the $\Theta$-subproblem, we can again show that \eqref{eq:admm_genside_theta} is equivalent to

\begin{equation}
\begin{aligned}
  \Theta = & \argmin_{\Theta \in \R^{m \times m}}  \Big\{ - \log \det \Theta \\
  & \quad  + \rho \norm*{\Theta - \frac{\rho V_1^{k+1} + \rho V_2^{k+1} - \Gammahat - \Lambda_1^k - \Lambda_2^k}{2 \rho}}_F^2 \Big\}
\end{aligned}
\end{equation}

Let us define the operator
\begin{equation} \label{eq:logdet_prox}
\begin{aligned}
\tilde{T}_\rho(A)
  &= \argmin_{\Theta} \left\{ - \log \det \Theta + \rho \norm{\Theta - A}_F^2 \right\} \\
  &= \frac{1}{2} (A + (A^2 + (2/\rho) I)^{1/2})
\end{aligned}
\end{equation}
whose solution is derived in Section~\ref{sec:logdet_prox_ind} if we set $R=\infty$. Then the solution to \eqref{eq:admm_genside_theta} is
$\tilde{T}_\rho ( (\rho V_1^{k+1} + \rho V_2^{k+1} - \Gammahat - \Lambda_1^k - \Lambda_2^k)/(2 \rho) )$.

Using these results, the algorithm in \eqref{eq:admm_genside_v1}-\eqref{eq:admm_genside_lambda2} becomes
\begin{equation} \label{eq:admm_genside_solved}
\begin{aligned}
V_1^{k+1}
  &= \mt{Prox}_{g_{\lambda} / \rho} \left( \frac{\rho \Theta^k + \Lambda_1^k}{\rho} \right) \\
V_2^{k+1}
  &= \mt{Proj}_{\mathcal{X}_{h, R}} \left( \frac{\rho \Theta^k + \Lambda_2^k}{\rho} \right) \\
\Theta^{k+1}
  &= \tilde{T}_\rho \left( \frac{\rho V_1^{k+1} + \rho V_2^{k+1} - \Gammahat - \Lambda_1^k - \Lambda_2^k}{2 \rho} \right) \\
\Lambda_1^{k+1} &= \Lambda_1^k + \rho(\Theta^{k+1} - V_1^{k+1}) \\
\Lambda_2^{k+1} &= \Lambda_2^k + \rho(\Theta^{k+1} - V_2^{k+1})
\end{aligned}
\end{equation}

\subsection{Convergence}
\label{subsec:ADMM_genside_convergence}

Analogues to Propositions~\ref{cor:admm_conv_convex} and~\ref{thm:admm_stationary} can also be shown for this algorithm using similar methods. To do this, we first note that we can rewrite the optimization problem \eqref{eq:glasso_obj_foradmm_genside} as
\begin{equation} \label{eq:glasso_obj_alt_genside}
\begin{aligned}
\min_{\Theta, V} &\ \phi(\Theta, V)
  = \min_{\Theta, V} \left\{ f_1(\Theta) + f_2(V) \right\} \\
  \mt{s.t. } &\ A \mt{vec}(V) + B \mt{vec}(\Theta) = 0
\end{aligned}
\end{equation}
where
\begin{align*}
f_1(\Theta) &= \tr(\Gammahat \Theta) - \log \det (\Theta) \\
f_2(V) &= g_\lambda(A_1 V) + \I_{\mathcal{X}_{h, R}} (A_2 V)
\end{align*}
and
\begin{gather*}
A = - I_{2m^2}
\qquad B = \begin{pmatrix} I_{m^2} \\ I_{m^2} \end{pmatrix} \\
V = \begin{pmatrix} V_1 \\ V_2 \end{pmatrix}
\qquad A_1 = \begin{pmatrix} I_m & 0 \end{pmatrix}
\qquad A_2 = \begin{pmatrix} 0 & I_m \end{pmatrix}
\end{gather*}
This results in the following augmented Lagrangian that is equivalent to \eqref{eq:glasso_obj_auglag_genside}.
\begin{equation*}
\begin{aligned}
\mathcal{L}_\rho(\Theta, V, \Lambda)
  &= f_1(\Theta) + f_2(V) \\
  & \quad  + \frac{\rho}{2} \norm{B \Theta + A V}_F^2 + \inner{\Lambda, B \Theta + A V}
\end{aligned}
\end{equation*}
Even though we present our algorithm as a three-block ADMM in Section~\ref{subsec:ADMM_genside_derivation}, this formulation makes it clear that we are using a two-block splitting scheme where \eqref{eq:admm_genside_v1} and \eqref{eq:admm_genside_v2} are the separable subproblems of the $V$-step.

Showing similar convergence results to Propositions~\ref{cor:admm_conv_convex} and~\ref{thm:admm_stationary} can then be done using the same techniques as in Sections~\ref{subsec:admm_conv_convex_pf} and~\ref{subsec:admm_stationary_pf}.

\section{Additional simulation results}
\label{supp:simulation_supp}

\subsection{Tuning parameter selection}
\label{subsec:tuning}

Note that in practice tuning parameters must be selected for all these methods. In particular, we must tune $\lambda$ and possibly the side-constraint $R$. Note that one often has a reasonable prior for the magnitude of the spectral norm of the true precision matrix, so if that is the case a multiple of that can often be used to choose $R$. Also, as noted in Section~\ref{subsec:sim_optperf}, when using the $\ell_1$ penalty the choice of $R$ primarily affects how difficult tuning $\lambda$ will be. Though it is important to tune correctly when using nonconvex penalties, we do not recommend those methods when there is significant missing data. Therefore we will focus on tuning $\lambda$ here, though the same methods can be used to choose $R$ as well.

Two possible methods are to use cross-validation or a modified BIC criterion. Though the particular implementation of both of these will depend on the data model that is being used, as these methods can be applied to any method that generates an indefinite initial estimate of the covariance.

For the missing data case we can follow \citet{stadler2012missing}, which uses the same data model. Recall the notation in Section~\ref{sec:data_models}, where $X_{ij}$ denotes the $i$th value of variable $j$ and $U_{ij}$ tracks if that value is observed. Here, we define the observed log-likelihood of an observation $X_i$ given a precision matrix estimate $\hat{\Theta}$ as
\begin{equation*}
\ell(X_{i}, U_{i}; \hat{\Sigma})
    = \log \phi(X_{i, U_i}; \hat{\Sigma}_{U_i, U_i})
\end{equation*}
where $X_{i, U_i}$ is the vector of values that are observed for observation $i$, $\hat{\Sigma} = \hat{\Theta}^{-1}$, and $\phi$ is the multivariate normal density. The BIC criterion, which we minimize, is therefore
\begin{equation*}
\mt{BIC}(\lambda) = -2 \sum_i \ell(X_{i}, U_{i}; \hat{\Sigma}) + \log(n) \sum_{j \leq j'} \I\{\hat{\Theta}_{jj'\} \neq 0}
\end{equation*}

To cross-validate, we can divide the data into $V$ folds, where the $v$th fold contains indices $N_v$. The cross-validation score, which we maximize, is therefore
\begin{equation*}
\mt{CV}(\lambda)
    = \sum_v \sum_{i \in N_v} \ell(X_{i}, U_{i}; \hat{\Sigma}_{-v})
\end{equation*}
where $\hat{\Sigma}_{-v} = \hat{\Theta}_{-v}^{-1}$ and $\hat{\Theta}_{-v}$ is the estimate based on the sample omitting the observations in $N_v$.

\begin{figure}[tbh] \centering
\includegraphics[width=0.95\linewidth]{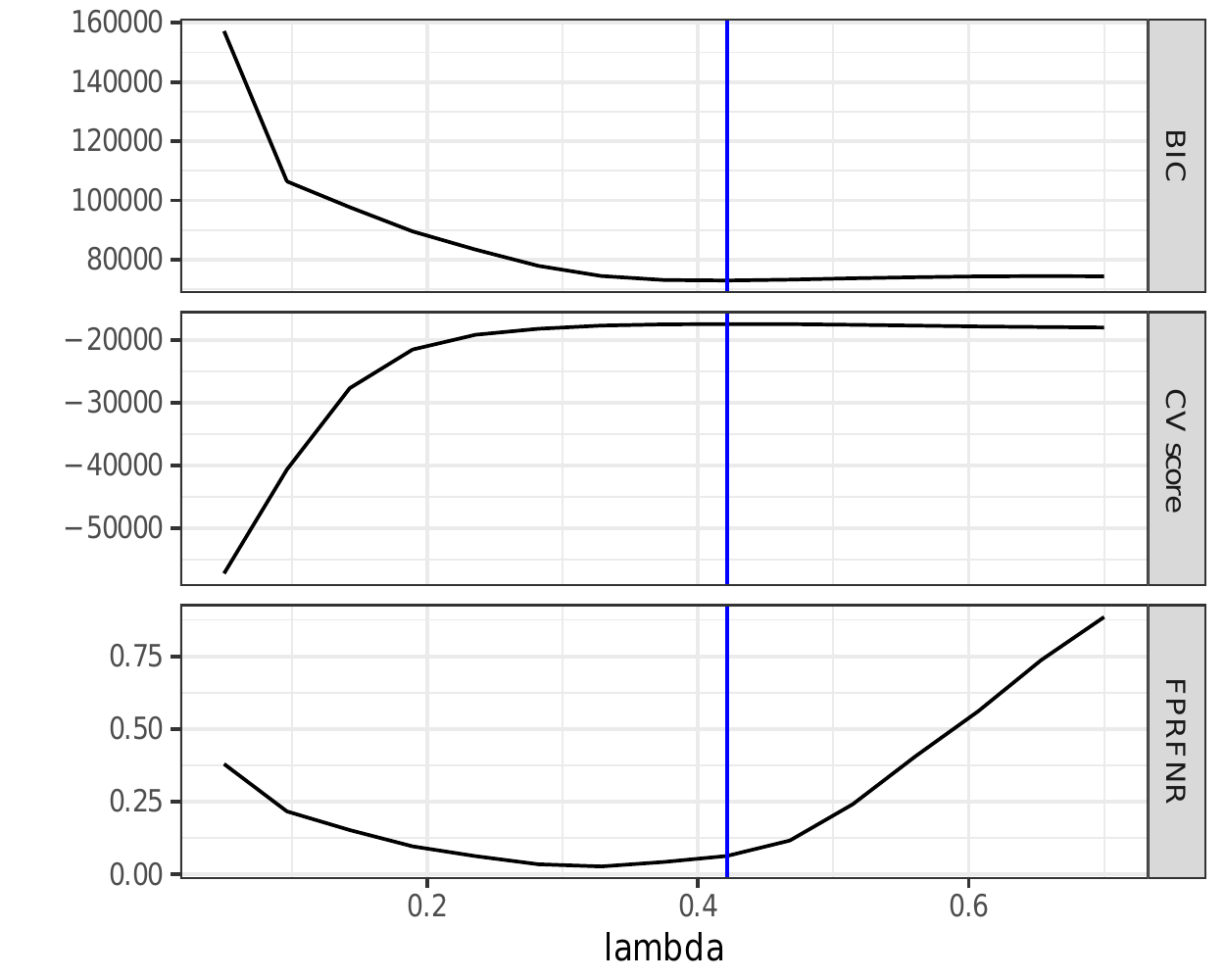}
\caption{Example parameter tuning using BIC and CV. We additionally present the FPR+FNR rate of the estimate. The vertical lines show the optimal $\lambda$ values for BIC and CV, which here happen to be identical. We set $m=400$ and $n=80$, the sampling rate to $\zeta = 0.8$, and let $A$ be from an $\mt{AR}(0.6)$ model.}
\label{fig:cv_bic}
\end{figure}

Figure~\ref{fig:cv_bic} presents an example of parameter tuning on a simulated scenario. We see that both BIC and CV select slightly higher-than-optimal levels of penalization in terms of model selection, but that selected model still achieves fairly good model selection.

\subsection{Kronecker sum (KS) model}
\label{subsec:kronsummodel}
\citet{parketal17_kroneckersum} present a graphical model with additive noise that is dependent across observations. This noise structure was first studied in the regression setting in \citet{rudelsonzhou17errinvardependent} with a Kronecker sum covariance.

Let $W_1, W_2 \in \R^{n \times m}$ be independent mean-zero subgaussian random matrices. The data matrix is generated as
$X = W_1 A^{1/2} + B^{1/2} W_2 \dist \mathcal{M}_{n, m}(0, A \oplus B)$,
where $\mathcal{M}_{n, m}$ is the matrix variate normal distribution and for covariance matrices $A \in \R^{m \times m}$ and $B \in \R^{n \times n}$. Note that $A \oplus B = A \otimes I_n + I_m \otimes B$, where $\otimes$ denotes the Kronecker product. Here $X_0 = W_1 A^{1/2}$ contains the signal and has independent rows, while $W = B^{1/2} W_2$ is the noise matrix with independent columns but dependent rows. We are interested in estimating the signal precision matrix $\Theta = A^{-1}$, which has sparse off-diagonal entries. For our simulations, we normalize $B$ so that $\tr(B) = n \tau_B$, where $\tau_B$ is a measure of the noise level. Then the initial covariance estimate for $A$ is given by
\begin{equation}
\Gammahat = \frac{1}{n} X^T X - \frac{\hat{\tr}(B)}{n} I_m
\end{equation}
as shown in \citet{rudelsonzhou17errinvardependent}.
Note that, in this model, $\Gammahat$ is guaranteed to not be positive semidefinite when $m > n$, as $X^T X$ will have zero eigenvalues.

\subsection{Covariance models}

We look into three different models from which $A$
Let $\Omega = A^{-1} = (\omega_{ij})$. We consider simulation settings using the following covariance models, which are also used in \citet{zhou14}.

\begin{itemize}
\item
  \textbf{AR1($r$)}: The covariance matrix is of the form $A = (r^{\abs{i-j}})_{ij}$.

\item
  \textbf{Star-Block (SB)}: Here the covariance matrix is block-diagonal, where each block's precision matrix corresponds to a star-structured graph with $A_{ii} = 1$. For the corresponding edge set $E$, then $A_{ij} = r$ if $(i, j) \in E$ and $A_{ij} = r^2$ otherwise.

\item
  \textbf{Erdos-Renyi random graph (ER)}: We initialize $\Omega = 0.25 I$ then randomly select $d$ edges. For each selected edge $(i, j)$, we randomly choose $w \in [0.6, 0.8]$ and update $\omega_{ij} = \omega_{ji} \to \omega_{ij} - w$ and $\omega_{ii} \to \omega_{ii} + w$, $\omega_{jj} \to \omega_{jj} + w$.
\end{itemize}

\subsection{Optimization performance}
\label{subsec:optimization_performance}

\begin{figure}[htbp] \centering
\begin{subfigure}[t]{0.48\linewidth} \centering
\includegraphics[width=\linewidth]{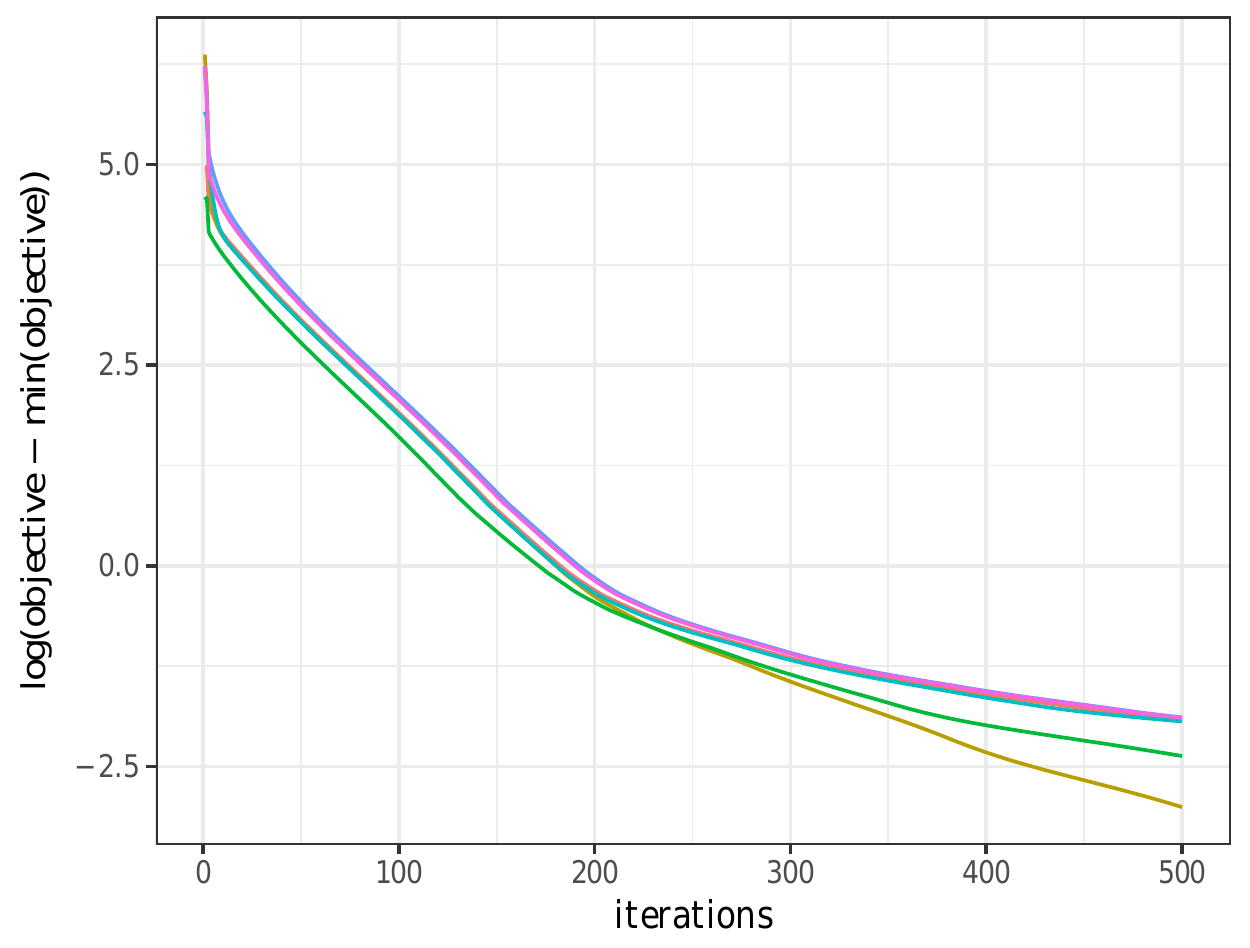}
\caption{KS, AR}
\label{subfig:admm_conv_obj_ks_ar}
\end{subfigure}
\begin{subfigure}[t]{0.48\linewidth} \centering
\includegraphics[width=\linewidth]{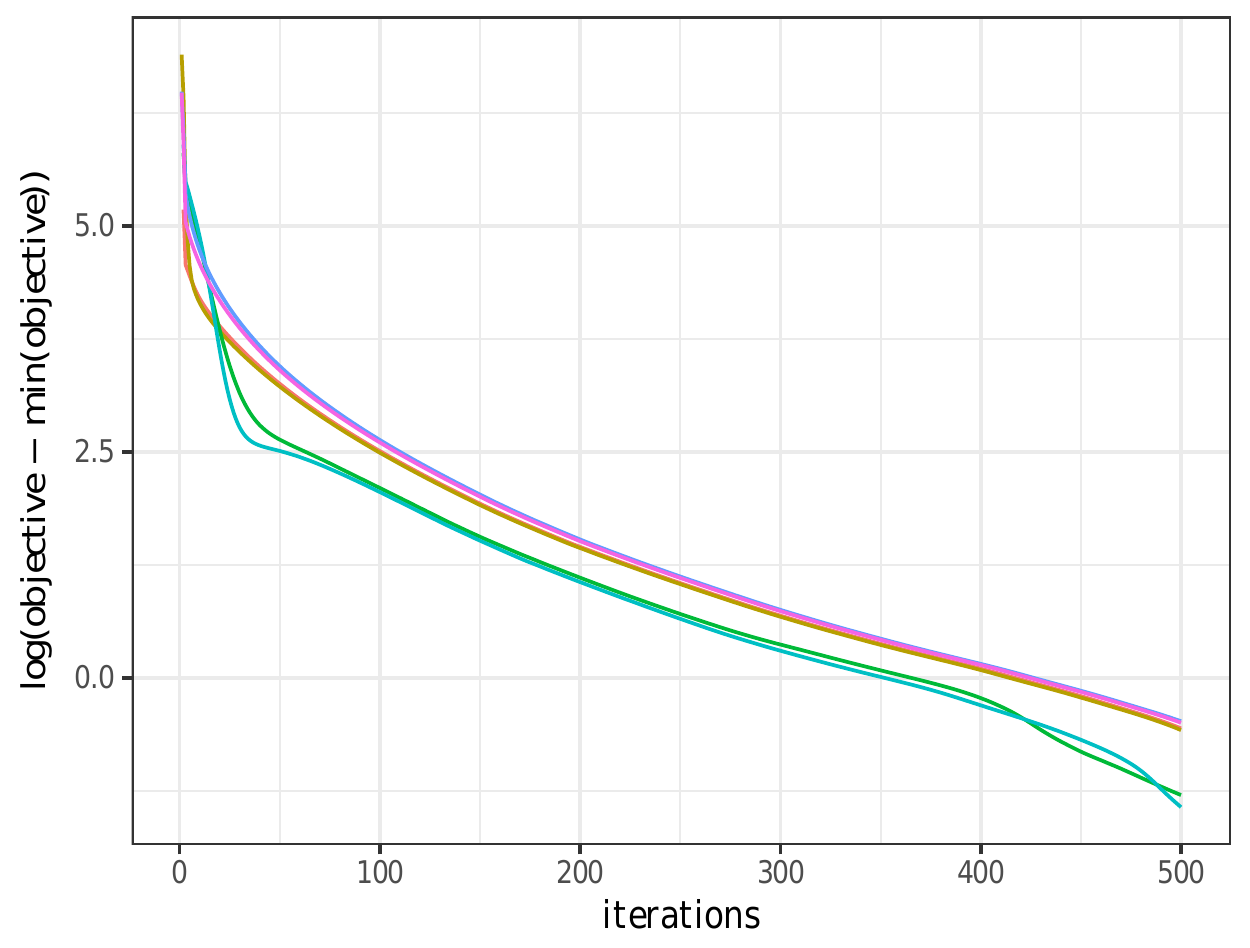}
\caption{MD, ER}
\label{subfig:admm_conv_obj_md_er}
\end{subfigure}

\caption{Convergence behavior of the ADMM algorithm for two objectives. Panel~\subref{subfig:admm_conv_obj_ks_ar} shows the optimization convergence under the Kronecker sum model with $A = \mt{AR1}(0.6)$, $B = \mt{ER}$, $m=300$, $n=140$, $\tau_B = 0.3$, and $\lambda=0.2$, while Panel~\subref{subfig:admm_conv_obj_md_er} is for the missing data model with $A = ER$, $m=400$, $n=140$, $\zeta = 0.7$, and $\lambda=0.2$. We choose $\rho = 12$ and the SCAD penalty is used with $a=2.1$.}
\label{fig:admm_conv_obj}
\end{figure}

Figure~\ref{fig:admm_conv_obj} shows the convergence behavior for several initializations in terms of objective value. Our algorithm seems to attain a linear convergence rate in terms of the objective values even with a nonconvex penalty regardless of the initialization. We find that the algorithm consistently converges well over a range of tested scenarios.

\begin{figure}[htbp] \centering
\includegraphics[width=0.9\linewidth]{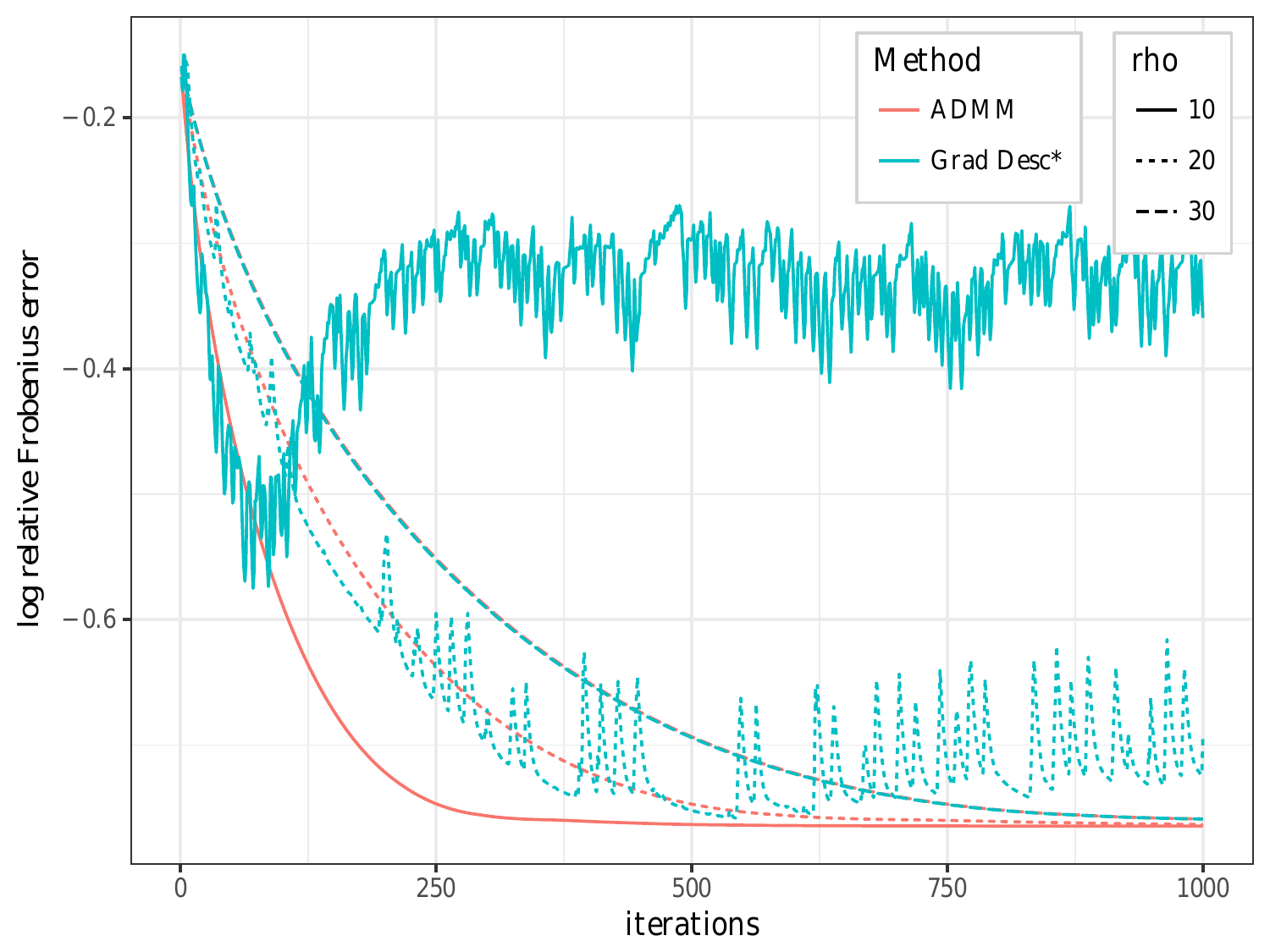}

\caption{Comparing the convergence behavior of ADMM to gradient descent. Here we use an $\mt{AR}1(0.8)$ model with $m=200$, $n=150$, $\zeta=0.6$, and use an $\ell_1$ penalty with $\lambda = 0.11$. For gradient descent, $\rho$ is the inverse of the step size. Note that since proximal gradient descent is difficult to do in this problem, this version performs the proximal gradient step without the side-constraint then projects back to the space.}
\label{fig:convcomp_admm_grad}
\end{figure}

\textbf{Comparison to gradient descent.} Figure~\ref{fig:convcomp_admm_grad} compares the optimization performance of our ADMM algorithm to gradient descent. Note that since proximal gradient descent is difficult to do in this setting, requiring an interior optimization step, we use a heuristic version similar to that suggested by \citet{agarwaletal12} that does the proximal gradient step ignoring the side-constraint then projects back to the side-constraint space. Note that since $\rho$ in ADMM is roughly equivalent to the inverse step size in gradient descent, we compare for difference values of $\rho$. These methods also take roughly the same computational time per iteration, as they are both dominated by either an eigenvalue decomposition or matrix inversion.

We can see that for large enough values of $\rho$, these methods are nearly identical. Although there is no known theoretical guarantee of convergence, it seems that this heuristic gradient descent still convergence well for small enough step sizes.

But for smaller values, i.e. larger step sizes, ADMM still performs well and obtains faster convergence rates while gradient descent is unstable and inconsistent. This combined with the convergence guarantee of ADMM leads us to recommend this algorithm.

\subsection{Penalty nonconvexity and \texorpdfstring{$R$}{R} }
\label{subsec:R_mu_cond}

Suppose $g_\lambda$ is $\mu$-weakly convex and $R \leq \sqrt{\frac{2}{\mu}}$. Then, as shown in Lemma 6 of \citet{lohwainwright17}, the overall objective function is strictly convex over the feasible set, and Proposition~\ref{thm:admm_stationary} therefore shows that any limiting point of ADMM algorithm corresponds to the unique global optimum of the objective. However, this choice of $R$ radius on the $\norm{\cdot}_2$ side constraint is quite restrictive. In particular, since we also require $R \geq \norm{\Theta^*}_2$ we therefore need to choose large values of $a$ in the SCAD or MCP penalties to make $\mu$ small enough, which means in practice we simply recover the performance of the $\ell_1$ penalized methods. Though \citet{lohwainwright17} show statistical properties for when the parameters are chosen satisfying this condition, in practice we can often do better by allowing the objective to be nonconvex even though no global optimum will exist.

\begin{figure*}[tbh] \centering
\begin{subfigure}[t]{0.7\linewidth} \centering
\includegraphics[width=\linewidth]{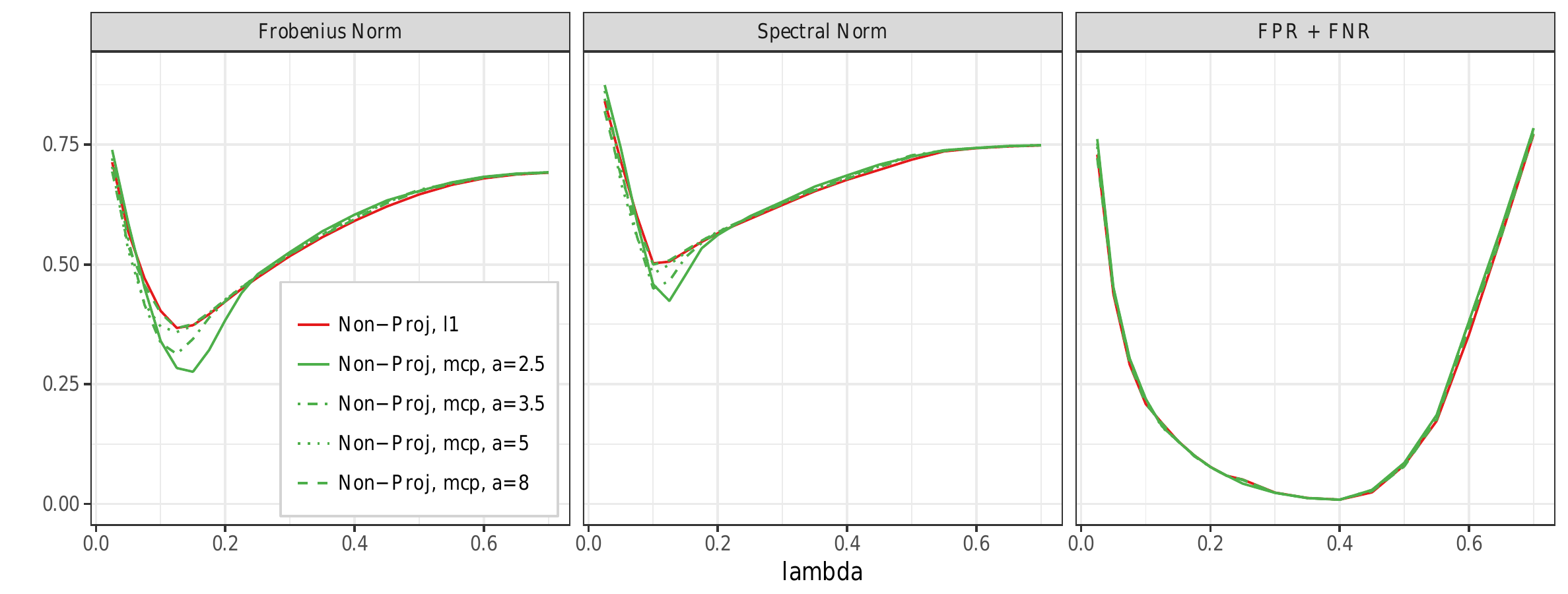}
\caption{Nonprojected estimators}
\end{subfigure}
\par\bigskip
\begin{subfigure}[t]{0.7\linewidth} \centering
\includegraphics[width=\linewidth]{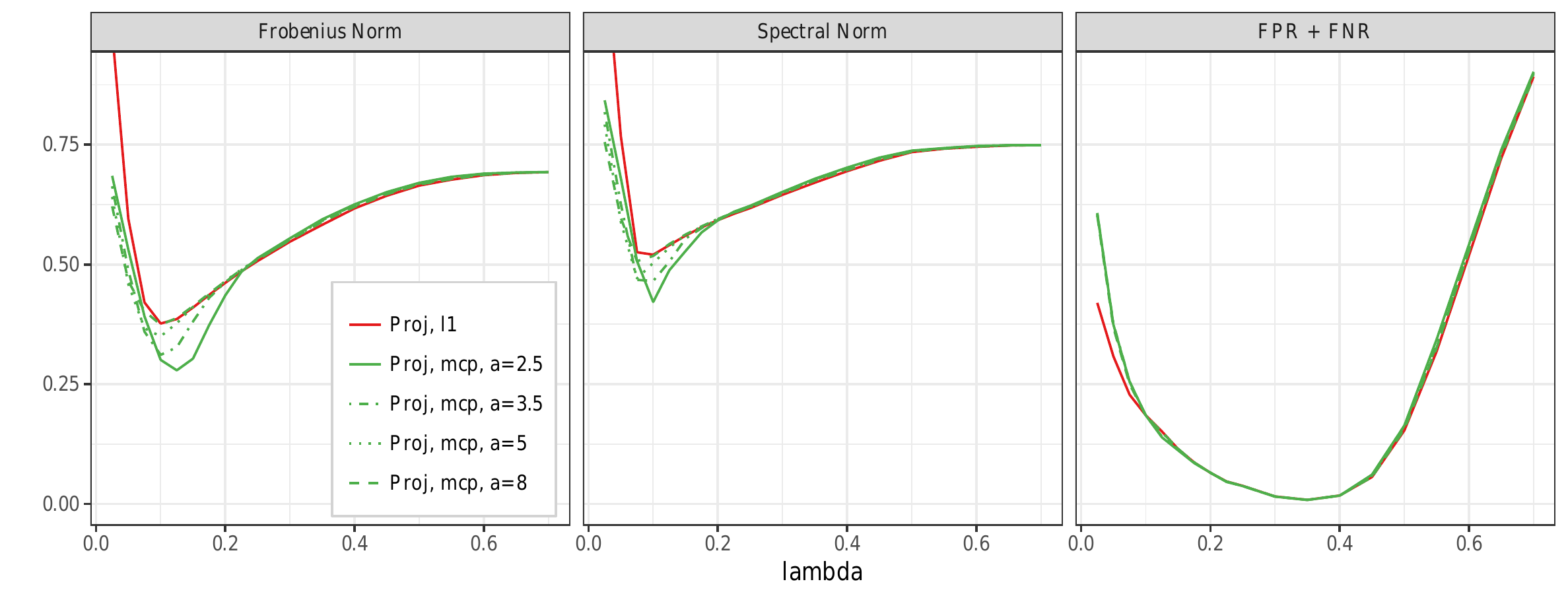}
\caption{Projected estimators}
\end{subfigure}

\caption{Comparing the performance of the graphical Lasso estimators as $a$ (and therefore the weak convexity constant $\mu$) is changed. Here we present the results using the MCP penalty, so $\mu = 1/a$. We set $R$ to be the oracle Note that $a=8$ is the only value of $a$ that satisfies the $R \leq \sqrt{2/\mu}$ condition from \citet{lohwainwright17}. Data is from a missing data model with $A=\mt{AR1}(0.6)$, $m=400$, $n=80$, and $\zeta=0.9$.}
\label{fig:rcond}
\end{figure*}

Once we relax this condition ($R > \sqrt{2/\mu}$), the objective becomes nonconvex, and Proposition~\ref{thm:admm_stationary} simply shows that any limiting point of our ADMM algorithm will be a stationary point of the objective. In our simulations, we generally set $\mu$ and $R$ such that this condition is violated, and yet we show that our algorithm still results in good estimators. In fact, Figure~\ref{fig:rcond} demonstrates how, in practice, choosing $\mu$ such that this condition is met tends to eliminate the advantages that nonconvex penalties provide. Here the choice of $a=8$ is the only one that satisfies the condition, and this choice has identical performance as the $\ell_1$ penalty. Using a smaller value of $a$ violates this condition but allows the estimator to take advantage of the unbiasedness of the penalty, resulting in better performance in this setting.

Note that for both of these cases, our ADMM algorithm provides a new feasible method of implementing estimation of this type of side-constrained graphical Lasso objective. This consideration is related to tuning, where satisfying the $(R, \mu)$ condition allows the support recovery without incoherence statistical results of \citet{lohwainwright17} but in practice results in suboptimal performance, as the nonconvex penalties have to be chosen such that they lose their unbiased advantage over the $\ell_1$ penalty.

\subsection{Method comparisons}
\label{supp:method_comp}

Tables~\ref{tab:methodcomp_kronsum_ar}-\ref{tab:methodcomp_missdat_er} present more detailed comparison based on the models from the Kronecker sum (KS) and the missing data (MD) models. We compared performance in terms of relative Frobenius and nuclear norm to the true precision matrix, as well as false positive rate plus false negative rate (FPRFNR). The Kronecker sum results are reported for two sample sizes and two values of the noise parameter $\tau_B$, while the missing data results are reported for two covariance models and three settings of the sample size and sampling rate $\zeta$.\footnote{Note that in the initial covariance estimator for the missing data model the effective sample size for estimating an off-diagonal element of the covariance is $n \zeta^2$; four settings are designed to keep this effective sample size roughly constant while changing the sampling rate $\zeta$. The effective sample sizes for the $n=80$, $n=130$, and $n=250$ settings are $64.8$, $63.7$, and $62.5$, respectively.}

\begin{table*}[tbh]
\centering \small
\caption{The relative norm error and $\mt{FPR}+\mt{FNR}$ performance of the Kronecker sum estimator using different methods. Here we set $A$ to be from an $\mt{AR}(0.5)$ model and choose $B$ from an Erdos-Renyi random graph. We set $m=400$ and let $\tau_B = 0.5$. Metrics are reported as the minimum value over a range of penalty parameters $\lambda$. The MCP penalty is chosen with $a = 2.5$, and we set $R = 1.5\norm{A}_2$.}
\begin{tabular}{rrcrcccc}
\toprule
 n & $\tau_B$ & method & penalty & Frobenius & Spectral & Nuclear & FPRFNR \\
\midrule
\multirow{5}[7]{*}{80} & \multirow{5}[7]{*}{0.3}
  & \multirow{2}{*}{Nonproj}
     & $\ell_1$ & 0.422 & 0.598 & 0.406 & 0.107 \\
   &&& MCP      & 0.450 & 0.613 & 0.422 & 0.106 \\ \cmidrule(l){3-4}
 && \multirow{2}{*}{Proj}
     & $\ell_1$ & 0.424 & 0.610 & 0.411 & 0.113 \\
   &&& MCP      & 0.444 & 0.616 & 0.429 & 0.111 \\ \cmidrule(l){3-4}
 && Nodewise
     & $\ell_1$ & 0.391 & 0.517 & 0.383 & 0.130 \\
\midrule
\multirow{5}[7]{*}{160} & \multirow{5}[7]{*}{0.3}
  & \multirow{2}{*}{Nonproj}
     & $\ell_1$ & 0.342 & 0.509 & 0.327 & 0.013 \\
   &&& MCP      & 0.363 & 0.518 & 0.345 & 0.013 \\ \cmidrule(l){3-4}
 && \multirow{2}{*}{Proj}
     & $\ell_1$ & 0.356 & 0.525 & 0.343 & 0.016 \\
   &&& MCP      & 0.341 & 0.493 & 0.321 & 0.015 \\ \cmidrule(l){3-4}
 && Nodewise
     & $\ell_1$ & 0.288 & 0.429 & 0.280 & 0.017 \\
\midrule
\multirow{5}[7]{*}{80} & \multirow{5}[7]{*}{0.5}
  & \multirow{2}{*}{Nonproj}
     & $\ell_1$ & 0.469 & 0.642 & 0.452 & 0.174 \\
   &&& MCP      & 0.481 & 0.659 & 0.458 & 0.177 \\ \cmidrule(l){3-4}
 && \multirow{2}{*}{Proj}
     & $\ell_1$ & 0.464 & 0.651 & 0.450 & 0.194 \\
   &&& MCP      & 0.483 & 0.658 & 0.467 & 0.197 \\ \cmidrule(l){3-4}
 && Nodewise
     & $\ell_1$ & 0.466 & 0.600 & 0.455 & 0.250 \\
\midrule
\multirow{5}[7]{*}{160} & \multirow{5}[7]{*}{0.5}
  & \multirow{2}{*}{Nonproj}
     & $\ell_1$ & 0.389 & 0.573 & 0.369 & 0.052 \\
   &&& MCP      & 0.422 & 0.596 & 0.393 & 0.054 \\ \cmidrule(l){3-4}
 && \multirow{2}{*}{Proj}
     & $\ell_1$ & 0.407 & 0.593 & 0.384 & 0.056 \\
   &&& MCP      & 0.399 & 0.587 & 0.377 & 0.055 \\ \cmidrule(l){3-4}
 && Nodewise
     & $\ell_1$ & 0.358 & 0.538 & 0.349 & 0.083 \\
\bottomrule
\end{tabular}
\label{tab:methodcomp_kronsum_ar}
\end{table*}

\begin{table*}[tbh]
\centering \footnotesize
\caption{The relative norm error and $\mt{FPR}+\mt{FNR}$ performance of the missing data estimator using different methods. Here we set $A$ to be from an $\mt{AR}(0.6)$ model and set $m=400$. Recall that $\zeta$ is the sampling rate. Metrics are reported as the minimum value over a range of penalty parameters $\lambda$. The MCP penalty is chosen with $a = 2.5$, and we set $R$ to be $1.5$ times the oracle value for each method.}
\begin{tabular}{rrrcrcccc}
\toprule
 $A$ Model & n & $\zeta$ & method & penalty & Frobenius & Spectral & Nuclear & FPRFNR \\
\midrule
\multirow{20}[23]{*}{$\mt{AR}(0.6)$}
& \multirow{5}[7]{*}{80} & \multirow{5}[7]{*}{0.9}
  & \multirow{2}{*}{Nonproj}
     & $\ell_1$ & 0.367 & 0.506 & 0.363 & 0.0089 \\
  &&&& MCP      & 0.308 & 0.533 & 0.296 & 0.0088 \\ \cmidrule(l){4-5}
&&& \multirow{2}{*}{Proj}
     & $\ell_1$ & 0.377 & 0.520 & 0.375 & 0.0085 \\
  &&&& MCP      & 0.308 & 0.527 & 0.284 & 0.0083 \\ \cmidrule(l){4-5}
&&& Nodewise
     & $\ell_1$ & 0.292 & 0.487 & 0.280 & 0.0097 \\
  \cmidrule(l){2-9}
& \multirow{5}[7]{*}{130} & \multirow{5}[7]{*}{0.7}
  & \multirow{2}{*}{Nonproj}
     & $\ell_1$ & 0.397 & 0.597 & 0.388 & 0.017 \\
  &&&& MCP      & 0.384 & 0.632 & 0.363 & 0.016 \\ \cmidrule(l){4-5}
&&& \multirow{2}{*}{Proj}
     & $\ell_1$ & 0.417 & 0.599 & 0.407 & 0.019 \\
  &&&& MCP      & 0.348 & 0.626 & 0.326 & 0.018 \\ \cmidrule(l){4-5}
&&& Nodewise
     & $\ell_1$ & 0.356 & 0.592 & 0.347 & 0.029 \\
  \cmidrule(l){2-9}
& \multirow{5}[7]{*}{250} & \multirow{5}[7]{*}{0.5}
  & \multirow{2}{*}{Nonproj}
     & $\ell_1$ & 0.420 & 0.619 & 0.403 & 0.028 \\
  &&&& MCP      & 0.457 & 0.680 & 0.436 & 0.026 \\ \cmidrule(l){4-5}
&&& \multirow{2}{*}{Proj}
     & $\ell_1$ & 0.437 & 0.626 & 0.429 & 0.031 \\
  &&&& MCP      & 0.391 & 0.600 & 0.369 & 0.032 \\ \cmidrule(l){4-5}
&&& Nodewise
     & $\ell_1$ & 0.412 & 0.632 & 0.400 & 0.078 \\
  \cmidrule(l){2-9}
& \multirow{5}[7]{*}{700} & \multirow{5}[7]{*}{0.3}
  & \multirow{2}{*}{Nonproj}
     & $\ell_1$ & 0.431 & 0.633 & 0.411 & 0.043 \\
  &&&& MCP      & 0.505 & 0.718 & 0.470 & 0.040 \\ \cmidrule(l){4-5}
&&& \multirow{2}{*}{Proj}
     & $\ell_1$ & 0.450 & 0.644 & 0.431 & 0.034 \\
  &&&& MCP      & 0.422 & 0.664 & 0.391 & 0.031 \\ \cmidrule(l){4-5}
&&& Nodewise
     & $\ell_1$ & 0.555 & 0.704 & 0.517 & 0.131 \\
\bottomrule
\end{tabular}
\label{tab:methodcomp_missdat_ar}
\end{table*}

\begin{table*}[tbh]
\centering \footnotesize
\caption{The relative norm error and $\mt{FPR}+\mt{FNR}$ performance of the missing data estimator using different methods. Here we set $A$ to be from an Erdos-Renyi random graph and set $m=400$. Recall that $\zeta$ is the sampling rate. Metrics are reported as the minimum value over a range of penalty parameters $\lambda$. The MCP penalty is chosen with $a = 2.5$, and we set $R$ to be $1.5$ times the oracle value for each method.}
\begin{tabular}{rrrcrcccc}
\toprule
 $A$ Model & n & $\zeta$ & method & penalty & Frobenius & Spectral & Nuclear & FPRFNR \\
\midrule
\multirow{20}[23]{*}{ER}
& \multirow{5}[7]{*}{80} & \multirow{5}[7]{*}{0.9}
  & \multirow{2}{*}{Nonproj}
     & $\ell_1$ & 0.398 & 0.426 & 0.369 & 0.133 \\
  &&&& MCP      & 0.379 & 0.444 & 0.355 & 0.132 \\ \cmidrule(l){4-5}
&&& \multirow{2}{*}{Proj}
     & $\ell_1$ & 0.405 & 0.420 & 0.375 & 0.129 \\
  &&&& MCP      & 0.367 & 0.383 & 0.346 & 0.126 \\ \cmidrule(l){4-5}
&&& Nodewise
     & $\ell_1$ & 0.349 & 0.357 & 0.334 & 0.160 \\
  \cmidrule(l){2-9}
& \multirow{5}[7]{*}{130} & \multirow{5}[7]{*}{0.7}
  & \multirow{2}{*}{Nonproj}
     & $\ell_1$ & 0.409 & 0.495 & 0.372 & 0.137 \\
  &&&& MCP      & 0.410 & 0.562 & 0.372 & 0.137 \\ \cmidrule(l){4-5}
&&& \multirow{2}{*}{Proj}
     & $\ell_1$ & 0.423 & 0.497 & 0.385 & 0.135 \\
  &&&& MCP      & 0.388 & 0.465 & 0.354 & 0.131 \\ \cmidrule(l){4-5}
&&& Nodewise
     & $\ell_1$ & 0.372 & 0.463 & 0.346 & 0.194 \\
  \cmidrule(l){2-9}
& \multirow{5}[7]{*}{250} & \multirow{5}[7]{*}{0.5}
  & \multirow{2}{*}{Nonproj}
     & $\ell_1$ & 0.421 & 0.556 & 0.379 & 0.163 \\
  &&&& MCP      & 0.463 & 0.680 & 0.401 & 0.170 \\ \cmidrule(l){4-5}
&&& \multirow{2}{*}{Proj}
     & $\ell_1$ & 0.437 & 0.556 & 0.394 & 0.163 \\
  &&&& MCP      & 0.406 & 0.535 & 0.364 & 0.171 \\ \cmidrule(l){4-5}
&&& Nodewise
     & $\ell_1$ & 0.431 & 0.654 & 0.376 & 0.241 \\
  \cmidrule(l){2-9}
& \multirow{5}[7]{*}{700} & \multirow{5}[7]{*}{0.3}
  & \multirow{2}{*}{Nonproj}
     & $\ell_1$ & 0.427 & 0.604 & 0.383 & 0.193 \\
  &&&& MCP      & 0.485 & 0.701 & 0.415 & 0.189 \\ \cmidrule(l){4-5}
&&& \multirow{2}{*}{Proj}
     & $\ell_1$ & 0.445 & 0.575 & 0.401 & 0.184 \\
  &&&& MCP      & 0.423 & 0.638 & 0.380 & 0.191 \\ \cmidrule(l){4-5}
&&& Nodewise
     & $\ell_1$ & 0.500 & 0.719 & 0.413 & 0.276 \\
\bottomrule
\end{tabular}
\label{tab:methodcomp_missdat_er}
\end{table*}

Comparing the projected and nonprojected methods, we see that these two methods are fairly competitive. In terms of model selection, the nonprojected methods tend to perform similarly or better than the projected methods. This improvement is particularly evident in the $n=80$ settings in Table~\ref{tab:methodcomp_kronsum_ar}. If we focus on the methods using the $\ell_1$ penalty, the nonprojected method performs at least similarly and sometimes significantly better than the projected method in terms of norm error. The lower sampling rate regime in Tables~\ref{tab:methodcomp_missdat_ar} and~\ref{tab:methodcomp_missdat_er} shows this trend as well. Overall these results suggest a small but sometimes significant advantage for the nonprojected methods, supporting the idea that the projected methods pay a cost in terms of efficiency due to the loss of information in the projection.

There is no significant difference in model selection between MCP and the $\ell_1$ penalty. In fact, the different penalties perform almost identically across scenarios regardless of the $\ell_\infty$-projection step. Intuitively, the primary benefit of nonconvex penalties is their ability to more accurately estimate large entries, which are easy for the estimators to select.

In terms of norm error, however, there are significant differences depending on the indefiniteness of the optimization problem. Table~\ref{tab:input_spectrums} reports some statistics on the eigenspectrum of the input matrix. Nonprojected methods with MCP tends to perform relatively better than its $\ell_1$ counterpart if the input matrix is close to the positive semidefinite space. Simulation results from the missing data model Tables~\ref{tab:methodcomp_missdat_ar} and~\ref{tab:methodcomp_missdat_er} further support this relationship between the most negative eigenvalue and the relative performance. Here we see how the MCP nonprojected estimator goes from being significantly better than its $\ell_1$ counterpart in terms of Frobenius error in the $\zeta=0.9$ case to significantly worse when $\zeta=0.5$. In the projected case, which projects away this indefinite issue, the MCP estimator consistently outperforms its $\ell_1$ counterpart in terms of Frobenius error.

The nonconvexity of the penalty interacts poorly with indefiniteness of the input matrix. When the $\ell_1$ penalty is used, it is better able to ``control'' the indefiniteness of the input due to its linear scaling, resulting in better norm error performance. The nonconvex penalty's inability to resolve the indefiniteness issue results in a degradation of its relative performance as the input matrix becomes more indefinite.

\begin{table*}[htbp]
\centering \small
\caption{Measures of the indefiniteness of the input matrix $\Gammahat$. $\sigma_i$ denote the eigenvalues of $\Gammahat$, while $\sigma_i^+$ denote the eigenvalues of $\Gammahat^+$ as defined in Section~\ref{sec:alternatives}. We set $m=400$. For data generated from each model, we report the most negative eigenvalue, the maimum eigenvalues of both the nonprojected and projected sample covariances, the sum of the negative eigenvalues, and the number of negative eigenvalues.}
\begin{tabular}{rccrrrrc}
\toprule
 Model & $A$ & n & $\min \sigma_i$ & $\max \sigma_i$ & $\max \sigma_i^+ $ & $\sum_{\sigma_i < 0} \sigma_i$ & $\#\{\sigma_i < 0\}$ \\
\midrule
\multirow{4}{*}{KS} & \multirow{4}{*}{$\mt{AR}(0.5)$}
  & $n=80, \tau_B=0.3$  & -0.51 & 17.0 & 15.3 & -100.5 & 320 \\
 && $n=160, \tau_B=0.3$ & -0.42 & 10.3 &  9.6 &  -74.1 & 240 \\
 && $n=80, \tau_B=0.5$  & -0.93 & 21.3 & 18.1 & -170.1 & 320 \\
 && $n=160, \tau_B=0.5$ & -0.78 & 12.0 & 10.7 & -124.6 & 243 \\
 \cmidrule(lr){1-8}
\multirow{6}[4]{*}{MD}
& \multirow{3}{*}{$\mt{AR}(0.6)$}
  & $n=80, \zeta=0.9$   & -0.26 & 14.2 & 13.6 &  -36.2 & 320 \\
 && $n=130, \zeta=0.7$  & -0.63 & 12.3 & 11.0 & -116.6 & 270 \\
 && $n=250, \zeta=0.5$  & -1.19 & 11.4 &  9.7 & -183.6 & 218 \\
 && $n=700, \zeta=0.3$  & -2.17 &  9.2 &  7.5 & -228.9 & 188 \\
 \cmidrule(lr){2-8}
& \multirow{3}{*}{$\mt{ER}$}
  & $n=80, \zeta=0.9$   & -0.26 & 13.4 & 12.7 &  -36.6 & 320 \\
 && $n=130, \zeta=0.7$  & -0.62 & 11.7 & 10.4 & -116.7 & 270 \\
 && $n=250, \zeta=0.5$  & -1.20 & 10.3 &  8.7 & -180.7 & 214 \\
 && $n=700, \zeta=0.3$  & -2.17 &  8.5 &  6.9 & -223.0 & 184 \\
\bottomrule
\end{tabular}
\label{tab:input_spectrums}
\end{table*}

Turning to the nodewise estimator, we see similar patterns. Again referring to Table~\ref{tab:input_spectrums}, it seems that the relative performance of the nodewise estimator varies significantly with the indefiniteness of the input matrix. When the input matrix is closer to positive semidefinite, such as the $n=160$ situations in Table~\ref{tab:methodcomp_kronsum_ar} or the $\zeta=0.9$ cases in Tables~\ref{tab:methodcomp_missdat_ar} and~\ref{tab:methodcomp_missdat_er}, it performs comparably in terms of model selection and significantly better in terms of norm error. But when the input matrix is very indefinite, such as the $\zeta=0.5$ cases in Tables~\ref{tab:methodcomp_missdat_ar} and~\ref{tab:methodcomp_missdat_er} its relative performance quickly degrades.

\begin{figure*}[tbh] \centering

\begin{subfigure}[t]{0.75\linewidth} \centering
\includegraphics[width=1\linewidth]{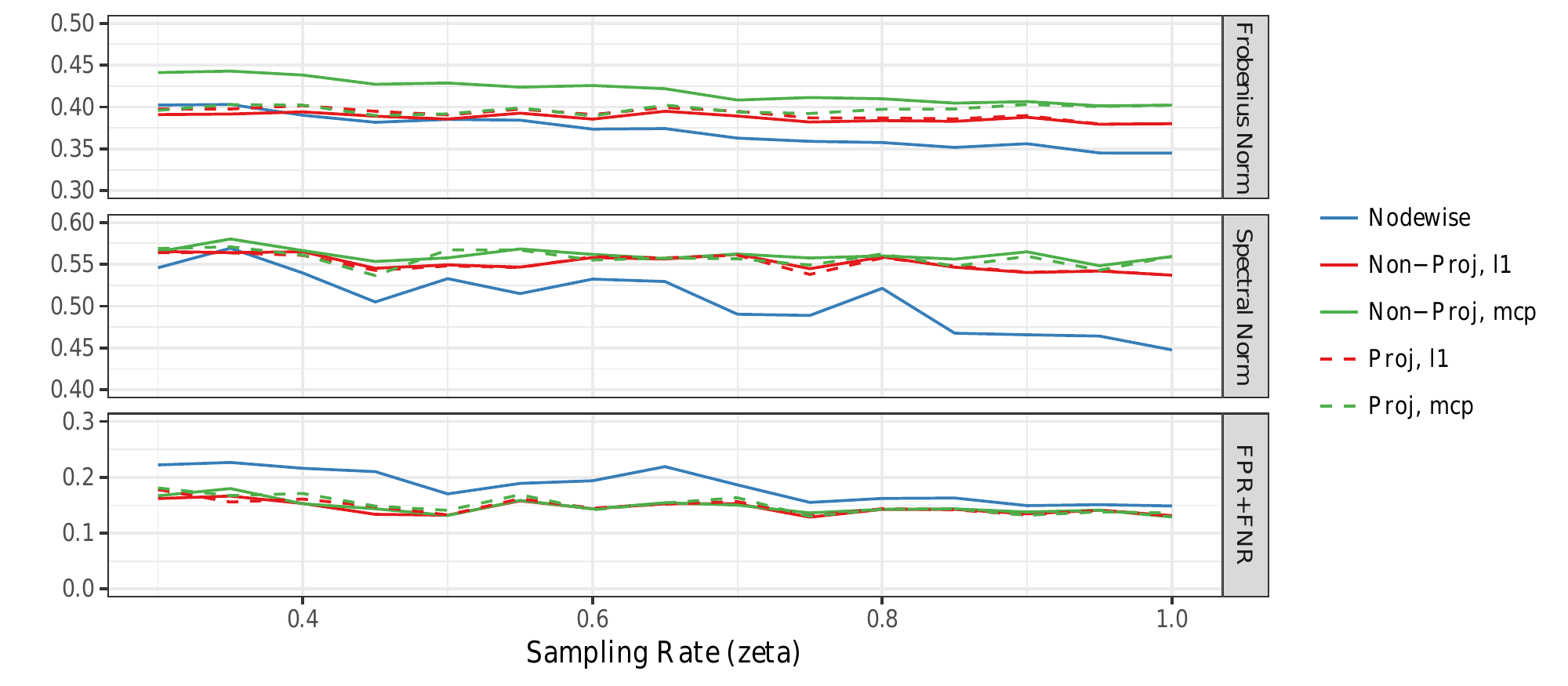}
\end{subfigure}
\begin{subfigure}[t]{0.75\linewidth} \centering
\includegraphics[width=1\linewidth]{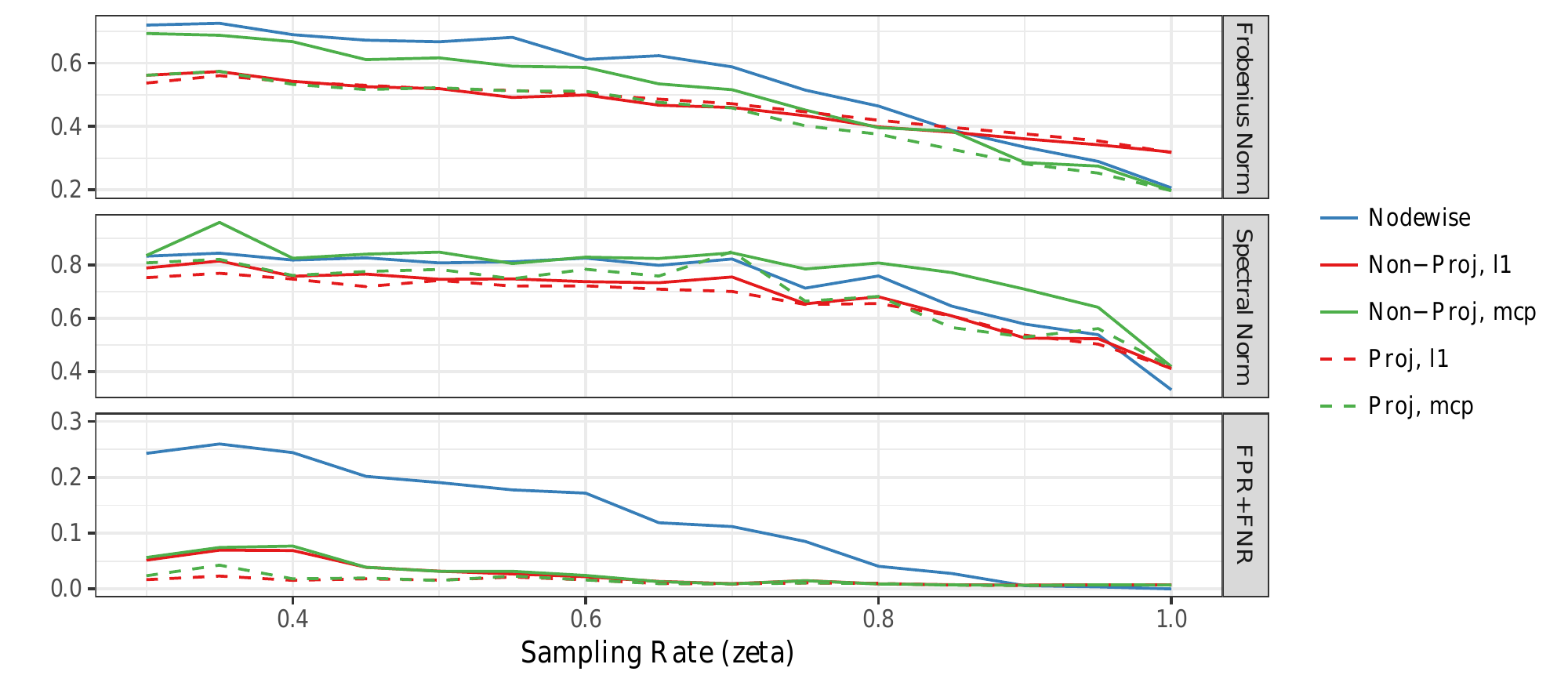}
\end{subfigure}

\caption{The performance of the various estimators for the missing data model as we vary the sampling rate. Note that these are minimums over a range of $\lambda$ values. For each $\zeta$, $n$ is chosen so that the effective sample size for estimating off-diagonal entries of the covariance is constant, so $n \zeta^2=80$. On the top panel, we set $A$ to be from an $\mt{AR}(0.4)$ with $m=400$, while the bottom panel uses $A$ as $\mt{AR}(0.8)$. In both cases, the effective sample size ($n\zeta^2$) is set at 80. The MCP penalty is chosen with $a=1.5$, and we set $R$ to be the 2 times the oracle value for each method.}
\label{fig:miss_ar_err_varyZeta}
\end{figure*}

Figure~\ref{fig:miss_ar_err_varyZeta} demonstrates the patterns that we observed in Figures~\ref{fig:missdat_ar_projvsnon}. Again, we vary the sampling rate $\zeta$ and fix the effective sample size for estimating off-diagonal entries of the covariance matrix ($n \zeta^2$), so the $\ell_\infty$ rate of $\Gammahat$ is kept constant. As the sampling rate decreases, the magnitude of the most negative eigenvalue in the covariance estimate increases, which we can see negatively affects the relative performance of the nodewise and nonprojected MCP methods. These methods are the best for high sampling rates but the worst when there is more missing data. The other methods are not as sensitive.

We can also see how the size of the off-diagonal entries in the precision matrix affect the potential benefits of the nonconvex penalties. In the top panel, which has small off-diagonal entries, the MCP estimators are consistently worse. But in the bottom panel, which has larger off-diagonal entries, the nonconvex penalties have better Frobenius norm performance when the sampling rate is high, though this advantage goes away as the sampling rate drops.

\subsection{Comparison of side constraints}
\label{subsec:side_comp}

Here we compare using the operator norm side-constraint in \eqref{eq:glasso_opt_problem_side} to the $\ell_1$-side constrained version considered in \citet{lohwainwright15}. Note that theoretically \citet{lohwainwright17} show that under certain conditions the former can attain model selection without incoherence and spectral norm convergence under the scaling $n > d^2 \log p$ (where $d$ is the maximal node degree), which has not been shown with the latter.

\begin{figure*}[tbhp] \centering
\captionsetup[subfigure]{justification=centering}

\begin{subfigure}[t]{0.45\linewidth} \centering
\includegraphics[width=\linewidth]{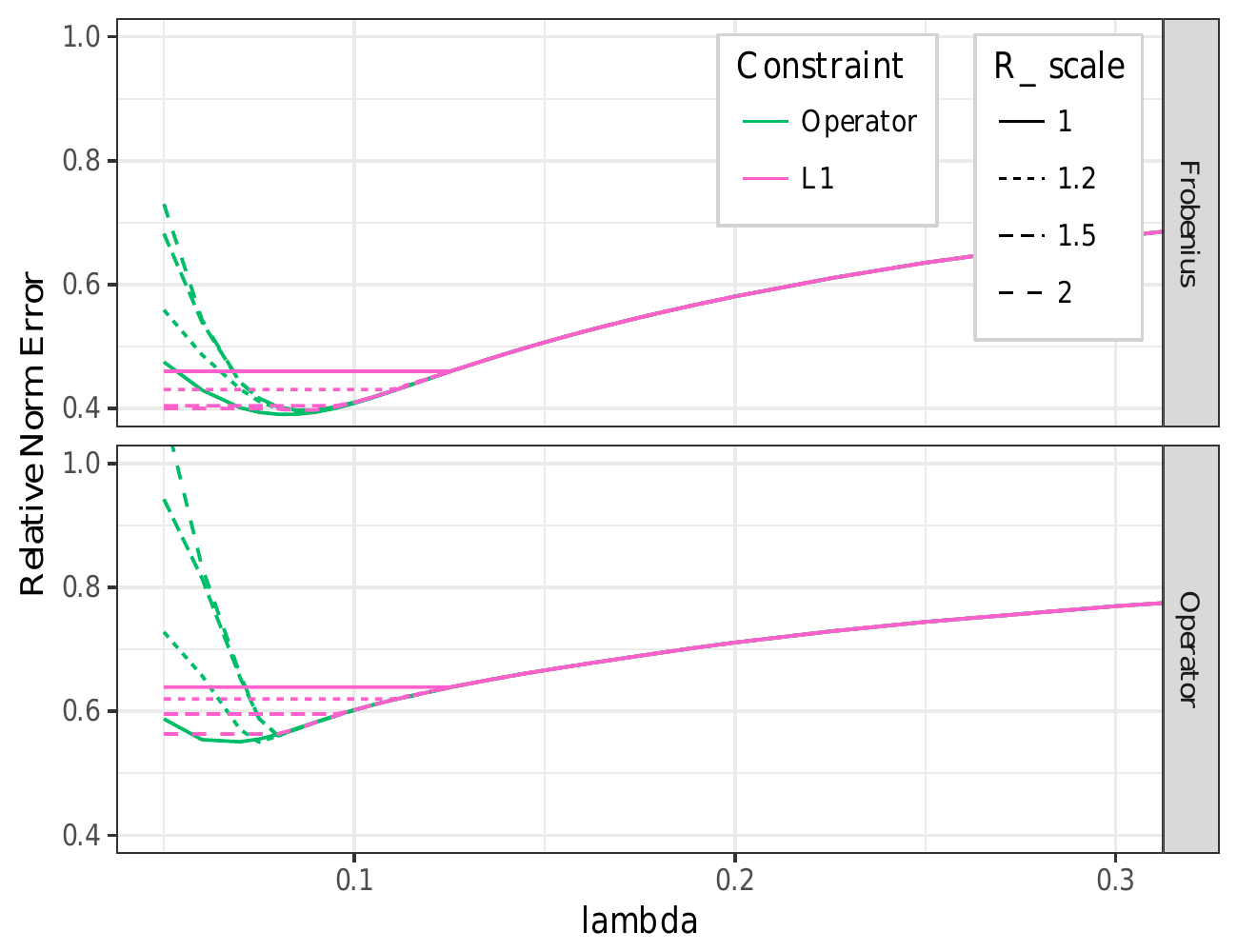}
\caption{$\mt{AR1}(0.8)$, $m=200$, $n=100$, $\zeta=0.8$}
\end{subfigure}
\begin{subfigure}[t]{0.45\linewidth} \centering
\includegraphics[width=\linewidth]{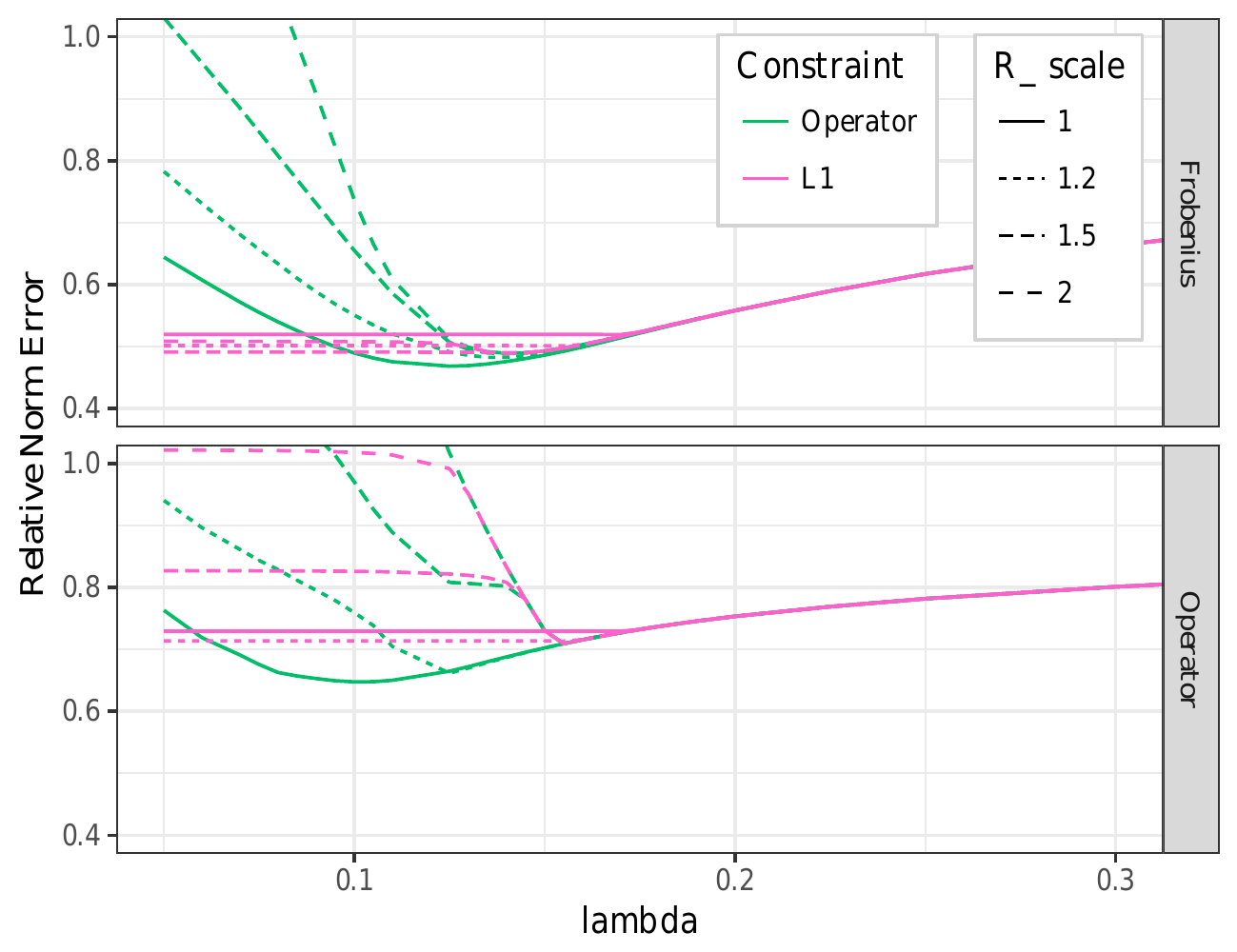}
\caption{$\mt{AR1}(0.8)$, $m=200$, $n=150$, $\zeta=0.6$}
\end{subfigure}
\\ \vspace{5pt}
\begin{subfigure}[t]{0.45\linewidth} \centering
\includegraphics[width=\linewidth]{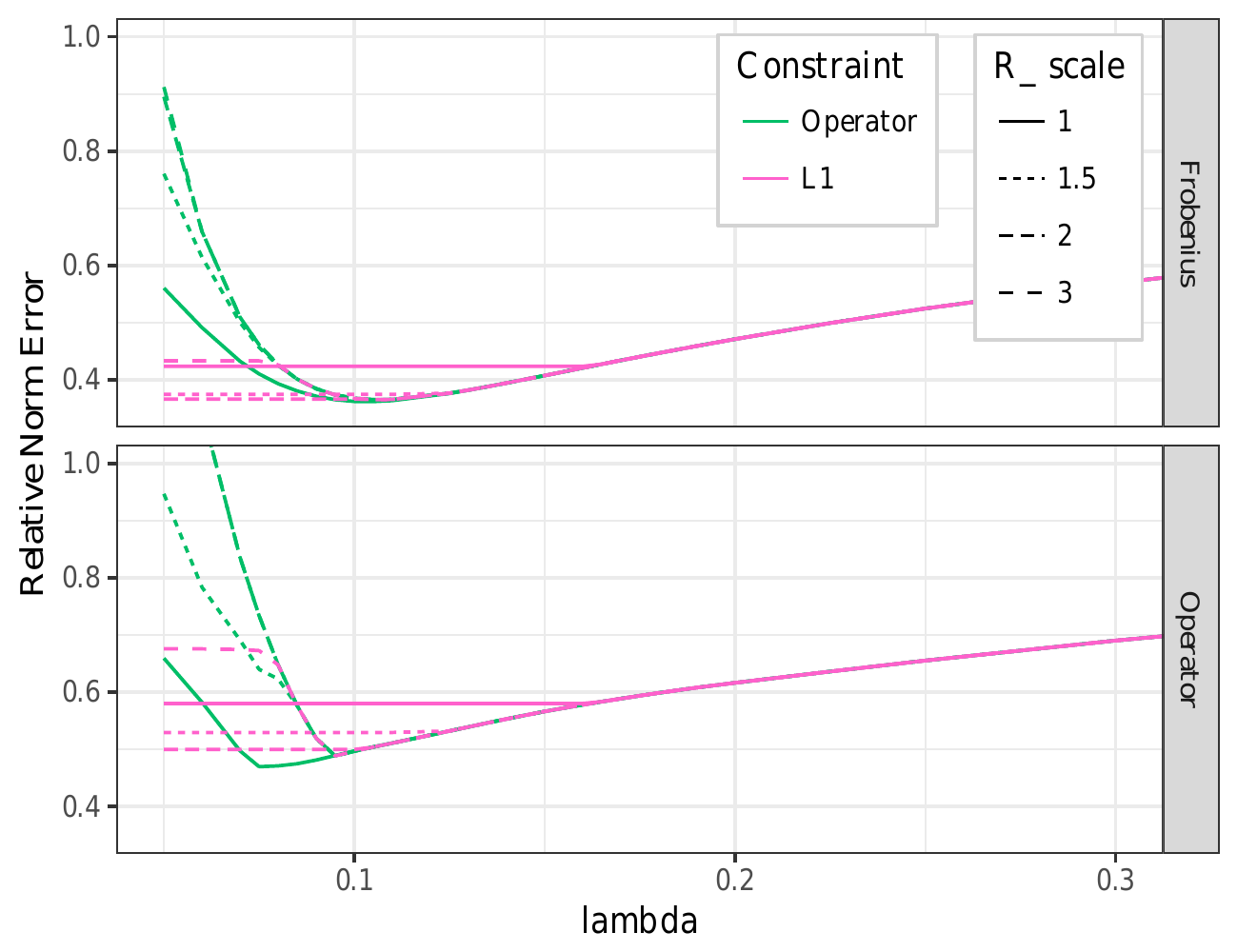}
\caption{$\mt{SB}(0.7)$, $m=120$, $n=100$, $\zeta=0.8$}
\end{subfigure}
\begin{subfigure}[t]{0.45\linewidth} \centering
\includegraphics[width=\linewidth]{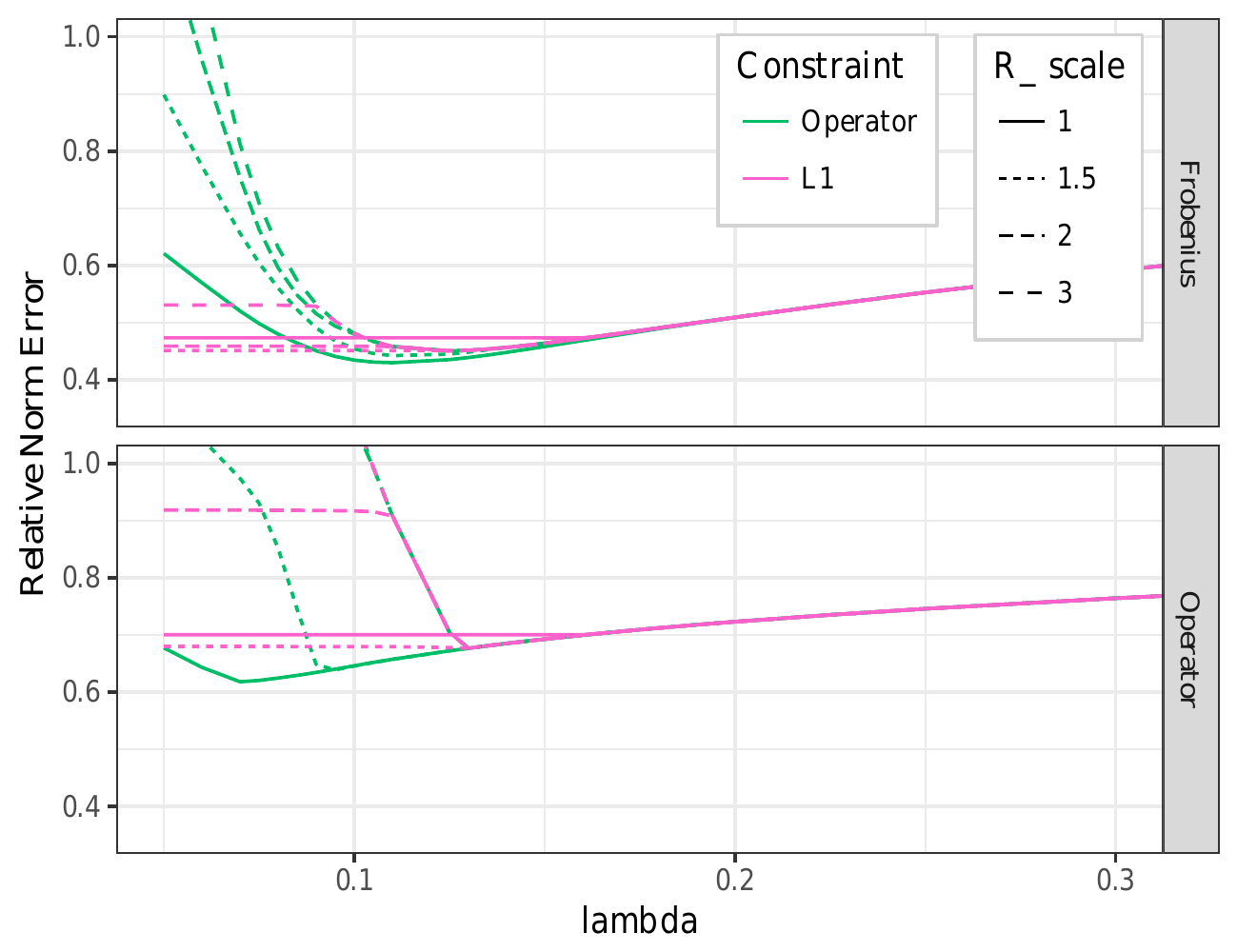}
\caption{$\mt{SB}(0.7)$, $m=120$, $n=200$, $\zeta=0.6$}
\end{subfigure}

\caption{Comparing the performance of the operator- and $\ell_1$-norm side constrained estimators. For each model, errors in terms of relative Frobenius norm (top panel) and relative operator norm (bottom panel) are shown. For each method, $R$ was set to R\_scale times the oracle value and an $\ell_1$ penalty was used. The Star-Block model uses block sizes of 30.}
\label{fig:miss_l1side_comp}
\end{figure*}

Figure~\ref{fig:miss_l1side_comp} shows the performance in terms of relative norm error for various missing data model scenarios. For large values of $\lambda$ we can see that the two estimators are identical, as the side-constraints are not active.

As the penalty $\lambda$ shrinks, when the $\ell_1$ side constraint is used, selecting $R$ is akin to performing selection on the minimum value of $\lambda$ to use, since only one of the penalty and side-constraint can be active at a time. Following the green lines, we can see that the regularization from the operator norm side-constraint can improve results. This means that $\ell_1$ side-constrained estimator misses this additional improvement. At worst, for larger values of $R$ the operator norm-constrained simply has identical performance to the $\ell_1$-constained version as long as $\lambda$ is appropriately selected.

\clearpage
\section{Additional data analysis}
\label{sec:data_analysis_supp}

As discussed in Section~\ref{sec:data_analysis}, we collected voting records data from the Senate during the 112th US Congress, which was from January 3, 2011 to January 3, 2013. This data is part of the public record and open-source code to download and process the data can be found at \url{https://github.com/unitedstates/congress}.

Due to changes in membership, there are data on 102 senators, which we drop three of due to serving incomplete terms. The data contains 486 votes in total. We drop votes that are unanimous or unanimous within both parties, resulting in 426 votes. Roughly 2.6\% of values are missing in this data.

We use the ADMM algorithm from Section~\ref{sec:ADMM} to estimate the nonprojected version of \eqref{eq:glasso_opt_problem_side} with an $\ell_1$ penalty to estimate conditional dependence graphs among senators. Since this is an exploratory analysis, we ran estimators using various levels of the penalization parameter $\lambda$ and have chosen plots to display based on the number of estimated edges and maintaining visual clarity.

For our preliminary analysis, we use a modified version of the missing data estimator as described in \citet{zhou15}, where bills have varying missing probabilities while we estimate the edges among senators. We also demean each vote by political party, similar to the demeaning done in \citet{hornstein2018joint}. See our future work for a more detailed study of this estimator and its properties.

\begin{figure}[tbh] \centering
\includegraphics[width=0.95\linewidth]{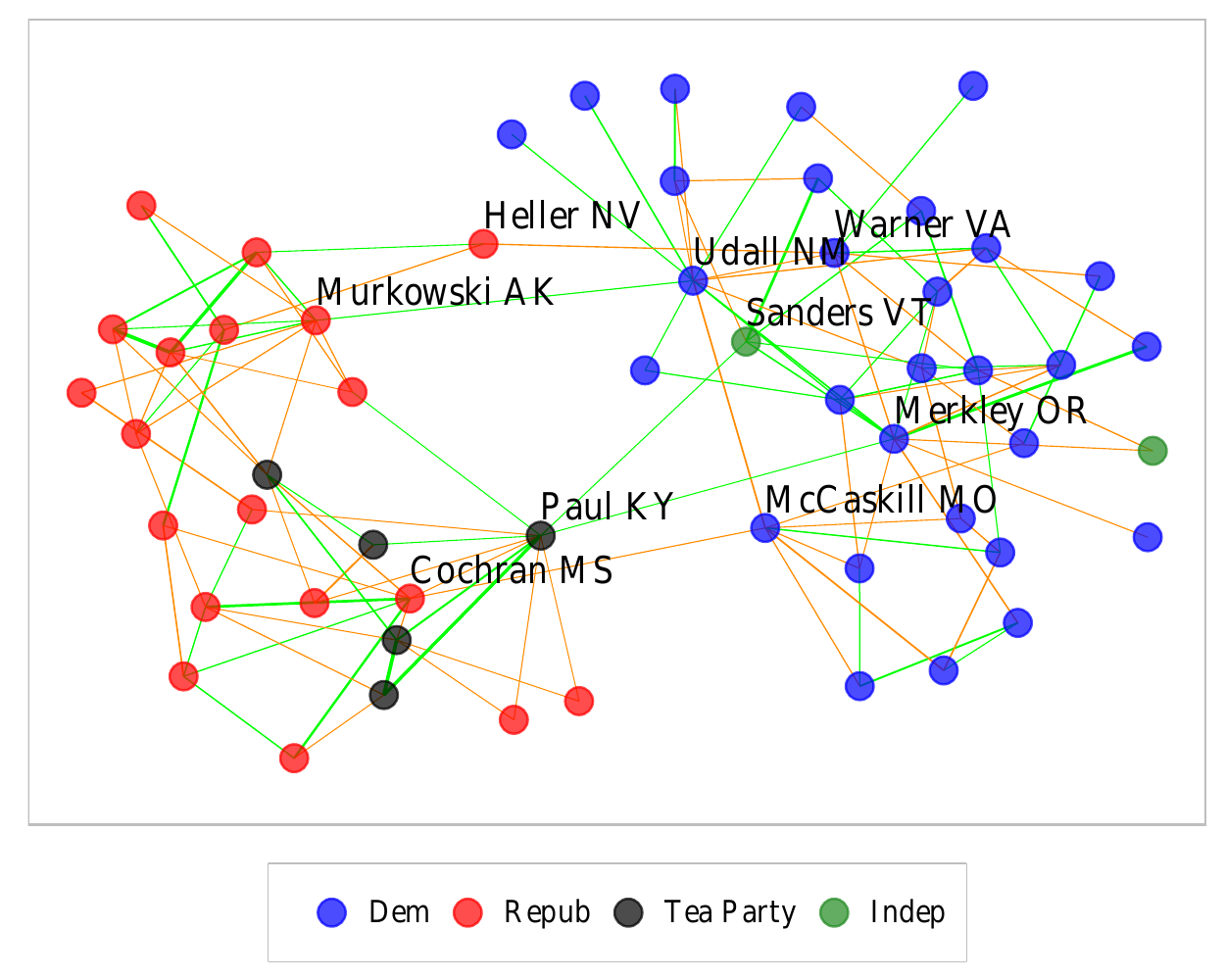}

\caption{The subgraph of nodes that are 1- or 2-steps removed from the opposing party from \ref{fig:senate_graphs_2}. Edge color and thickness is determined by the size of the partial correlation estimate, with green denoting positive orrelation and orange denoting negative.}
\label{fig:senate_graphs_links}
\end{figure}

\begin{figure}[tbh] \centering
\includegraphics[width=0.95\linewidth]{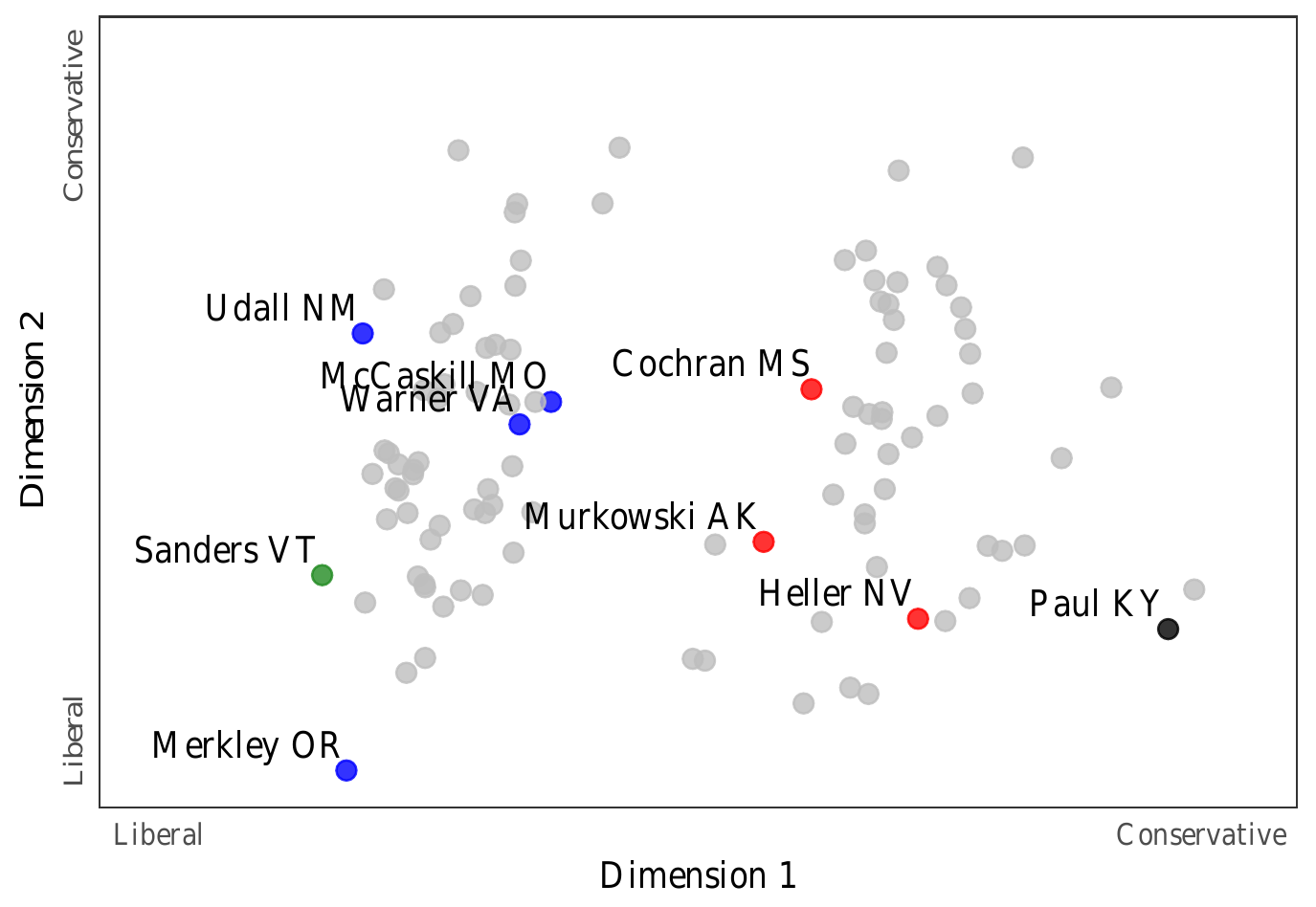}

\caption{NOMINATE scores for the senators of the 112th Congress. Data are from \url{https://voteview.com/about}.}
\label{fig:nominate}
\end{figure}

Figure~\ref{fig:senate_graphs_links} plots the subgraph of senators with cross-party connections or links to those with cross-party connections from  Figure~\ref{fig:senate_graphs_2}. We look at the NOMINATE scores of these senators (Figure~\ref{fig:nominate}) to determine their positions on the political spectrum. NOMINATE is a probabilistic geometric model that places each senator in a two-dimensional space representing their ideological beliefs \citep{poole2005spatial}.

Although most of the linked senators are are either on the extremes or are moderates (see Section~\ref{sec:data_analysis}), the main exception to this is Dean Heller, who is linked to Mark Warner. This outlier connection is perhaps worth further investigation as to why they are linked.

The McCaskill-Cochran connection is unsurprising, as both are among the most moderate senators from their respective parties. The Udall-Murkowski connection is more interesting since Tom Udall is viewed as a relatively liberal Democrat while Murkowski is a moderate Republican. Murkowski connecting across the aisle is expected, but why she would be linked to Udall is unknown.

Paul-Sanders and Paul-Merkley are both interesting connections since they are between left-wing and right-wing senators. In fact, these three senators form a triangle of positively correlated edges. This could be because they are all ideological extremists and therefore vote together against bills with moderate support. Or this could be explained by the fact that all three senators are liberal on NOMINATE's second dimension, and therefore vote together on those issues.

\begin{figure}[tbh] \centering
\begin{subfigure}[t]{0.95\linewidth} \centering
\includegraphics[width=\linewidth]{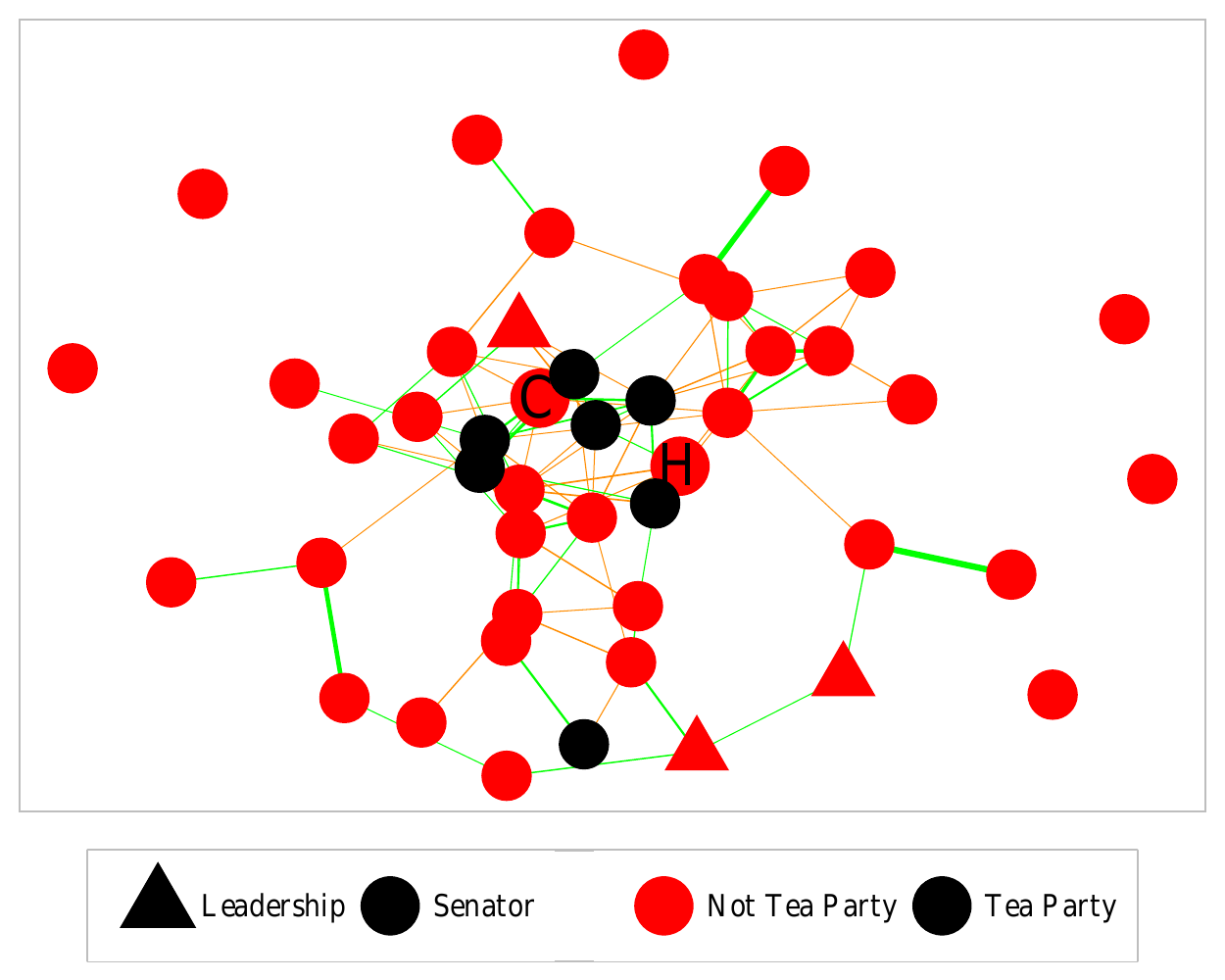}
\caption{Republican subgraph}
\end{subfigure}
\begin{subfigure}[t]{0.95\linewidth} \centering
\includegraphics[width=\linewidth]{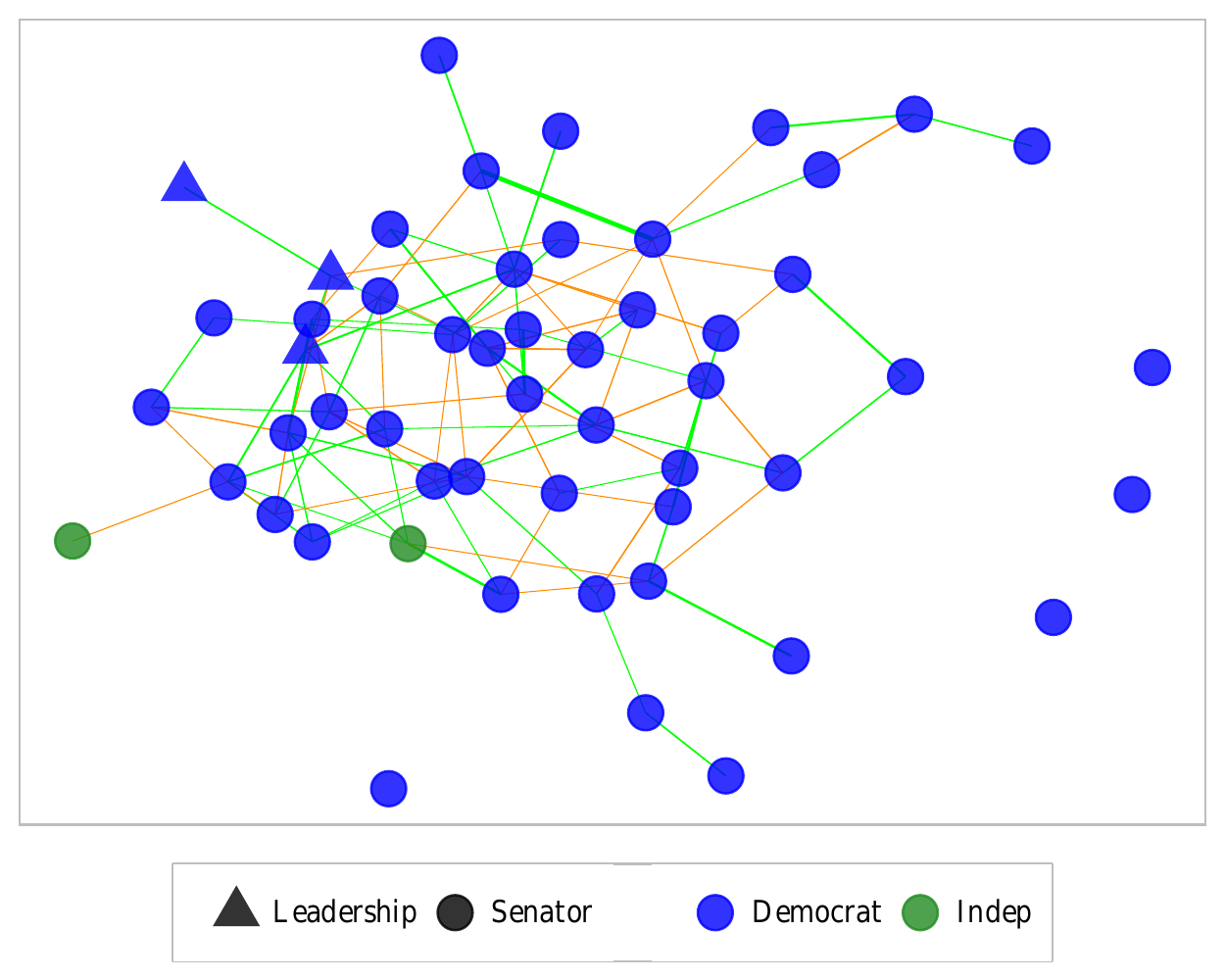}
\caption{Democratic subgraph}
\end{subfigure}

\caption{Party subgraphs from Figure~\ref{fig:senate_graphs_1} (equivalent to Figure~\ref{fig:senate_graphs_r}). Edge color and thickness is determined by the size of the partial correlation estimate, with green denoting positive correlation and orange denoting negative. These are estimated with $\lambda = 0.21$ and are thresholded at 0.04.}
\label{fig:senate_graphs_party}
\end{figure}

Figure~\ref{fig:senate_graphs_party} shows the estimated Republican and Democratic subgraphs with edge weights and directions as determined by the partial correlation esimates, while Table~\ref{tab:edges} lists the top ten edges by strongest positive partial correlation.

\begin{table}[htbp]
\centering \small
\caption{Highest estimated partial correlations among Republicans (top) and Democrats (bottom).}
\begin{tabular}{llccc}
\toprule
 \multicolumn{2}{c}{Senators} & \multicolumn{2}{c}{States} & Tea Party \\
\cmidrule(lr){1-2} \cmidrule(lr){3-4} \cmidrule(lr){5-5}
Enzi & Barrasso & WY & WY & No-No \\
Crapo & Risch & ID & ID & No-No \\
Chambliss & Isakson & GA & GA & No-No \\
Snowe & Collins & ME & ME & No-No \\
DeMint & Lee & SC & UT & Yes-Yes \\
Collins & Brown & ME & MA & No-No \\
Lee & Paul & UT & KY & No-No \\
Cochran & Hoeven & MS & ND & No-No \\
Coburn & Johnson & OK & WI & Yes*-Yes \\
Cochran & Wicker & MS & MS & No-No \\
\midrule
Reed & Whitehouse & RI & RI \\
Murray & Cantwell & WA & WA \\
Wyden & Merkley & OR & OR \\
Udall & Bennet & CO & CO \\
Leahy & Sanders & VT & VT \\
Carper & Coons & DE & DE \\
Mikulski & Akaka & MD & HI \\
Feinstein & Boxer & CA & CA \\
Baucus & Tester & MT & MT \\
Akaka & Gillibrand & HI & NY \\
\bottomrule
\end{tabular}
\label{tab:edges}
\end{table}

We can see that for both parties most of the strongest links are between senators from the same state, which is expected since these senators must appeal to the same constituents and therefore are likely to vote similarly. Of the 33 states where both senators are in the same party, 19 of the pairs are linked and 13 of those appear in the top connections in Table~\ref{tab:edges}.

Looking at the exceptions closely, we can often identify why these links are particularly strong. For instance, DeMint-Lee and Coburn-Johnson are both between senators we have identified as being tea party-linked (Coburn due to his close proximity and support in the media). Also, Collins-Brown is between two moderate Republicans who are both from Democratic New England states.

Table~\ref{tab:edges_neg} also contains the most negatively estimated partial correlations. Note that the negative relationships we find are in general of much smaller magnitude than the positive correlations. For instance, the most negative correlation (Akaka-McCaskill at -0.13) has roughly same magnitude as the 34th-most positive correlation (DeMint-Johnson).

\begin{table}[htbp]
\centering \small
\caption{Most negative estimated partial correlations among senators.}
\begin{tabular}{llccc}
\toprule
 \multicolumn{2}{c}{Senators} & \multicolumn{2}{c}{States} & Party \\
\cmidrule(lr){1-2} \cmidrule(lr){3-4} \cmidrule(lr){5-5}
Akaka & McCaskill & HI & MO & D \\
Hatch & Hoeven & UT & ND & R \\
Hoeven & Toomey & ND & PA & R \\
Alexander & Rubio & TN & FL & R \\
Levin & Tester & MI & MT & D \\
McCaskill & Udall & MO & NM & D \\
Corker & Wicker & TN & MS & R \\
Collins & Johnson & ME & WI & R \\
Cardin & Hagan & MD & NC & D \\
Bingaman & Casey & NM & PA & D \\
\bottomrule
\end{tabular}
\label{tab:edges_neg}
\end{table}

\subsection{Nodewise regression}

\begin{figure}[tbh] \centering
\begin{subfigure}[t]{0.95\linewidth} \centering
\includegraphics[width=\linewidth]{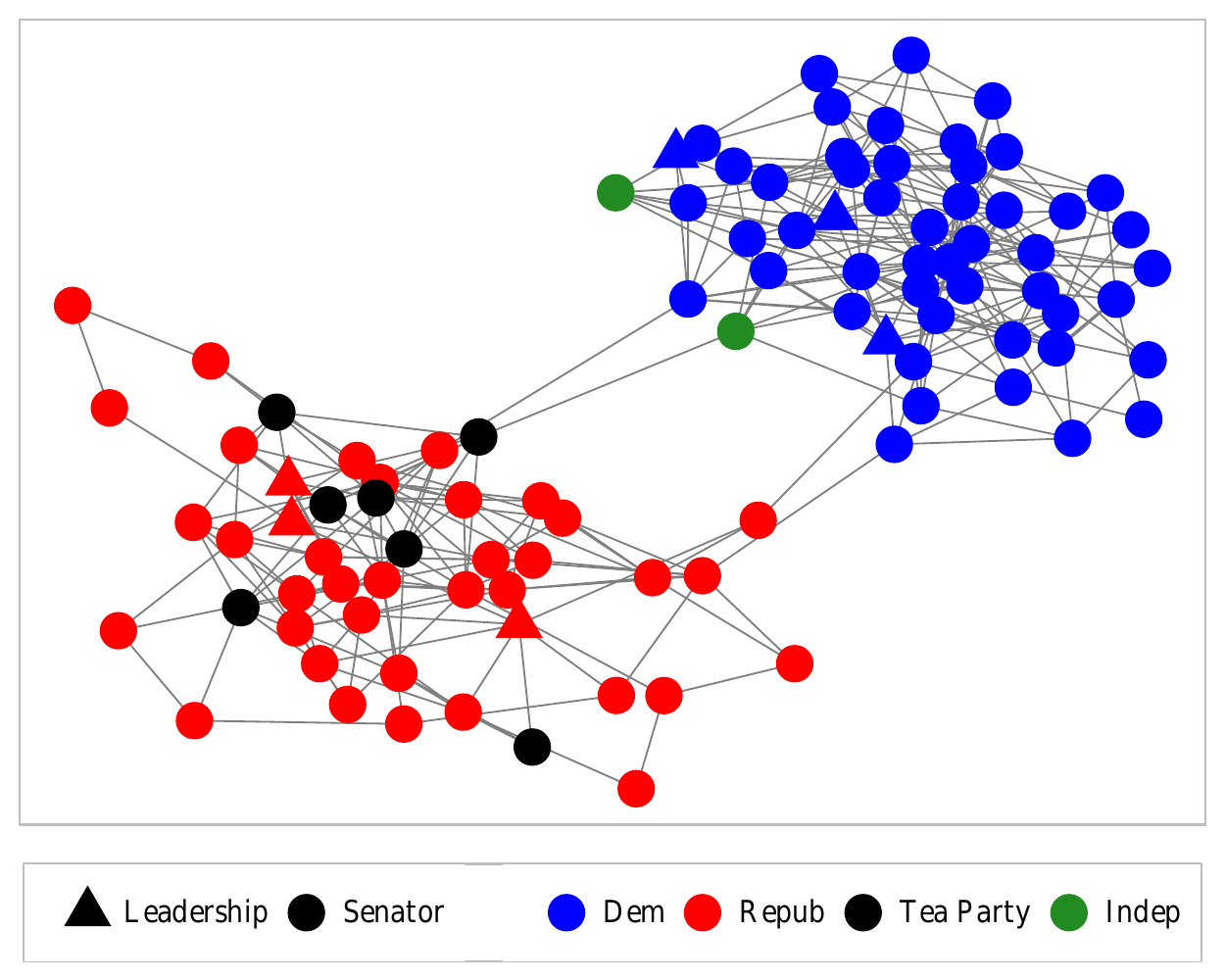}
\caption{All Senators}
\end{subfigure} \\
\begin{subfigure}[t]{0.95\linewidth} \centering
\includegraphics[width=\linewidth]{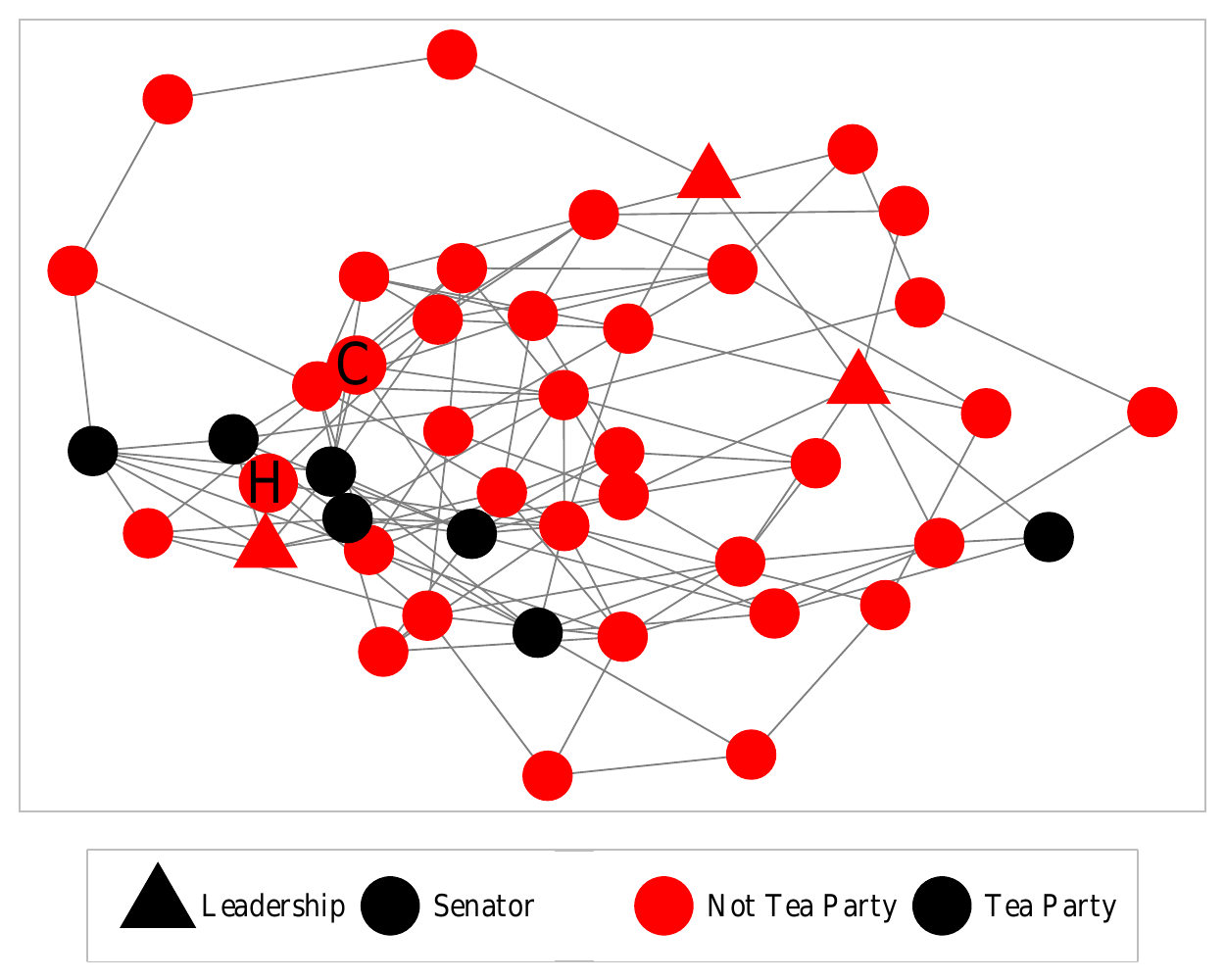}
\caption{Republican subgraph}
\label{fig:senate_graphs_nodewise_r}
\end{subfigure}

\caption{Graphs among 112th Congress senators estimated with nodewise regression. We set $\lambda=0.12$ and $R=10$. After estimation, the precision matrix is thresholded at 0.03.}
\label{fig:senate_graphs_nodewise}
\end{figure}

We also demonstrate the usage of nodewise regression in Figure~\ref{fig:senate_graphs_nodewise}. Since the sampling rate for this dataset is fairly high, we expect nodewise regression to perform well, and the results are overall similar to those in Figure~\ref{fig:senate_graphs_2}. Of the four cross-party links, three (Cochran-McCaskill, Sanders-Paul, and Warner-Heller) are also present in the previously estimated graph. The link between Democrat Mark Begich and Lisa Murkowski is new, a natural one since they both represent Alaska in the Senate.

Figure~\ref{fig:senate_graphs_nodewise_r} exhibits similar patterns to those identified in Figure~\ref{fig:senate_graphs_r}. Hatch and Coburn, marked as `H' and `C,' still appear to be closely connected to the tea party cluster. Moran is also still disconnected from the rest of the tea party despite attending the inaugural meeting of the Senate Tea Party Caucus.

\end{document}